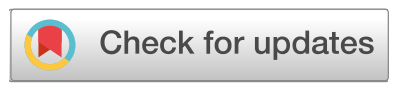

OPEN

# Impact of canny edge detection preprocessing on performance of machine learning models for Parkinson's disease classification

Sameer Bhat✉ & Piotr Szczuko

**This study investigates the classification of individuals as healthy or at risk of Parkinson's disease using machine learning (ML) models, focusing on the impact of dataset size and preprocessing techniques on model performance. Four datasets are created from an original dataset: $DS_0$ (normal dataset), $DS_1$ ($DS_0$ subjected to Canny edge detection and Hessian filtering), $DS_2$ (augmented $DS_0$), and $DS_3$ (augmented $DS_1$). We evaluate a range of ML models-Logistic Regression (LR), Decision Tree (DT), Random Forest (RF), Gradient Boosting (GB), XGBoost (XBG), Naive Bayes (NB), Support Vector Machine (SVM), and AdaBoost (AdB)-on these datasets, analyzing prediction accuracy, model size, and prediction latency. The results show that while larger datasets lead to increased model memory footprints and prediction latencies, the Canny edge detection preprocessing supplemented by Hessian filtering (used in $DS_1$ and $DS_3$) degrades the performance of most models. In our experiment, we observe that Random Forest (RF) maintains a stable memory footprint of 61 KB across all datasets, while models like KNN and SVM show significant increases in memory usage, from 5.7-7 KB on $DS_0$ to 102-220 KB on $DS_2$, and similar increases in prediction time. Logistic Regression, Decision Tree, and Naive Bayes show stable memory footprints and fast prediction times across all datasets. XGBoost's prediction time increases from 180-200 ms on $DS_0$ to 700-3000 ms on $DS_2$. Statistical analysis using the Mann-Whitney U test with 100 prediction accuracy observations per model (98 degrees of freedom) reveals significant differences in performance between models trained on $DS_0$ and $DS_2$ (p-values < 1e-34 for most models), while the effect sizes measured by estimating Cliff's delta values (approaching $-1$) indicate large shifts in performance, especially for SVM and XGBoost. These findings highlight the importance of selecting lightweight models like LR and DT for deployment in resource-constrained healthcare applications, as models like KNN, SVM, and XGBoost show significant increases in resource demands with larger datasets, particularly when Canny preprocessing is applied.**

Over one million people in the U.S. have Parkinson's disease (PD), a number that is expected to rise to 1.2 million by 2030. This condition affects more than 10 million people globally, with approximately 60,000 new diagnoses occurring annually in the United States alone[1]. PD is a neurodegenerative condition that affects motor function and, in severe cases, can also impact cognitive abilities[2]. Symptoms of PD can vary widely, encompassing both movement difficulties and non-motor signs like cognitive impairment and depression[3]. Primarily affecting older adults, PD is caused by the death of dopaminergic neurons in the brain's substantia nigra[4]. One of the early signs of PD can be issues with handwriting, which affects millions of people worldwide. While clinical diagnosis relies on examining symptoms and medical history, recent studies are exploring handwriting changes due to its prevalence[4–9]. Modern healthcare requires effective early detection tools, particularly since initial signs of conditions may appear before retirement, which can reduce patients' work capacities. The Unified Parkinson's Disease Rating Scale (UPDRS) and its revised version by the Movement Disorder Society (MDS-UPDRS) serve as guidelines for clinical assessments of PD[10–13]. However, the accuracy of clinical diagnoses can vary, which has led to the exploration of alternative methods, such as computer-based handwriting analysis[4,6,14–18]. Changes in handwriting, especially micrographia, may serve as a potential biomarker for PD[19–21].

Faculty of Electronics, Telecommunications and Informatics, Multimedia Systems Department, Gdansk University of Technology, Narutowicza 11/12, 80-233 Gdansk, Poland. ✉email: sameer.bhat@pg.edu.pl

This paper reviews the diagnosis of PD through handwriting spiral analysis, emphasizing the role of machine learning (ML). Given ML's promise in identifying PD via such analyses, it has the potential for early detection, monitoring, and improving patient outcomes[22,23]. Handwriting is a complex skill that involves cognitive, visual, and motor abilities. Although handwriting difficulties are not a diagnostic criterion for PD, they often lead individuals to seek medical attention. Recent studies have investigated the use of ML to assess the severity of PD. These studies propose improved algorithms and utilize MRI scans and deep learning techniques for more accurate classification of the disease[16,24–26]. Other approaches utilize ML to predict PD with non-motor symptoms, extracting features from digital drawing tests or employing convolutional neural networks (CNNs) for end-to-end processing of handwriting images[7,27,28]. Research by[29] introduces a Continuous Convolution Network (CC-Net) designed to recognize writing disorders based on hand-drawn images.[30] evaluates the role of ML in assessing neurocognitive features associated with PD, achieving high accuracy in distinguishing between individuals with PD and healthy controls. Additionally,[31] conducts spiral drawing assessment tests using ML on both static and dynamic drawings, demonstrating impressive accuracy in predicting PD. Recent developments have explored innovative methods that surpass existing benchmarks,[32] demonstrates superior performance using deep convolutional neural networks (CNNs) combined with transfer learning. Additionally,[33] utilizes computer vision and machine learning to predict PD from hand-drawn wave and spiral images. Furthermore, ML, particularly CNNs, has been shown to outperform traditional classifiers in the analysis of EEG signals[34]. Diverse CNNs, SVM classifiers, and SqueezeNet CNNs are effective in analyzing spiral drawings[35,36]. Tablet-based technologies utilizing random forest classifiers demonstrate high accuracy in Parkinson's disease detection[37]. ML demonstrates considerable potential for detecting Parkinson's disease through handwriting analysis. It plays a crucial role in the accurate diagnosis, classification, and monitoring of the disease by utilizing early signs and various modalities, including handwriting analysis, neurocognitive features, and imaging techniques. These advancements illustrate the evolving landscape of Parkinson's disease diagnosis and highlight ML's contribution to improving diagnostic accuracy, accessibility, and early detection. This promising direction suggests a bright future for research in neurodegenerative diseases[4,14].

As advancements in PD detection continue, this study contributes to the evolving field by addressing an important aspect that is often overlooked: the memory requirements of ML models used for predicting PD. While many studies have developed ML and deep learning models for PD diagnosis, this research fills a critical gap by focusing on the memory footprint and prediction latency of these models. The main motivation for this study arises from the important role that ML models play in predicting and classifying PD. This aligns with the growing demand for applications in healthcare systems that prioritize low computational power and minimal memory usage. By examining the memory requirements of ML models, particularly in resource-constrained devices or edge computing platforms, this study aims to improve the practicality and accessibility of PD diagnosis. Linking this effort to the broader context of PD research highlights a multidimensional approach. Advancements in ML techniques not only enhance diagnostic accuracy but also help address practical challenges in healthcare applications. As the field of PD detection continues to evolve, studies like these pave the way for improved diagnostic capabilities and better patient outcomes.

### Research contribution

The main contributions of this study are as follows:

- Comparative analysis of ML models: The study compares various ML models to classify patients as either healthy or affected by PD. This analysis helps identify the most effective models for accurate diagnosis.
- Exploration of feature space: The research explores the feature space to detect essential characteristics for training and evaluating the ML models. This investigation sheds light on critical features that contribute to accurate predictions
- Explainability of ML models: The study prioritizes explainable ML, providing insights into the feature extraction process and analysis. Understanding the reasoning behind model decisions enhances trust and interpretability in clinical applications
- Memory-efficient models: By examining the memory footprint of different ML models trained on two datasets, the study identifies models with low memory requirements. This finding is crucial for implementing ML-based systems on memory-constrained or edge-computing devices.
- Prediction time analysis: The study discusses prediction times for each ML model. It compares the prediction times of different models, providing valuable information on the computational efficiency of the models, which is essential for real-time clinical applications.

## Literature review

### Handwriting analysis

Handwriting issues, often an early sign prompting medical attention in PD, are a focus of recent studies. Aural cueing positively influences handwriting, contrasting with varied findings for visual feedback. Handwriting problems, including micrographia, may serve as clinical markers for early-stage diseases. Kinematic analysis using digital tablets has significantly advanced PD handwriting research. Studies like McLennan's[20] on micrographia gathered samples from PD patients, noting the challenge of quantifying such qualitative information. M. Ranzato et al.'s work[21] identifies PD using Restricted Boltzmann Machines and characteristics from handwritten images, emphasizing the significance of analyzing tremors. The study utilized photographs from the "HandPD" collection, including data from tasks performed by healthy individuals and PD patients undergoing handwriting evaluations. In PD diagnosis, handwriting complexities are pivotal, with various studies exploring kinematics, micrographia, and image-based approaches. As a clinical marker and an area rich in qualitative information, handwriting analysis holds promise for enhancing early PD detection and understanding disease progression[19–21].

## Machine learning and Parkinson's disease

ML algorithms showcase significant promise in identifying Parkinson's Disease (PD) through handwriting spiral analysis, particularly in the context of the spiral drawing test. This fundamental test for assessing fine motor skills in PD patients provides rich data that ML algorithms can effectively analyze to detect and monitor the disease. Studies have employed ML algorithms to assess various parameters of spiral drawings, including size, shape, and speed, demonstrating high accuracy in differentiating between PD patients and healthy controls. Moreover, ML algorithms prove valuable in tracking changes in spiral drawing parameters over time, offering insights into disease progression and treatment response. For instance, in a study by[38], ML algorithms effectively detected motor function improvements in PD patients following deep brain stimulation surgery. Integration of multiple data sources, such as clinical information alongside spiral drawing analysis, further enhances diagnostic accuracy. Studies by[4,14] exemplify this approach, showing improved PD detection accuracy through combined analyses. ML emerges as a robust tool for detecting PD using handwriting spiral analysis, offering not only accurate identification of PD but also the potential for monitoring disease dynamics and treatment outcomes. As digital health technology advances, ML's ability to analyze diverse characteristics becomes increasingly valuable, paving the way for enhanced diagnostic capabilities and personalized patient care[22,23].

## Related works

In the evolving landscape of Parkinson's disease (PD) diagnosis, advanced technologies, particularly deep learning and artificial intelligence (AI), have played a pivotal role. Notably, studies such as Khatamino et al.[39] showcased the effectiveness of CNNs in classifying the HW dataset, achieving high accuracy in distinguishing PD patients. This approach, marked by early stopping and efficient feature extraction, sets a promising precedent for accurate PD diagnosis. Unique perspectives emerged in studies like[40], where patterns from digital Luria's alternating series tests were analyzed for PD diagnosis. The emphasis on kinematic and geometric features not only aids in disease detection but also enhances the interpretability of classification decisions. Johri and Tripathi et al.[41] introduced a cost-effective classifier targeting gait and speech impairment, leveraging bi-directional GRUs and 1D convolution principles. The study's modular approach, focusing on walking styles and speech distortion, contributes to promising performance values in PD diagnosis. Tuncer and Dogan's[42] novel multi-pooling technique addressed gender, PD, and combined gender-PD classification problems, showcasing a lightweight yet accurate methodology. In[18], a numerical method for grading functional tremor variations demonstrated superior performance using a perceptual colour representation-based classifier (PCRC). The integration of PCRC into a decision support system highlighted its potential for clinical applications and at-home monitoring. Studies examining minute handwriting and voice variances, such as[43], demonstrated the efficacy of CNNs in extracting visual features for PD diagnosis. Additionally, the system developed in[17], utilizing SVM for pattern identification, provided a promising and accessible solution.Finally, the collective findings, underscored by studies like[44], Chakraborty et al.[45], Nõmm et al.[46], and Tuncer et al.[47], collectively contribute to advancing PD diagnostics through diverse methodologies, paving the way for enhanced diagnostic capabilities and improved patient outcomes.

In recent developments in Parkinson's disease (PD) detection, several studies have introduced innovative methods, surpassing existing benchmarks and showcasing significant advancements. In[32], a deep convolutional neural network with transfer learning and data augmentation techniques demonstrated superior performance on a benchmark dataset, achieving 98.28% accuracy through fine-tuning with ImageNet. Handwriting and hand-drawn subjects serve as crucial indicators of PD, as suggested by[33]. The study leverages computer vision and machine learning techniques to predict PD from hand-drawn wave and spiral images. Using classification algorithms with the HOG feature descriptor algorithm, Gradient Boosting and K-Nearest Neighbors emerged as top performers with 86.67% and 89.33% accuracy rates, respectively. Functional tremor variations were scrutinized in[8], where a proposed model using polar expression features and a decision-making classifier achieved remarkable accuracy rates of 98.93%. This outperformance of traditional machine learning classifiers underlines the effectiveness of the suggested model. Vision transformers (ViT) entered the scene in[48], where a ViT-based model for handwritten data achieved an accuracy of 92.37% for detecting PD based on spiral and meander drawings. Various modalities were explored beyond handwriting in PD detection. In[34], electroencephalography (EEG) signals were employed, resulting in a 99.46% accuracy in classifying the control group, PD patients with medication, and PD patients without medication. A diverse set of CNNs was evaluated in[35], where deep CNNs using hand-drawn images outperformed other approaches, particularly ResNet50 and MobileNet-V2. Spiral drawings became a focal point in[36], where an SVM classifier and a SqueezeNet CNN as a feature extractor achieved an impressive accuracy of 91.26%, outperforming other models and classifiers. Tablet-based technologies were explored in[37], introducing the ParkinsonHW dataset and utilizing random forest classifiers. The interpretable features based on spiral shape properties resulted in an AUC of 0.999, emphasizing the value of straightforward models for enhancing interpretability. These studies collectively highlight the diverse approaches and modalities employed in PD detection, showcasing advancements that surpass previous benchmarks and underscore the potential of innovative technologies in improving diagnostic accuracy[6,35].

Apparently, recent studies have made significant strides in Parkinson's disease (PD) detection, emphasizing the potential of machine learning (ML) and innovative technologies for diagnosis and severity assessment. One notable contribution comes from[24], introducing a hybrid CNN/LSTM model that achieves high accuracy (85.5% and 99.3%) in diagnosing PD severity using handwritten tests. The study highlights the quantitative and comprehensive assessment of PD using deep learning technologies. Several studies explored diverse approaches and modalities. Fang[25] proposed an improved KNN algorithm for PD detection with a notable accuracy of 93.8%. Kuplan et al.[26] introduced an MRI-based classification model achieving impressive accuracies of 99.22%, 98.70%, and 99.53% for different PD-related conditions. Gazda et al.[16] presented an ensemble of deep-learning architectures achieving a remarkable accuracy of 96.3% for spiral drawing tasks. Additionally,[7] augmented hand-drawing datasets, demonstrating superior performance with a hybrid model composed of ResNet50 + SVM. Researchers also delved into neurocognitive features.[30] employed ML to evaluate sensor-based features, achieving 92.6% accuracy in discriminating PD from healthy controls. Furthermore, handwriting features were explored by[27], revealing that ML methods and digital drawing tests hold promise for automated neurodegenerative disease detection. Novel methods for early detection were suggested. An end-to-end CNN approach proposed by[28] achieved 94.7% accuracy in classifying PD from offline handwriting.[29] introduced a Continuous Convolution Network (CC-Net) for predicting tremor, shape, and spacing characteristics with 89.3% accuracy. Additionally,[31] applied ML to static and dynamic drawing tests, achieving an impressive accuracy of 95.32% in predicting PD. These studies collectively highlight the evolving landscape of PD detection, showcasing the effectiveness of ML across various modalities, from handwriting and neurocognitive features to MRI scans. The proposed methods demonstrate potential improvements in accuracy, accessibility, and early diagnosis, reflecting a promising direction for future research in the field[4,14].

### Problem statement

Attribute selection for Parkinson's Disease (PD) patients' handwriting analysis lacks clarity, hindering precise categorization based on hand-drawn spirals. Moreover, prevailing studies achieve commendable accuracy in PD detection, often exceeding 90%, yet their less transparent and somewhat opaque feature selection processes undermine their applicability in the medical realm. This study addresses the gap by not only identifying an optimal feature space for enhanced PD detection accuracy but also by conducting a comparative analysis of model complexities and prediction times against established methods. Additionally, the study recognizes the dearth of suitable datasets and aims to overcome this hurdle by introducing synthesized datasets for comprehensive training of machine learning and deep learning models.

### Motivation

Neglected in PD diagnosis, the feature selection process significantly influences model complexities and prediction times. This study addresses this gap by identifying a feature space for enhanced PD detection accuracy. It compares model complexities and prediction times with current methods, offering a transparent view of machine learning models. Synthesizing datasets addresses data scarcity, ensuring large, diverse training data for improved prediction accuracy with reduced memory and time requirements using ML models.

## Method

The study employs a four-fold approach to evaluate the models' performances based on the Parkinson's Image dataset (PID)[49], referred to as PID or $DS_0$. The PID consists of 72 training and 30 testing image samples of spiral and wave curves. The PID dataset consists of two classes: healthy individuals and those diagnosed with Parkinson's disease. The dataset is balanced, with an equal number of samples in each class: healthy individuals and those diagnosed with Parkinson's disease. This study only employs an image set of spiral curves and leaves out a wave dataset for future works. The reason for selecting spiral curves is that spirography is an established technique for assessing Parkinson's disease symptoms, including tremors and micrographia and examining how individuals make up for these impairments[50,51]. Researching micrographia and motor strategies in Parkinson's patients can be significantly aided by hand-drawn spirals. Studies reveal that during spiral drawing tasks, people with Parkinson's disease generally show less growth in spiral radius than healthy people, leading to narrower spirals[50]. Several neurological disorders have been evaluated by spirometry, including essential tremor[52], Parkinson's disease (PD) motor symptoms[50], multiple sclerosis (MS)[53], and Niemann-Pick disease[54]. An overview of these uses has been provided by a review[55]. Complete motor Unified Parkinson's Disease Rating Scale (UPDRS) scores validate that spiral severity, measured as the Degree of Severity (DoS), correlates with disease severity[50]. **Note** that the study employs ML models with default calibration settings available in the Python programming language.

The first fold employs the PID dataset referred to as $default_data_set$ or $DS_0$ containing original training images. The second fold includes data augmentation applied to $default_data_set$ or $DS_0$ to generate a new large balanced dataset, hereafter referred to as $augmented_default_dataset$ or $DS_2$. Next, all the ML models are trained on the datasets $DS_0$ and $DS_2$, and the ML model accuracies and memory footprints are observed for both datasets. Likewise, in the third fold, the $default_data_set$ or $DS_0$ is subject to a canny edge detector that results in a new dataset, hereafter referenced as $canny_data_set$ or $DS_1$. The fourth fold involves data augmentation to create balanced large-size datasets, hereafter referred to as $augmented_canny_dataset$ or $DS_3$ for training the models. Next, various ML models are trained using $DS_1$ and $DS_3$.

In all the folds, the **10-fold** cross-validation helps evaluate the performance of all ML classifiers. With stratified sampling, the dataset is randomly split while ensuring that each subset maintains an equal distribution of classes. The target variable (y) representing the class controls the sampling process. The classifiers are tested for accuracy using a reserved testing dataset or unknown sample set in the datasets $DS_0, DS_1, DS_2$, and $DS_3$. The study implements the following nine classifiers: Logistic Regression (LR) modeled class probabilities, Random Forest (RF) utilized ensemble learning with decision trees, and AdaBoost (AdB) combined weak classifiers for robust predictions. K Nearest Neighbor (KNN) relied on the majority class of neighbors, Decision Tree (DT) structured decisions in a tree-like manner, and Naive Bayes (NB) used Bayes' theorem for probabilistic classification. Support Vector Machines (SVM) facilitated supervised learning. Additionally, ensemble methods like Extreme Gradient Boosting (XGBoost) and Gradient Boosting Machine (GBM) optimized distributed gradient boosting for improved efficiency. This diverse set of classifiers provided a comprehensive exploration of both traditional and advanced machine learning techniques in the study's classification framework.

## Edge detection

*Histogram-of-oriented gradients*

The study employs a histogram of oriented gradients (HoG) to extract gradient information from images in the PID. HoG requires computing orientation gradients for each image area and normalising histograms in addition to separate groups of histogram plots. Let I(.) is an intensity function denoting the grey scale values of the image. The Eq. 1) and Eq. (2) enable computing of the horizontal gradient $G_x$ and vertical gradient $G_y$ of each pixel on grayscale spiral image $I(x, y)$, respectively:

$$G_x(x,y) = I(x+1,y) - I(x-1,y) \tag{1}$$

$$G_y(x,y) = I(x,y+1) - I(x,y-1) \tag{2}$$

The Eq. (3) and Eq. (4) express the magnitude of the gradient at a pixel and gradient orientation respectively,

$$G_y(x,y) = \sqrt{{G_x}^2 + {G_y}^2} \tag{3}$$

$$\theta(x,y) = \arctan(\frac{G_x}{G_y}) \tag{4}$$

Computing and counting the gradient direction histograms of local areas of the spiral image results in a HoG descriptor. Descriptors first divide the spiral image into small cells and then compute the gradient of the pixels in each cell. Next, the HoG creates a pixel block by integrating multiple cells.

*Canny edge detector*

The canny algorithm is the most popular edge detection technique with a high citation score ($\sim$ 41,786 total)[56]. The Canny edge detector stands out for its precise edge localization, low error rate, and ability to detect single-pixel width edges effectively. Its multi-stage algorithm includes noise reduction, gradient calculation, non-maximum suppression, and hysteresis thresholding, ensuring accurate edge detection while minimizing false positives. Gaussian smoothing reduces image noise, enhancing robustness, and parameter adjustability allows fine-tuning for optimal performance in various image processing applications. Apparently, Canny edge detection is highly reliable and accurate for detecting edges across different image conditions in computer vision tasks. Canny edge detection uses the calculus of variations to optimize a function that is a sum of four exponential terms approximating a Gaussian's first derivative. John Canny is the inventor of the potent computational method for edge identification known as Canny Edge identification[57]. The Canny edge detector provides enhanced edge detection and meets the three edge detection requirements[58]: (i) detection with a low error rate, (ii) the edge point should localize in the middle of the edge and; (iii) an edge should only be marked once, and picture noise should not form edges. The canny edge detector finds the edges of any image under analysis. The multi-step algorithm is based on the following steps: (1) uses a Gaussian filter to remove noise from the image; (2) calculates the gradient of the image pixels along the x- and y-directions; (3) applies non-maximum suppression to thin out edges; (4) applies double-threshold filtering to detect strong, weak, and non-relevant pixels; and (5) implements edge tracking by hysteresis to convert weaker pixels into stronger pixels if at least one of their neighbours is a stronger pixel.

The Canny edge detector identifies edges by comparing a pixel's gradient magnitude with its neighbors along the direction of maximum intensity change. Despite potentially missing some prominent edges[59], Canny outperforms comparable algorithms like Marr-Hildreth[60], which uses Gaussian smoothing and Laplacian of Gaussian filtering. Harris Corner Detection[61] identifies corners by analyzing gradient variations in local regions. While faint edge detection[62] methods offer robustness in diverse conditions, they may not align with our Parkinson's disease handwriting analysis focus on clear, well-defined edges crucial for diagnosis. These methods often require extensive noise reduction and preprocessing, complicating our workflow and increasing computational demands. Therefore, we chose the Canny edge detector for its precise edge localization, low error rate, and effectiveness in detecting single-pixel width edges, enhancing our ability to analyze fine details in handwriting samples. Let $\varphi(p,q)$ stand for the Gaussian function and $f(p, q)$ represent the input image:

$$\varphi(p,q) = e^{-\left(\frac{p^2+q^2}{2\sigma^2}\right)} \tag{5}$$

The convolution operation of $\varphi$ and $f$ given by Eq. (5) creates the smoothed image $f_s(p,q)$.

$$f_s(p,q) = \varphi(p,q) * f(p,q) \tag{6}$$

The sampling that generates a Gaussian mask and the scaling of coefficients that sum up to one generates a smoothed image represented by $f_s$. Fig. 1 depicts the representation of the Gaussian function $\varphi$ and a typical Gaussian mask.

Considering more about $3 \times 3$ patch in the smoothed image, $f_s(p,q)$, as shown in Fig. 2a, the gradient magnitude and direction (angle) can be estimated using Sobel operators, as depicted in Fig. 2b and Fig. 2c.

$$M(p,q) = \sqrt{\varphi_p^2 + \varphi_q^2}, \tag{7}$$

$$\alpha(p,q) = \tan^{-1}(\varphi_q/\varphi_p) \tag{8}$$

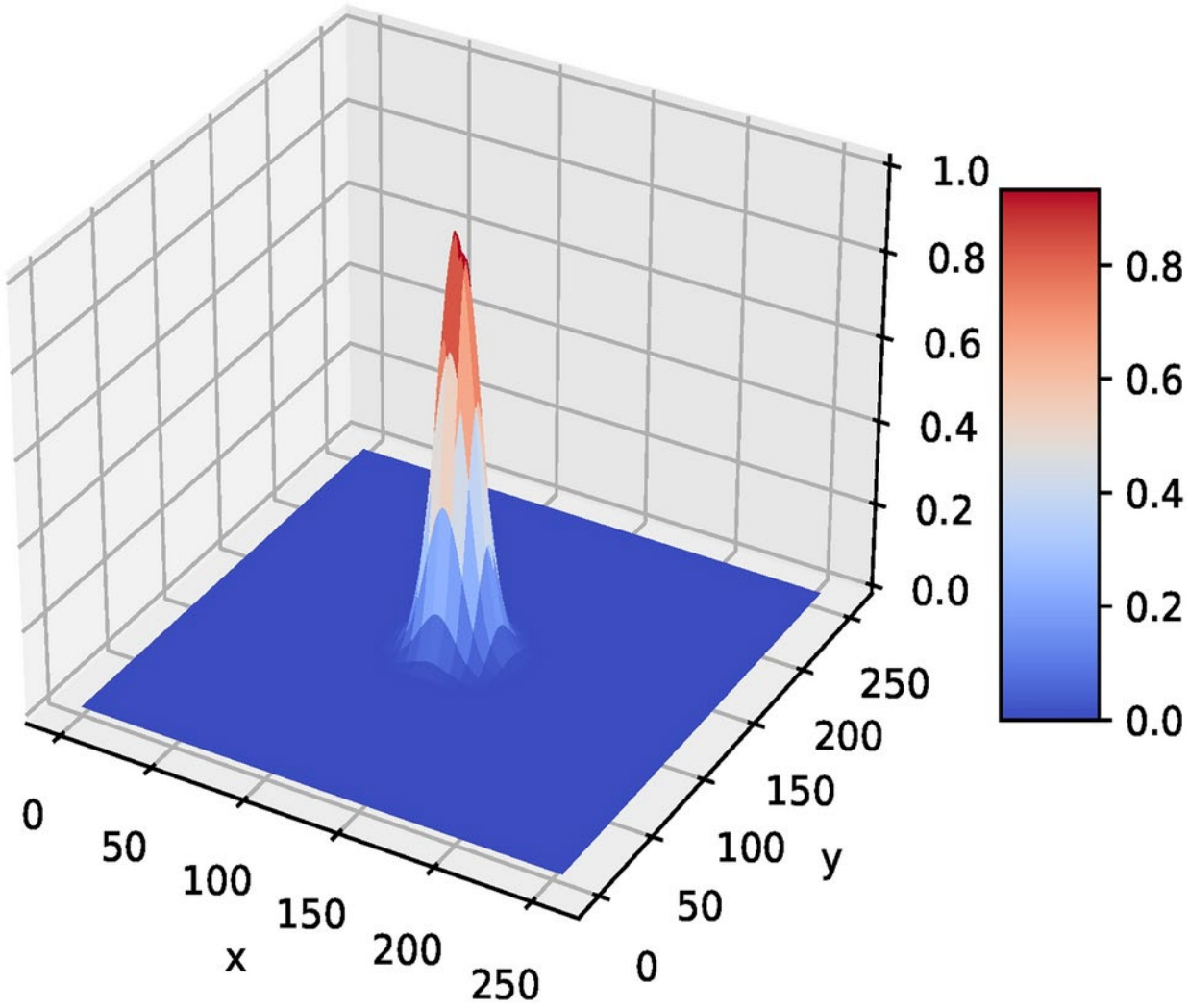

**Fig. 1**. A 3D plot of a Gaussian function $\varphi$ of smoothed image $f_s$.

| $s_1$ | $s_2$ | $s_3$ |
|---|---|---|
| $s_4$ | $s_5$ | $s_6$ |
| $s_7$ | $s_8$ | $s_9$ |

$(a)\ A\ \ 3\times 3\ region\ of\ f_s$

| $-1$ | $-2$ | $-1$ |
|---|---|---|
| 0 | 0 | 0 |
| 1 | 2 | 1 |

$(b)\ Sobel-p$

| $-1$ | 0 | 1 |
|---|---|---|
| $-2$ | 0 | 2 |
| $-1$ | 0 | 1 |

$(c)\ Sobel-q$

**Fig. 2**. A $3 \times 3$ region of smoothed image fs and two Sobel operators in p and q directions[60].

Where$\varphi_p^2 = \frac{\partial f_s}{\partial p} = (z_7 + 2z_8 + z_9) - (z_1 + 2z_2 + z_3)$and$\varphi_q^2 = \frac{\partial f_s}{\partial q} = (z_3 + 2z_6 + z_9) - (z_1 + 2z_4 + z_7)$

. Please note that the arrays $M(p, q)$ and $\alpha(p, q)$ are the same size as the image from which they were generated. $M(p, q)$ typically has large ridges around local maxima since it is produced using the gradient. The next stage is to thin out those ridges. Using non-maxima suppression is one strategy. There are numerous ways to accomplish this. However, the main strategy is to provide several distinct orientations of the edge normal (gradient vector).

For instance, in a $3 \times 3$ region, four orientations can be defined for an edge that passes through the region's centre point: horizontal, vertical, $+45^o$, and $-45^o$. The two alternative orientations of a horizontal edge are depicted in Fig. 3a. We must specify a range of edge directions we consider an edge horizontal because we must quantize all possible edge directions into four. The direction of the edge normal, which we explicitly extract from the image data, lets us estimate the edge direction Eq. (8). The edge is referred to as horizontal if the edge normal is in the range of directions from $22.5^o$ to $22.5^o$ or from $157.5^o$ to $157.5^o$, as shown in Fig. 3b. The angle ranges corresponding to the four considered orientations are depicted in Fig. 3c[63]. Let $\gamma_1$, $\gamma_2$, $\gamma_3$, and $\gamma_4$ represent the four basic edge directions for a $3 \times 3$ region that were previously discussed: horizontal, $45^o$, vertical, and $45^o$, respectively. The following non-maxima suppression strategy can be formulated for a $3 \times 3$ region centered at every point $(p, q)$ in $\alpha(p, q)$[63]:

- Locate the direction $\gamma_k$ that is most near $\alpha(p, q)$.
- If $M(p, q)$ is less than the sum of its two neighbors along $\gamma_k$, let $g_N(p, q) = 0$ (suppression); otherwise, let $g_N(p, q)$ equal $M(p, q)$, where $g_N(p, q)$ represents the non-maxima formulated image.

This study applies the canny edge detection algorithm to track down spiral edges in images from PID to constitute $DS_1$. The PID is subject to canny edge detection and then ridges in the images in the PID are filtered out using ridge operators.

*Edge intensity detection using Ridge operators*

Ridge operators in image processing enable the detection and enhancement of ridges or elongated structures in digital images. They help highlight prominent linear features, such as edges, vessel structures in medical images, or lines in fingerprint images. Ridge operators apply filters or convolution kernels to the image, which allows them to emphasize ridges and suppress other image features[64]. The techniques find applications in computer

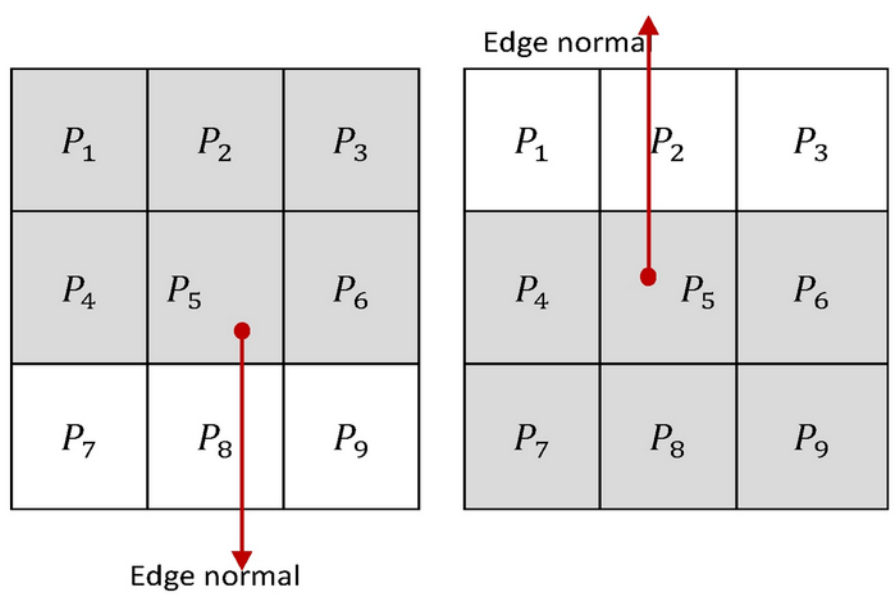

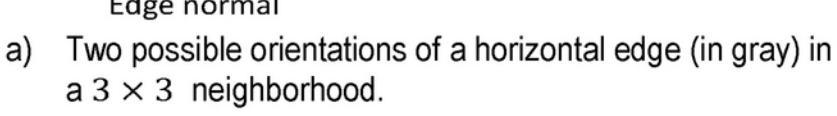
a) Two possible orientations of a horizontal edge (in gray) in a $3 \times 3$ neighborhood.

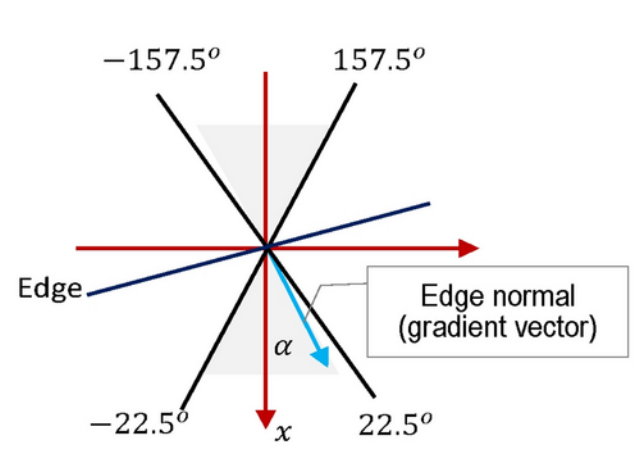

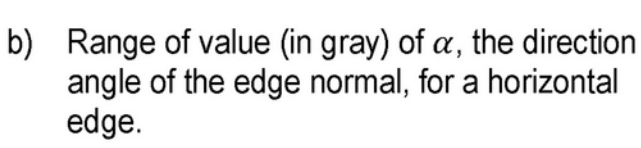
b) Range of value (in gray) of $\alpha$, the direction angle of the edge normal, for a horizontal edge.

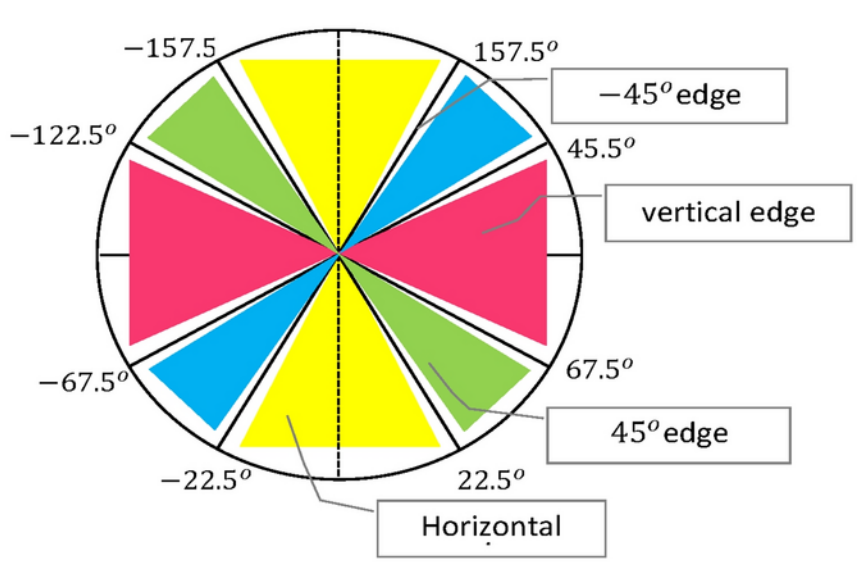

c) The four types of edge directions in a $3 \times 3$ neighborhood's edge normal angle ranges. Two ranges are specified with the same color for each edge direction.

**Fig. 3**. Angle range matching to edge normal and the four edge directions of the Canny edge detector[60].

vision, pattern recognition, and image analysis tasks where detecting and extracting ridge-like structures is crucial. Typical examples of ridge operators include the Ridgelet Transform, Hessian-based ridge detection, and the Radon Transform. They play a significant role in feature extraction, image segmentation, and object recognition tasks. The study implements a hessian-based ridge detection method to track ridges in the spiral curves[64]. The Hessian matrix helps detect and analyze special shapes, including simultaneous segmentation and reconstruction of curvilinear structures in medical images[65]. The hessian filter considers an image $I(x, y)$ as a three-dimensional (3D) surface, as shown in Fig. 1. The 3-dimensional surface curvature represents grayscale changes in the two-dimensional (2D) image. Hessian matrix $H(x, y)$ defines the curvature by the expression Eq. (9):

$$H(x,y) = \begin{bmatrix} I_{xx}(x,y) & I_{xy}(x,y) \\ I_{xx}(y,x) & I_{yy}(x,y) \end{bmatrix} \tag{9}$$

where $I_{xx}(x,y)$, $I_{xy}(x,y)$, $I_{yx}(x,y)$, $I_{yy}(x,y)$ are the four second partial derivatives of the original 2D image $I(x, y)$. The the second partial derivative in the x and y direction are expressed by Eqs. (10) and (11) respectively:

$$I_{xx} = \frac{\partial^2 I}{\partial x^2} = I(x-1,y) + I(x+1,y) - 2I(x,y) \tag{10}$$

$$I_{yy} = \frac{\partial^2 I}{\partial y^2} = I(x,y-1) + I(x,y+1) - 2I(x,y) \tag{11}$$

Similarly, Eq. (12) expresses the second mixed partial derivative in the x and y directions:

$$I_{xy} = \frac{\partial^2 I}{\partial x^2}y = I(x+1,y+1) + I(x,y) - I(x+1,y) - I(x,y+1) \tag{12}$$

$1(x, y)$ and $2(x, y)$ represent two eigenvalues of $H(x, y)$ since $I_{xy}$ equals $I_{yx}$ and $H(x, y)$ a real symmetric matrix. The maximum and minimum eigenvalues correspond to the intensity of the maximum and minimum curvature of the 3D surface, respectively.

The intensity of the minimum curvature. The edge intensity of each point is expressed by Eq. (13), representing the two eigenvalues computed by the Hessian matrix:

$$\lambda_1 = K + \sqrt{K^2 - Q^2}, \lambda_2 = K - \sqrt{K^2 - Q^2} \tag{13}$$

where $K = \frac{I_{xx}+I_{yy}}{2}$ and $Q = \sqrt{I_{xx}I_{yy} - I_{xy}I_{yx}}$. $I$ of each pixel ranges from 0 to 255 since all images are converted to a grey scale before being subject to hessian computation. In addition, the brightness is a function $I : Z \times Z \rightarrow Z0$, as the coordinates of the pixels on every image vary from 0 to a positive integer.

## Experiment setup

The testbed for the study is set up using a notebook with an Intel® Core$^{TM}$ i3-370M 2.40 GHz processor (with a 3*MB* cache) and 8*GB* of RAM. The program is developed using Pycharm programming integrated development environment employing Python version 3.0. The logical layout for the experiment is shown in Fig. 4. The experiment consists of a sub-list of tasks: pre-processing and feature extraction, feature selection, training and evaluation, testing, estimating memory footprint and model prediction time.

## Data set selection

The PID consists of two datasets: spiral and wave datasets[49]. Each type consists of a training and testing dataset with 72 and 30 images, respectively. The actual split in the PID consists of 204 images and is pre-divided into a training set and a testing set, which includes: a) Spiral: 102 images, 72 training, and 30 testing and b) Wave: 102 images, 72 training images, and 30 testing images. There are 36 images in each of the healthy and Parkinson's classes in the training set, while the test set contains 15 samples for each class. All images in the PID are in the CMGY format. The study employs a spiral curve dataset only for the experiment described.

**Fig. 4**. Logical layout of the experiment for classifying images of healthy and PD patients using spiral image dataset.

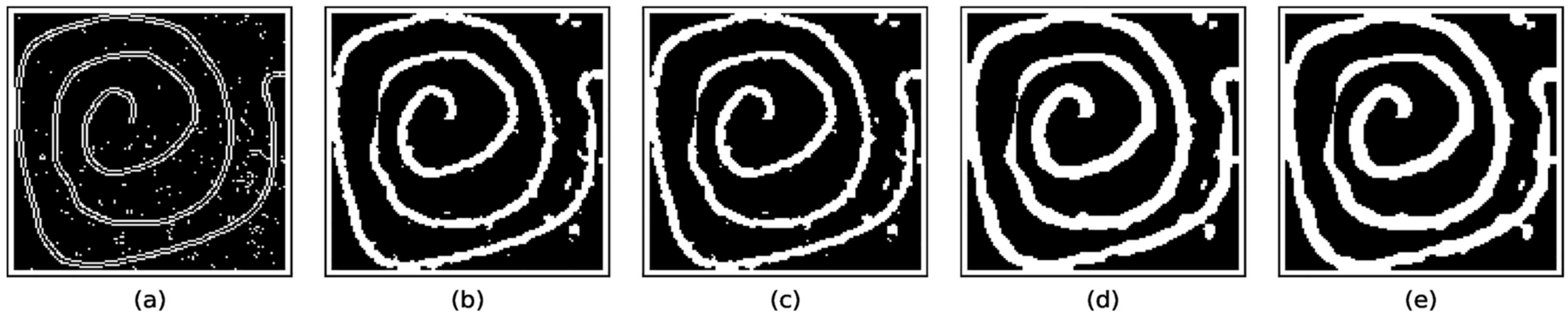

**Fig. 5**. Canny detector and sigma parameter value settings. **a**) Image acquired by applying canny edge detection; **b**) sigma = 1; **c**) sigma = 2; **d**) sigma = 2.8; **e**) sigma = 3. The value of sigma at index (**d**) represents the optimal value found during the pre-processing phase for training ML models on $DS_1$ and $DS_3$. $threshold1 = threshold2 = 51$, and $apertureSize = 5$.

## Data augmentation and pre-processing

Pre-processing resulted in the generation of four datasets ($DS_0$, $DS_1$, $DS_2$, and $DS_3$). Data augmentation techniques were employed to enhance the robustness of the dataset. These included rotation by up to 360 degrees, slight shifts in width and height (0.01 range), and horizontal and vertical flips. The brightness range was maintained at its default setting, as specified in the ImageDataGenerator configuration. These transformations were implemented to increase variability in the dataset, ensuring the model's ability to generalize effectively. Algorithm 1 provide the pseudocode used to process images for data augmentation.

The datasets were generated using the following steps: a) First, the raw images are extracted from the data source; b) Extracted images are corrected for the image dimensions ($image_index$, 128,128, channel). Here, $image_index$ is the index of the image in the dataset, and the channel represents three red, blue and green colour image channels; c) Next, the raw images from the data source are passed as a direct input to the algorithm computing HoG images or canny edge detection algorithm. Raw images are subject to Hessian filtering in the case of the canny image generation process (see algorithm 2). Different *sigma*, *threshold*1, *threshold*2, and *apertureSize* values were tried to find optimal presented in Fig. 5. The Hessian filter receives dataset filtered through canny edge detector algorithm and outputs Hessian filtered images as the dataset ($DS_1$); d) For generating augmented datasets, the raw images are first input to the augmentation algorithm generating augmenting datasets that serve as input either to the HoG images processing stage or the canny image processing stage. Figure 6 shows the samples of augmented dataset $DS_2$; e) Raw images from the data source as direct input to the HoG or canny algorithm generate the two datasets $DS_0$ and $DS_1$, respectively. The same raw images as input to the data augmentation algorithm generate large datasets or augmented datasets are inputs to the HoG and canny algorithms for generating datasets $DS_2$ and $DS_3$. The augmented dataset contains 2556 training samples of healthy and Parkinson's; hence a total of 5112 training samples. We do not employ test dataset for evaluation since the studies have confirmed that 10-fold cross validation is reliable metrics for evaluation of ML models (refer to the performance evaluation section).

The key difference between the original and Canny-processed datasets lies in the preprocessing applied to the images. The Canny edge detection algorithm detects edges within the spiral drawings, and Hessian filtering is then used to sharpen the image by emphasizing the ridges. This preprocessing enhances the structural details, making the features more defined. In contrast, the original dataset remains intact, preserving all the raw image data without any enhancement or noise reduction. The Canny-processed dataset provides a modified version of the images that highlights specific features relevant for analysis, while the original dataset retains its full spectrum of features, including both noise and structural elements. These different datasets contribute to the analysis by enabling a comparison of model performance using original and edge-detected features.

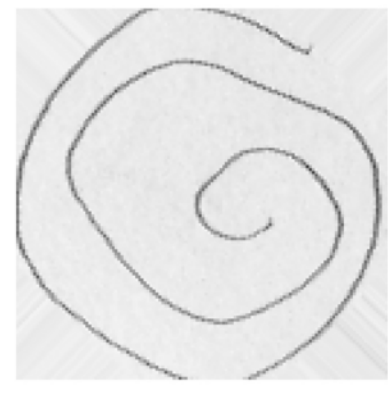
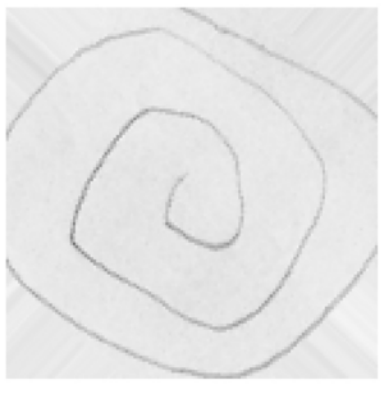
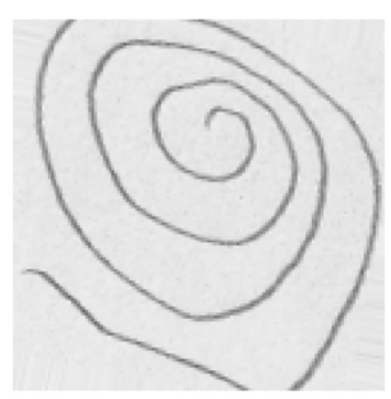

(a) Healthy

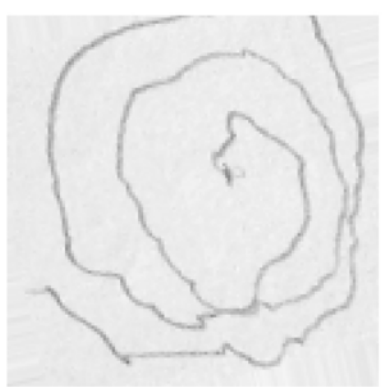
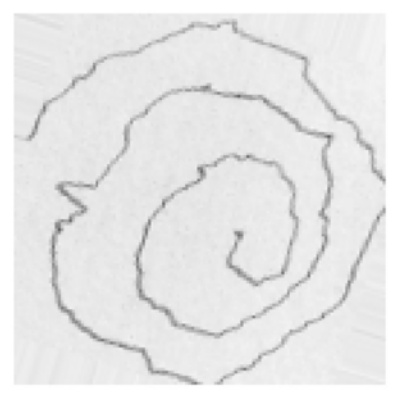
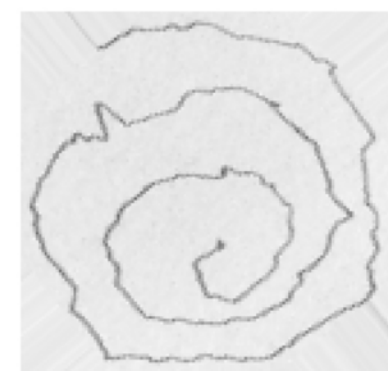

(b) Parkinson's

**Fig. 6**. Samples of real spiral images from augmented dataset ($DS_2$) representing hand-drawn spirals of healthy and Parkinson's classes.

1: **Input:** $x_{train}, y_{train}$ (Original training datasets)
2: **Output:** $x^{aug}_{train}, y^{aug}_{train}$ (Augmented datasets for both classes: Healthy and Parkinson's)
3: **Step 1: Initialize Augmentation Generators:**
4: Use `ImageDataGenerator` with parameters like rotation, shift, flip (horizontal and vertical), and brightness adjustment (default).
5: **Step 2: Augment Training Data:**
6: For each image in $x_{train}$:
7: Apply augmentation to generate 71 augmented images per original image (71x72=5112 image samples).
8: If the image belongs to the *Healthy* class, store augmented images in the Healthy class folder.
9: If the image belongs to the *Parkinson's* class, store augmented images in the Parkinson's class folder.
10: **Step 3: Resize Images:**
11: Resize all augmented images to $128 \times 128$ pixels.
12: **Step 4: Encode Labels:**
13: Apply `LabelEncoder` to encode the labels into numeric format, assigning one numeric label for *Healthy* and another for *Parkinson's*.
14: **Step 5: Return Augmented Data:**
15: Return $x^{aug}_{train}, y^{aug}_{train}$, where $x^{aug}_{train}$ contains augmented images stored in separate folders for *Healthy* and *Parkinson's*, and $y^{aug}_{train}$ contains the corresponding labels.

**Algorithm 1**. Data augmentation.

*Feature extraction*
This study's novelty relies on extracting the low-level features from the four datasets. The study employs manual feature extraction methods implementing statistical attributes extracted from images from each prepared dataset. The study employs a sequential feature extraction process while considering the two different approaches of edge detection: a) Histogram-of-oriented gradients (HoG) and b) Canny edge detection. As explained in the prior subsection, the first approach prepares HoG images by subjecting images in $DS_0$ to the HoG transform while the second approach implements a canny edge detector with a hessian filter to generate canny images for the $canny_data_set$ or $DS_1$. The datasets $DS_0$ and $DS_1$ input to the data augmentation algorithm generate the augmented datasets $DS_2$ and $DS_3$. The handcrafted feature extraction process required computing eleven statistical features vectors – **mean, standard deviation, skewness, kurtosis, energy, power, median, variance, minimum, maximum, and rms**; all annotated by numerals 0 through 10, respectively and estimated using Python packages – *numpy* and *scipy.stats*. As such, the feature space for training ML models can be represented by $F = [0, 1, 2, 3, 4, 5, 6, 7, 8, 9, 10]$. The Recursive Feature Selection (RFS) algorithm enabled finding the best possible features to generate a feature set, such as $feature_set_DS_0 = [2, 3, 4, 6, 9]$ from the realized feature space. The study uses the same convention throughout this article.

1: **Input:** $X_{train}^{aug}$ (Augmented Image Dataset)
2: **Output:** $X_{train}^{processed}$ (Processed Dataset with Edge Detection and Hessian Filtering)
3: **Step 1: Initialize Parameters:**
4: Set edge detection thresholds (`threshold1`, `threshold2`), `apertureSize`, `sigma`.
5: **Step 2: Apply Canny Edge Detection on Augmented Training Data:**
6: For each image in $X_{train}^{aug}$:
7: Apply Canny edge detection with specified parameters.
8: Perform Hessian filtering on the edges.
9: Store the processed image in $X_{train}^{processed}$.
10: **Step 3: Return Processed Data:**
11: Return $X_{train}^{processed}$.

**Algorithm 2**. Canny edge detection with hessian filtering.

The selection of the eleven statistical feature was specifically tailored to extract meaningful insights from spiral images in the dataset. Spiral patterns provide unique structural and textural characteristics indicative of fine motor skills, often impaired in Parkinson's disease. Features such as mean and median capture the overall intensity distribution within the spirals, providing a baseline for comparing normal and pathological patterns. Standard deviation and variance highlight the variability in stroke pressure and consistency, often more pronounced in Parkinson's spirals due to tremors. Higher-order statistics like skewness and kurtosis detect asymmetry and peak sharpness in the patterns, reflecting the irregularities typical of Parkinson's handwriting. Energy, power, and rms are texture-oriented features that quantify the concentration and contrast within the spiral strokes, emphasizing differences in drawing control and force. Minimum and maximum values define the intensity range, capturing boundary variations in the strokes that can signal tremor severity or rigidity. These features were chosen because they effectively summarize the structural irregularities and dynamic variations inherent in spiral drawings, making them highly relevant for distinguishing Healthy versus Parkinson's cases. Their selection is grounded in both their proven applicability to similar medical image analyses and their computational efficiency for robust feature extraction.

*Feature selection*

The Recursive Feature Elimination (RFE) process is a feature selection technique that recursively removes features from the dataset and builds a model to evaluate the performance after each elimination. Initially, the algorithm fits a model using all available features and ranks them based on their importance. Features with the highest importance are kept, while those with the lowest importance (i.e., the lowest ranking value) are eliminated. This method ensures that the final feature set is efficient and effective for training the model. The process is repeated until the desired number of features is selected. The RFE function in Python programming language ranks features inversely, where lower-ranked features are deemed more important, retaining features with lower rank values. In our implementation as shown in algorithm 3, the optimal feature set was determined by selecting features ranked lower in the RFE process, indicating higher importance for the model. Specifically, a threshold rank value of 2 was set, and features with a rank value of 2 or lower were retained, while others were discarded. This threshold was chosen based on balancing model complexity and performance, ensuring that only the most relevant features were used, reducing the risk of overfitting. RFE is chosen over Principal Component Analysis (PCA) because, while PCA effectively reduces dimensionality and removes redundant features, it can obscure the interpretability of the original features. PCA transforms the data into a set of orthogonal components, which may not correspond directly to the original features, making it harder to interpret the model's behaviour regarding the actual variables. Additionally, PCA assumes linear relationships between features, which may not capture complex, non-linear relationships present in the data. On the other hand, RFE directly evaluates feature importance based on model performance, allowing us to retain features that contribute most to predictive power while preserving their interpretability and relevance to the task at hand.

1: Load training data $(X_{\text{train}}, y_{\text{train}})$. // the original[6] or augmented data.
2: Extract HoG features from $X_{\text{train}}$ and normalize using Min-Max normalization.
3: **for** each model in predefined models **do**
4: Apply RFE with cross-validation to rank features based on importance.
5: Inverse ranking behavior: lower rank means higher feature importance in Python's RFE.
6: Retain low-ranked features (higher importance), eliminate high-ranked features (lower importance).
7: Store feature rankings and selected features.
8: **end for**
9: Set threshold to select features with rank $\leq 2$.
10: Threshold selection ensures that only the most important features are retained, minimizing overfitting and improving model performance.
11: Save selected features to CSV.

**Algorithm 3**. Feature ranking using RFE.

The study implemented the RFS algorithm for selecting the best possible $feature_sets$ out of statistical features as input to the RFS algorithm for training the models. RFS ranked features with scores providing a subset of features; however, the observation here is that different models require different $feature_sets$ for achieving high prediction accuracy. Intuitively, the need to find a $feature_set$ enabling all models to have almost the same prediction accuracy was observed. The study presents a method depicted in algorithm 4 that combines RFS determined optimal feature vectors for creating the $feature_set$. Observations show that the feature vectors in $feature_sets$ individually determined for models by the RFS showcase common feature vectors. Comprising $feature_set$ of feature vectors common to all individually determined feature sets for different models serves as the best $feature_set$, generalizing model prediction accuracies across the range of models used in the study. However, adding other feature vectors common to some models, although only in some, improves the corresponding models' accuracy. In addition, the observations show that only the most common features for small datasets enable models to achieve prediction accuracies higher than the accuracies resulting from the large dataset.

1: **Initialize** datasets $(X_{\text{train}}, y_{\text{train}})$ // this could be original or augmented data.
2: **Extract** features using feature extractor (HoG)
3: **Normalize** feature vector using Min-Max normalization
4: **Define** set of models (e.g., Logistic Regression, Decision Tree, Random Forest, etc.)
5: **for** each model **do**
6: Fit model to training data $(X_{\text{train}}, y_{\text{train}})$
7: Compute permutation importance to rank features based on model performance
8: Store ranking scores for each feature
9: **end for**
10: **Rank features** for each model:
11: Combine features and ranking scores
12: Sort features by ranking scores
13: Identify the best (lowest-ranked) and worst (highest-ranked) features for each model
14: **Store results** in DataFrame:
15: List best and worst features for each model
16: **Output** ranking of features and associated best and worst features for each model
17: **Optional:** Generate feature combinations using itertools for further analysis

**Algorithm 4**. Feature space generator.

*Model training and performance evaluation*
A review of the literature reveals that a 10-fold cross-validation improves ML model performance; therefore, the study evaluates classifiers' performances using the 10-fold stratified cross-validation technique[6,34,36,42,43,47]. All of the classifiers are evaluated using stratified $k-fold$ cross-validation. By randomly dividing the data into $k$ equal-sized groups or folds while maintaining the same number of samples per class, the method generates $k-1$ training and a test group[66]. The models are trained/tested ten times in each cross-validation iteration. The model training is based on $feature_set$ specific to each model.

As mentioned in the previous section, the study employs $10-fold$ StratifiedKFold cross-validation to validate the trained models. Nine models are trained in four folds on four different datasets $DS_0, DS_1, DS_2$ and $DS_3$. All nine models employ default calibration settings provided by the Python programming language.

## Performance evaluation metrics

Machine-learning algorithms are evaluated using a variety of metrics. The study employs the following performance metrics to evaluate the performances of employed ML classifiers. The metrics used in the study are accuracy, Receiver Operating Characteristic Curve (RoC) curves, F-score, and Area Under the Curve (AUC) confusion matrix. The bias-variance plots showcase the complexities of models on the number of features employed for training the models. In addition, the study also reports the memory footprint and prediction time for each of the employed models.

*Accuracy (ACC)*
Accuracy (ACC) is given by the Eq. (14) and defines the proportion of correct prediction of a given condition[67,68].

$$ACC = \frac{TP + TN}{TP + TN + FP + FN} \tag{14}$$

Where *TP* represents the number of true positives, *TN* represents the number of true negatives, *FP* represents the number of false positives, and *FN* represents the number of false negatives.

*Area under curve (AUC)*
A commonly used performance evaluation approach that applies to binary classification problems. A classifier's AUC estimates the probability score, defining that the model under evaluation would rank a randomly chosen positive rather than a randomly chosen negative example. Equation (15) provides the AUC, or area under the curve[69]. The model performs better at differentiating between people with and without tremor symptoms the higher the AUC is in a range from 0 to 1.

$$AUC = \int_0^1 RoC(t) \cdot dt \tag{15}$$

Estimation of AUC requires the definition of the following two parameters:

- True Positive Rate (Sensitivity or TPR): TPR is the percentage of positive data points that are accurately interpreted as positive compared to all positive data points[29,68–70]. TRP is expressed by the equation eq(16).

$$TPR = \frac{TP}{TP + FP} \tag{16}$$

- True Negative Rate (Specificity or TNR): TNR is the percentage of positive data points that are accurately interpreted as negative compared to all negative data points[29,68–70]. TRP is expressed by the equation eq(17).

$$TNR = \frac{TN}{TN + FP} \tag{17}$$

Further, the AUC is computed as follows:

$$AUC = \frac{1 + TPR - FPR}{2} \tag{18}$$

*F1 score*
The harmonic mean of the precision and recall gives the formula for the conventional F1 score expressed by the Eq. (19). F1-score. An F-score of 1 indicates a perfect model[29,68–70].

$$F1 = \frac{TP}{TP + \frac{1}{2}(FP + FN)} \tag{19}$$

| AUC Value | Connotation |
|---|---|
| 0.9 < AUC < 1.0 | Excellent |
| 0.8 < AUC < 0.9 | Good |
| 0.7 < AUC < 0.8 | Fair |
| 0.6 < AUC < 0.7 | Poor |
| 0.5 < AUC < 0.6 | Insignificant |

**Table 1.** Typical range of AUC values and their interpretation[71].

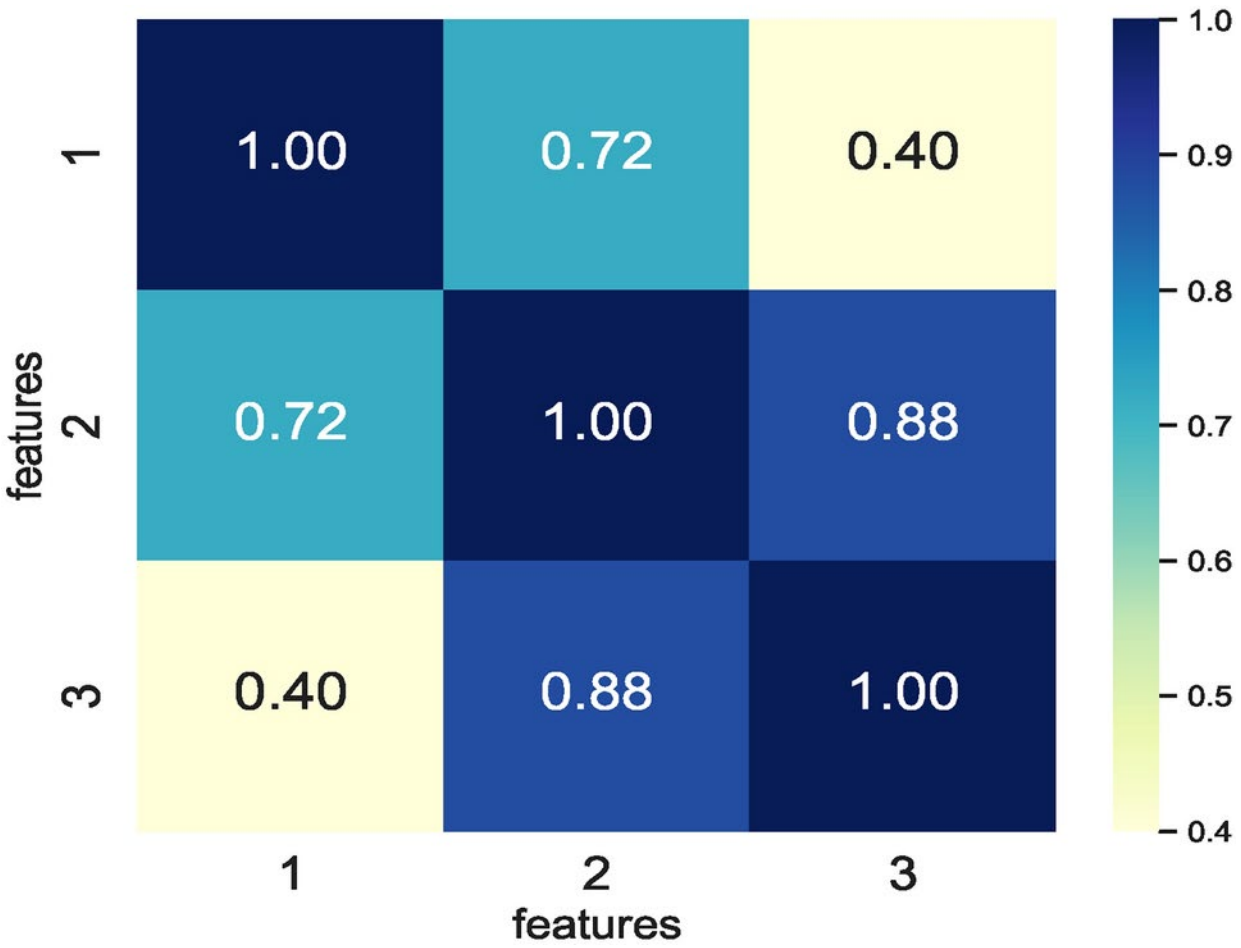

**Fig. 7**. Correlation matrix defining correlation of feature combinations handcrafted from $DS_0$.

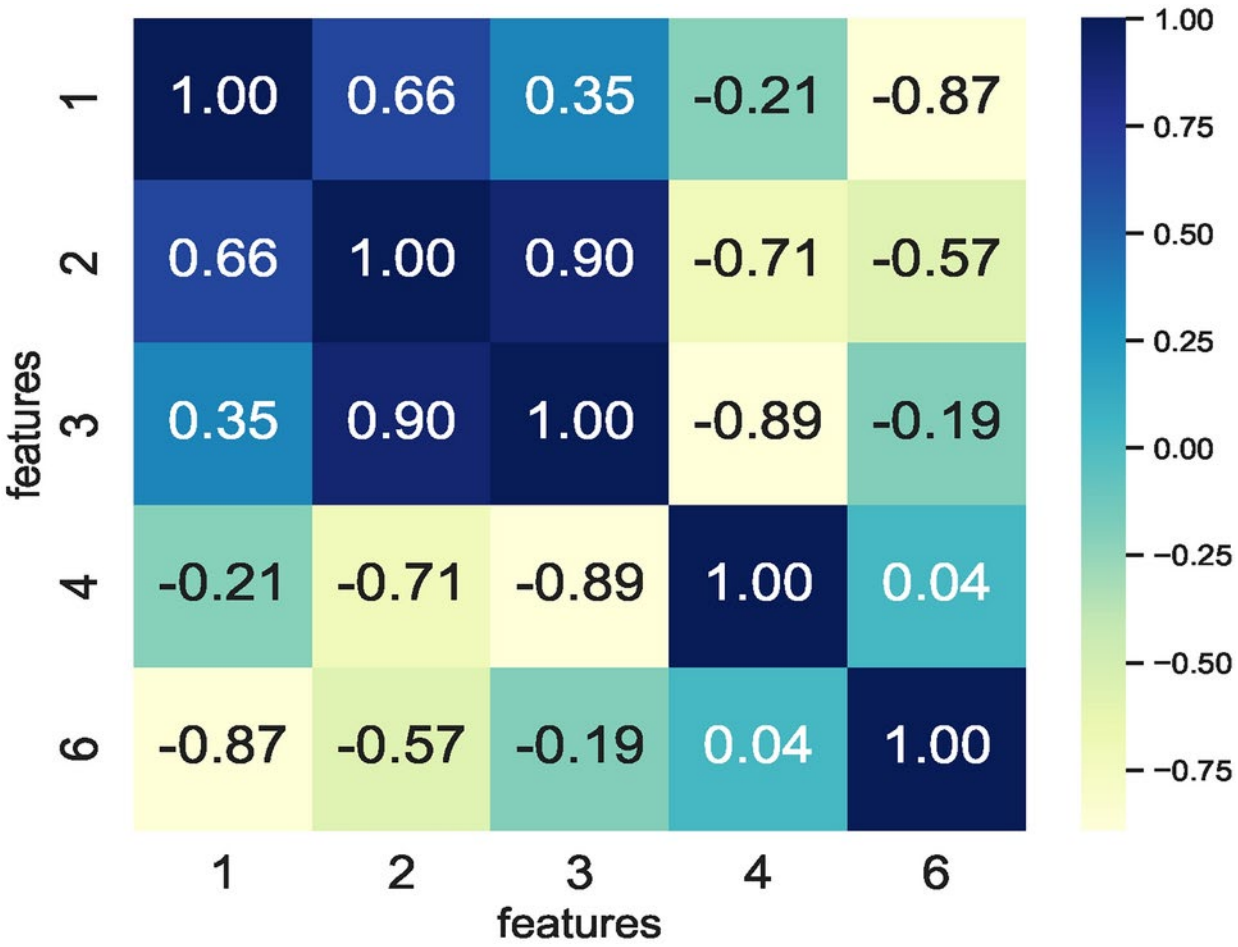

**Fig. 8**. Correlation matrix defining correlation of feature combinations handcrafted from $DS_2$.

*Receiver Operating Characteristic curve (RoC) Plot*
A RoC curve is a graph that depicts the performance of a classification model across all categorization levels. The curve compares the TPR and False Positive Rate (FPR) expressed by the Eqs. (16) and (20), respectively. The FPR is the percentage of negative data points incorrectly interpreted as positive compared to all negative data points.

$$FPR = \frac{FP}{TN + FP} \tag{20}$$

The TPR and the FPR values range from [0, 1]. Estimation of FPR and TPR at distinct threshold values, such as (0.00, 0.02, 0.04, ..., 1.00) can be plotted graphically, whereas the AUC defines the area under the curve of a plot of FPR vs TRP at different points in the range [0, 1]. Table 1 shows the typical values of AUC and their corresponding interpretation.

## Results and discussion

Training the models using a fourfold approach implementing four different datasets generates four different types of outputs that showcase performance at an individual level specific to every model. The next subsections showcase the experiment results and provide a detailed discussion.

### ML model training on $DS_0$ and $DS_2$

The correlation matrix formulated using a combination of various extracted features enabled identifying the correlation between the different feature vectors. The correlation matrix shown in Fig. 7 represents three feature vectors only, and a combination of these three feature vectors represents the feature space seen as input by the nine ML models. The RFS method removed the highly correlating features while generating feature sets specific to each model. However, a feature set enabling similar prediction accuracies by all nine models is selected. Therefore, feature set $feature_set_DS_0 = [1, 2, 3]$ represents the best possible feature set among all the RFS-generated features selected for training the models. In addition, eliminating the features 4 and 6 reduced the feature space for the small sized dataset $DS_0$.

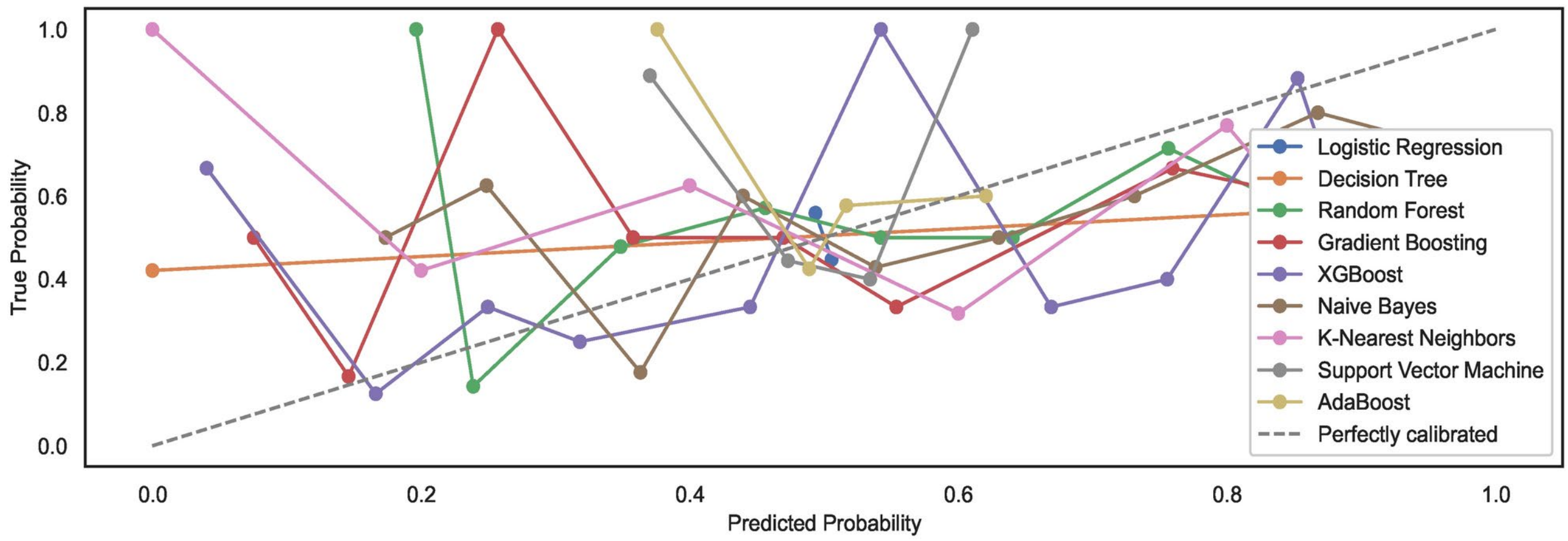

**Fig. 9**. Calibration curves for models trained on $DS_0$.

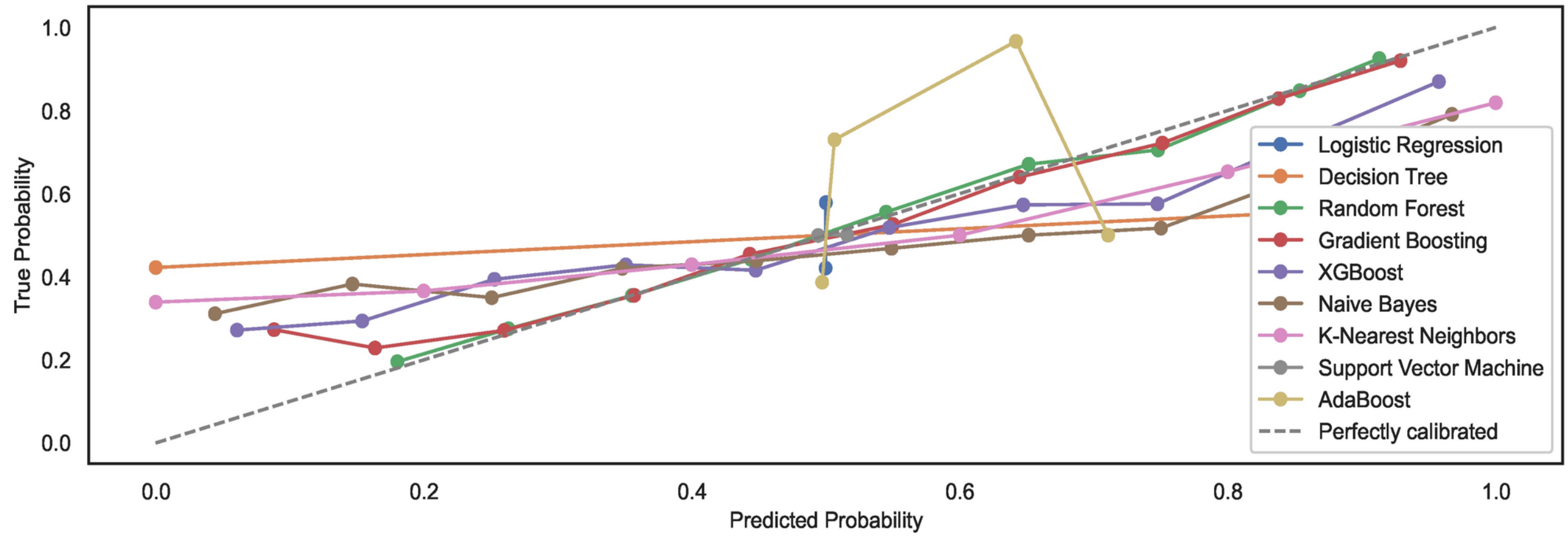

**Fig. 10**. Calibration curves for models trained on $DS_2$.

The correlation matrix in Fig. 8 enabled finding the feature space for the augmented dataset $DS_2$. Removing additional features other than those removed by the RFS results in lower prediction accuracies by the models. As such, the RFS generated features space comprising $feature_set_DS_2 = [1, 2, 3, 4, 6]$ enabled models' training and ensured similar prediction accuracies. While some models show the same behaviour on the reduced feature space, a decrease in memory footprint and prediction times of the ML models trained is observed (see Figs. 19 and 20).

*Model calibration*

As mentioned in the prior section, the study employs un-calibrated models with default configuration settings. Expected model behaviour showcases a non-alignment with the perfect calibration line when models are trained on $DS_0$. However, observations show an alignment of generated calibration curves with the perfect curve for models trained on a large-size dataset $DS_2$. The results reveal that large-size datasets automatically calibrate the models to some extent; however, they remain non-aligned to the ideal perfect calibration model state curve. Figures 9 and 10 show the calibration curves for the models trained on small and large datasets whilst considering the feature space size, respectively. The models trained on large datasets showcase curves nearly aligning with the perfect model calibration line although not exhibiting perfect $S-shape$ curves that typically result with calibrated models. The study highlights that calibrating models before training may improve the prediction accuracies of the models further. Nonetheless, the task has been left over for future work.

*Model training*

Figure 11 shows the cross-validation accuracy and F1-score for the nine models evaluated using stratified $10-fold$ cross-validation.

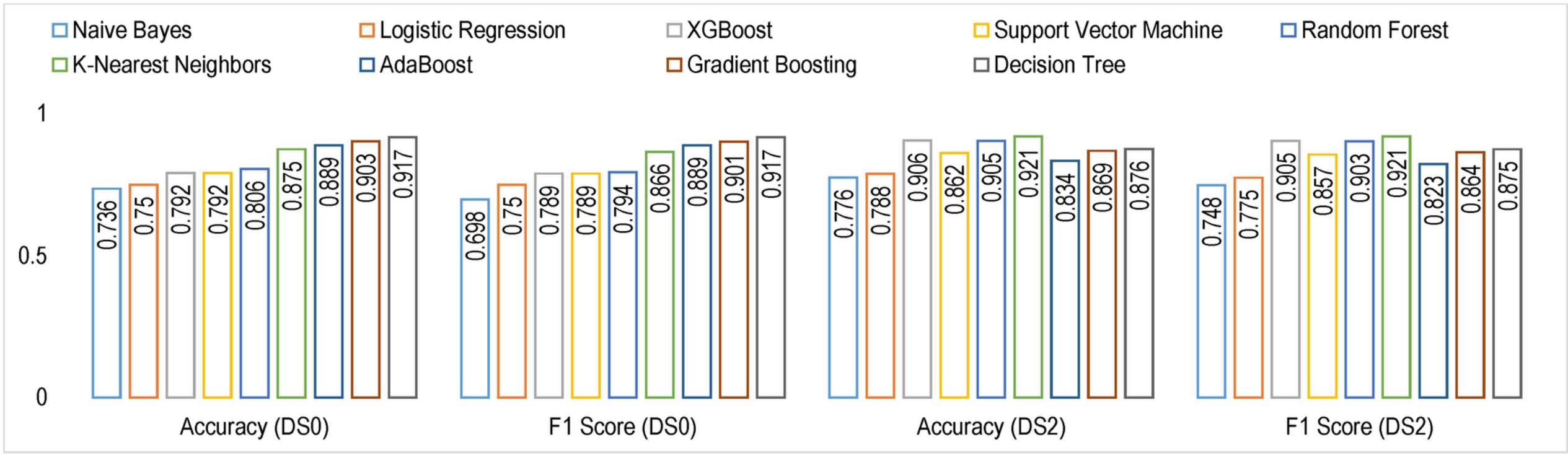

**Fig. 11**. Results of model evaluation results on $10-fold$ cross validation.

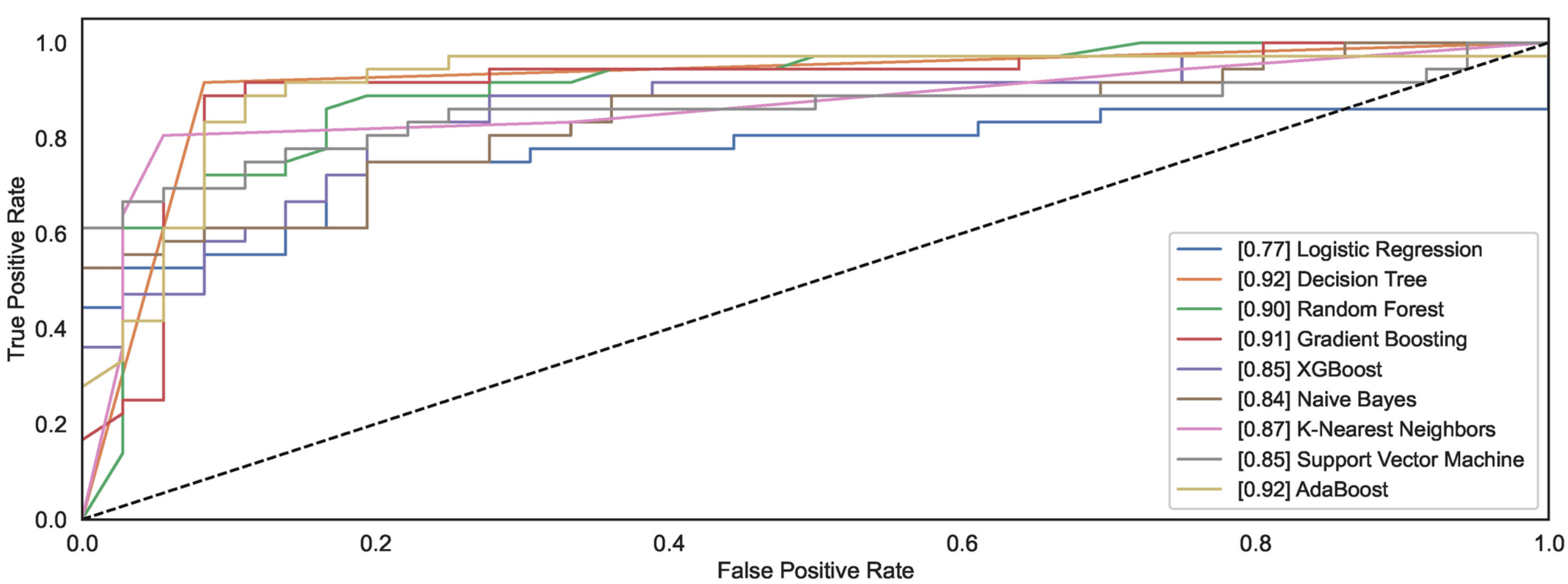

**Fig. 12**. ROC curves for models trained on $DS_0$.

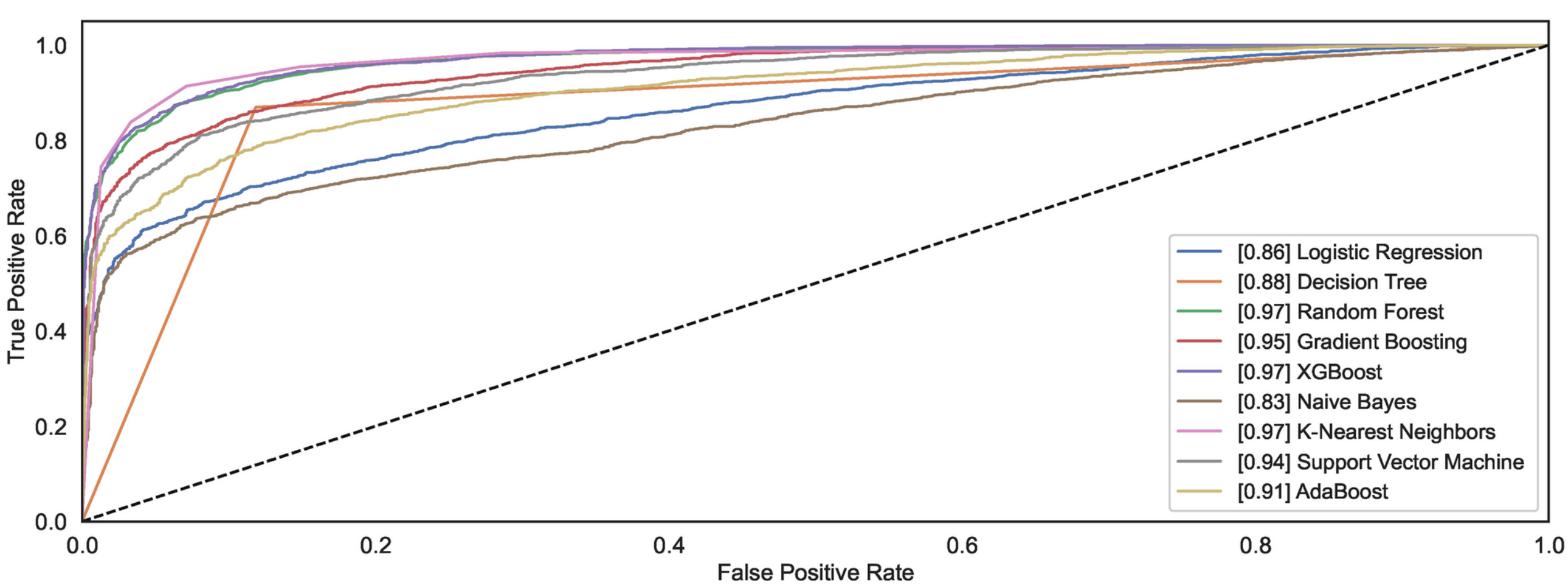

**Fig. 13**. ROC curves for models trained on $DS_2$.

## Performance evaluation on $DS_0$ and $DS_2$

*Model ROC*

The ROC curves demonstrate how different models behave towards the task of identifying the handwritten spiral drawings of healthy and PD patients. Figure 12 shows the ROC plots for the models trained on $DS_0$. Observing the results reveals that the models DT and ADB classify spiral drawing of healthy and PD patients with an AUC of 0.92; the highest AUC compared to other AUC of other classifiers. RF and GB show the AUC values of 0.90 and 0.91, respectively, thereby demonstrating equal competitiveness with the DT and ADB that attain the highest value of AUC among all the classifiers. While the other models reveal AUC values ranging from 0.85 to 0.87, the LR model remains the worst classifier to distinguish between the two classes.

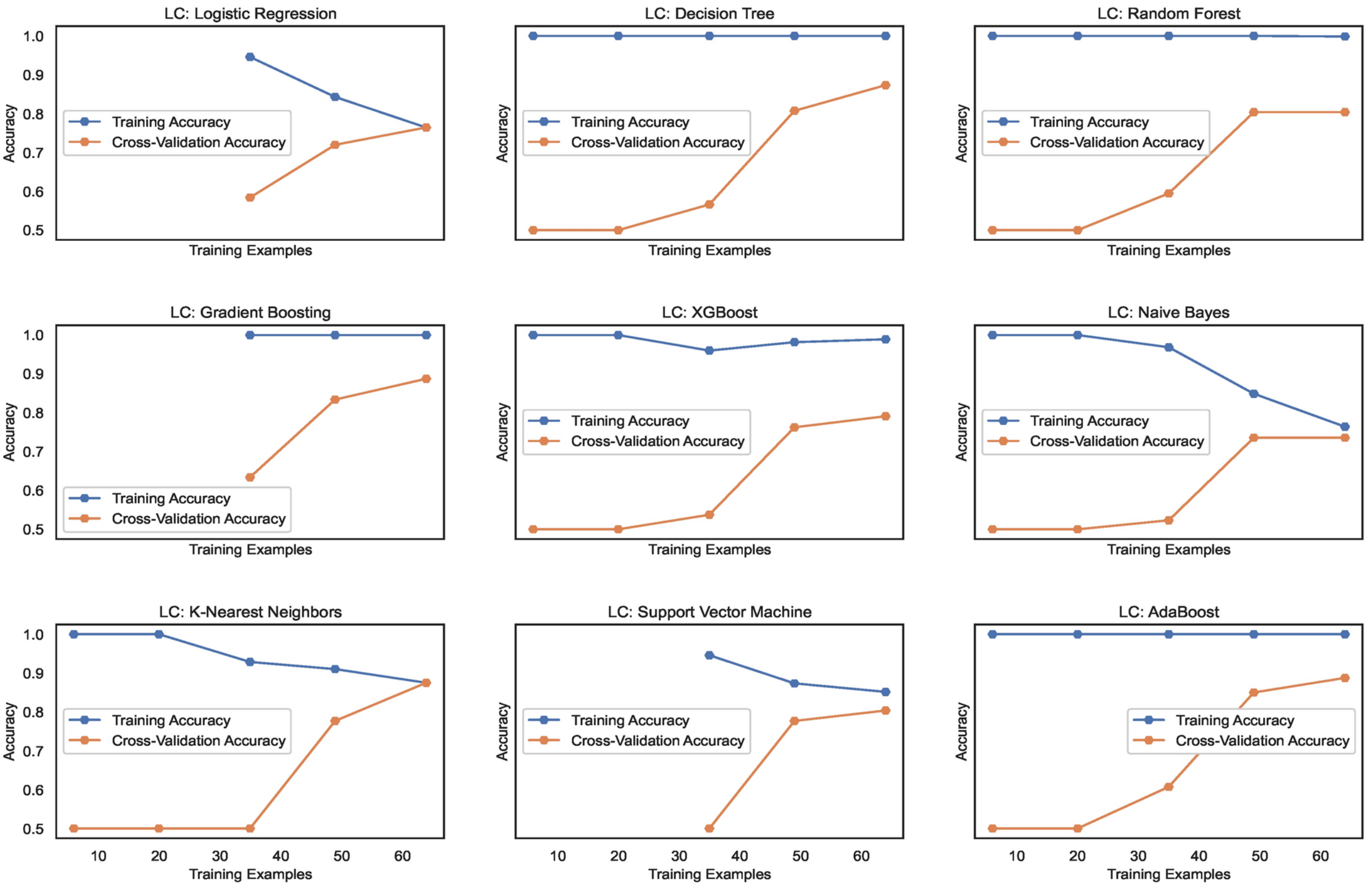

**Fig. 14**. Learning curves for models trained on $DS_0$.

A significant increase can be observed in the AUC for the models trained using augmented dataset $DS_2$. As shown in Fig. 13, out of nine, the AUC value for six models is above 0.9 while the AUC values for NB, LR, and DT are 0.83, 0.86, and 0.88. The observations reveal that classification results improve with large datasets consisting of healthy and PD patients' images. Amongst the nine models, the models capable of attaining the highest AUC value of 0.97 are RF, XGB, and SVM.

*Model learning*

The learning curves, as shown in Fig. 14, display the training accuracies for models trained on $DS_0$ range between 73% to 90% approximately. Logistic regression and Naive Bayes model show decreasing accuracies below 75% whereas the models Extreme Gradient Boosting)(XGB), Random Forest (RF) and Support Vector Machines (SVM) show nearly mid-level prediction accuracies reaching up to 80% approximately. Above the mid-level, however, lower than the highest level, the K-Nearest Neighbors(KNN) and AdaBoost (ADB) models exhibit prediction accuracies reaching approximately 89%. Only the Gradient Boosting(GB) and Decision tree (DT) models predict with the highest prediction accuracy of 91.7%, approximately. The average accuracy range of models is 83%. Likewise, the learning curves shown in Fig. 15 display the training accuracies for models trained on $DS_2$ range between 78% to 92% approximately. Naive Bayes and Logistic regression models show 77% and 78% accuracy, respectively. Only the ADB exhibits the mid-range of accuracy of 83%. The SVM, GB, and DT showcase nearly the same accuracy of 87%, approximately. Only the Decision tree (DT) model predicts with the highest prediction accuracy of 91.7%, approximately. Overall, the average accuracy range of models is 83%. The RF and XBG predict with 90% accuracy; however, the KNN model achieves the highest prediction accuracy of approximately 92%. The overall average accuracy of all the models is 86%, thereby showing a significant increase in accuracy by 3%.

*Model confusion matrix*

Confusion matrices in Fig. 16 show the estimated values of TP, TN, FP, and FN for the nine models. The confusion matrices framed on $DS_0$ show that nearly all the classifiers perform well, except the LR model, which shows a higher account of FPs and FNs. The models DT, GB, KNN, and ADB showcase consideration classification in terms of TPs and TNs, while the models RF, XGB, and SVM exhibit nearly medium classification levels with misclassification. The DT model shows the lowest FPs and FNs and the highest TPs and TNs among all the nine models. This implies DT classifies the spiral curves of healthy and PD patients, comprising a small dataset for classification. In contrast, the confusion matrices for the models trained on large dataset $DS_2$ show significant improvements in the nine models' estimated TP, TN, FP, and FN values. Observation reveals that the NB model shows the highest account of FPs and FNs despite the increase in the size of the dataset. The next classifier to showcase declining performance on the increased size of the dataset is the LR. The models DT, GB, SVM and ADB show the number of FPs and FNs higher than the FPs and FNs for the RF and KNN models and represent models with mid-range misclassifications. However, the KNN model shows the lowest number of FPs and FNs and the highest TPs and TNs among all the nine models. This implies that increasing the size of the dataset for the KNN model increases its classification accuracy on the image dataset consisting of spiral curves of healthy and PD patients.

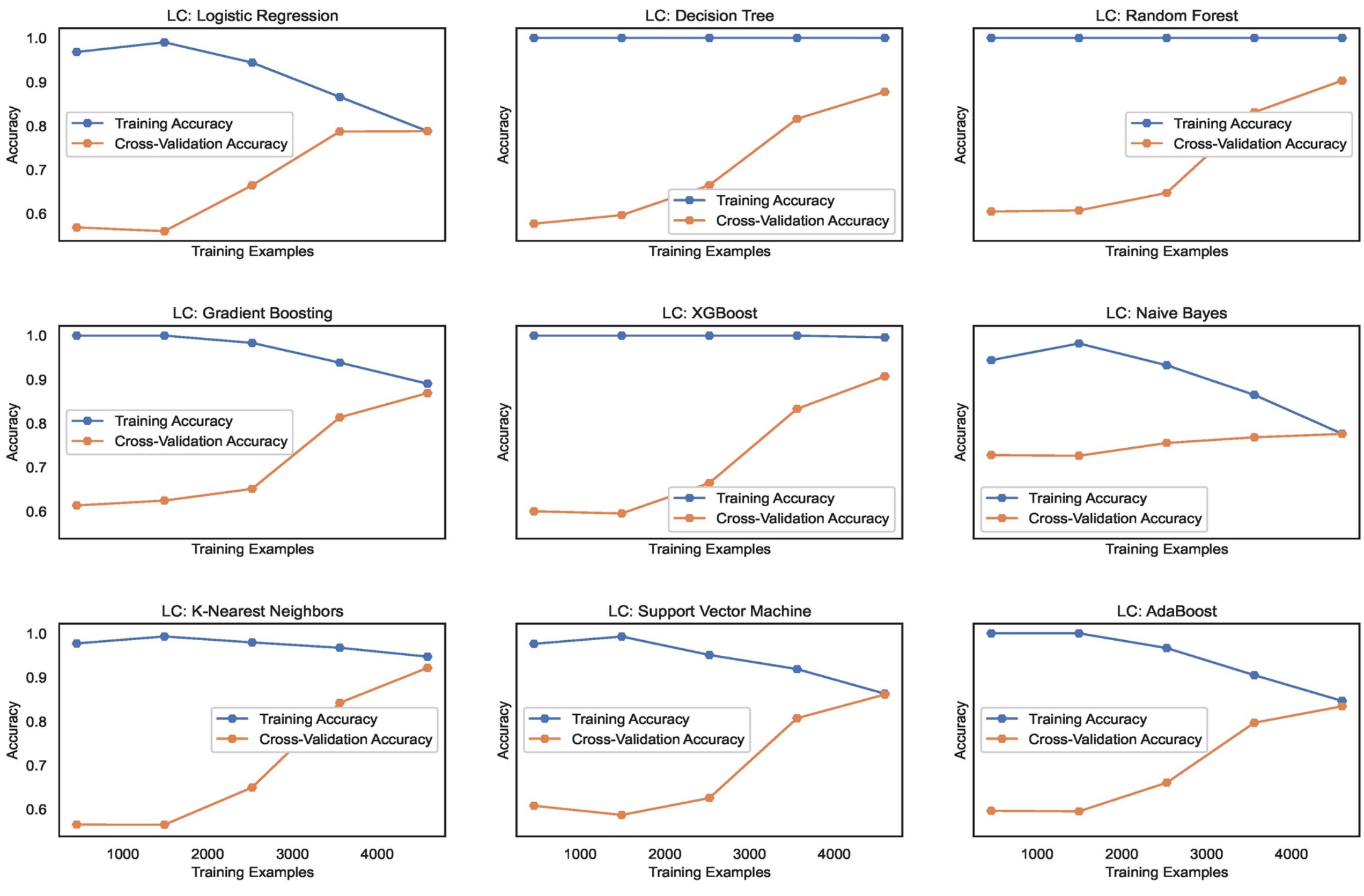

**Fig. 15**. Learning curves for models trained on $DS_2$.

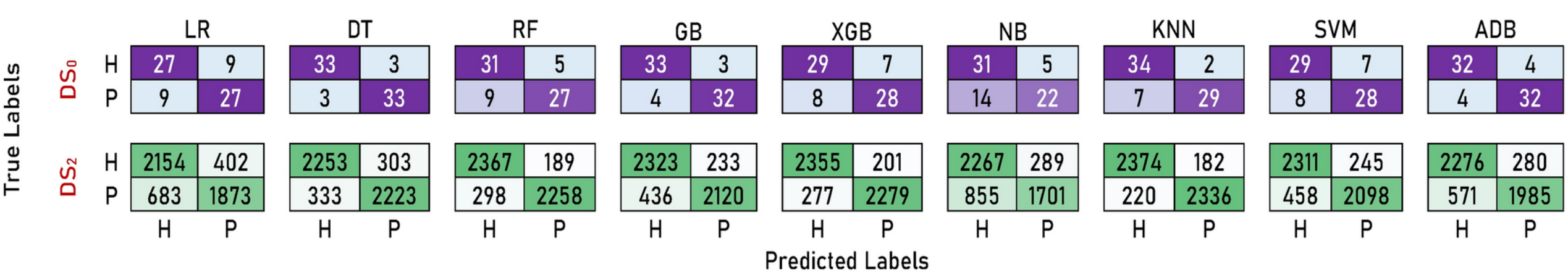

**Fig. 16**. Confusion matrices–results of 10-fold cross validation results on $DS_0$ and $DS_2$. Labels $H$ and $P$ denote the Healthy and Parkinson's class of patients, respectively.

*Bias-variance trade-off*
Bias and variance are the two sources of error that affect model performance. Bias-variance curves are a valuable visualization tool for analyzing the trade-off between bias and variance in ML models. The study considers elucidating the construction and interpretation of bias-variance curves essential. The study attempts to explore the trade-off between bias and variance for the nine ML models. Figures 17 and 18 show the bias-variance curves for models trained on datasets $DS_0$ and $DS_2$. The curves in Fig. 17 depict the bias-variance for models trained on the combinations of $feature_set_DS_0 = [1, 2, 3]$. The different feature combinations are generated for tracing the bias-variance curves representing the models' behaviour on different feature sets. The results reveal that the models trained on a low count of features showcase a near-perfect trade-off between bias and variance. These low count of feature vectors showcase unstablized trends in the means squared error (MSE) for all models except the LR and SVM model with optimal trade-off – stabilized MSE. Other models showcase varying behaviours, although the bias and variance trade-offs exist at certain combinations of features. This implies that the bias-variance trade-off is dynamic in nature and depends on the model type and possible feature combinations employed for classifying the spiral curve of healthy and PD patients.

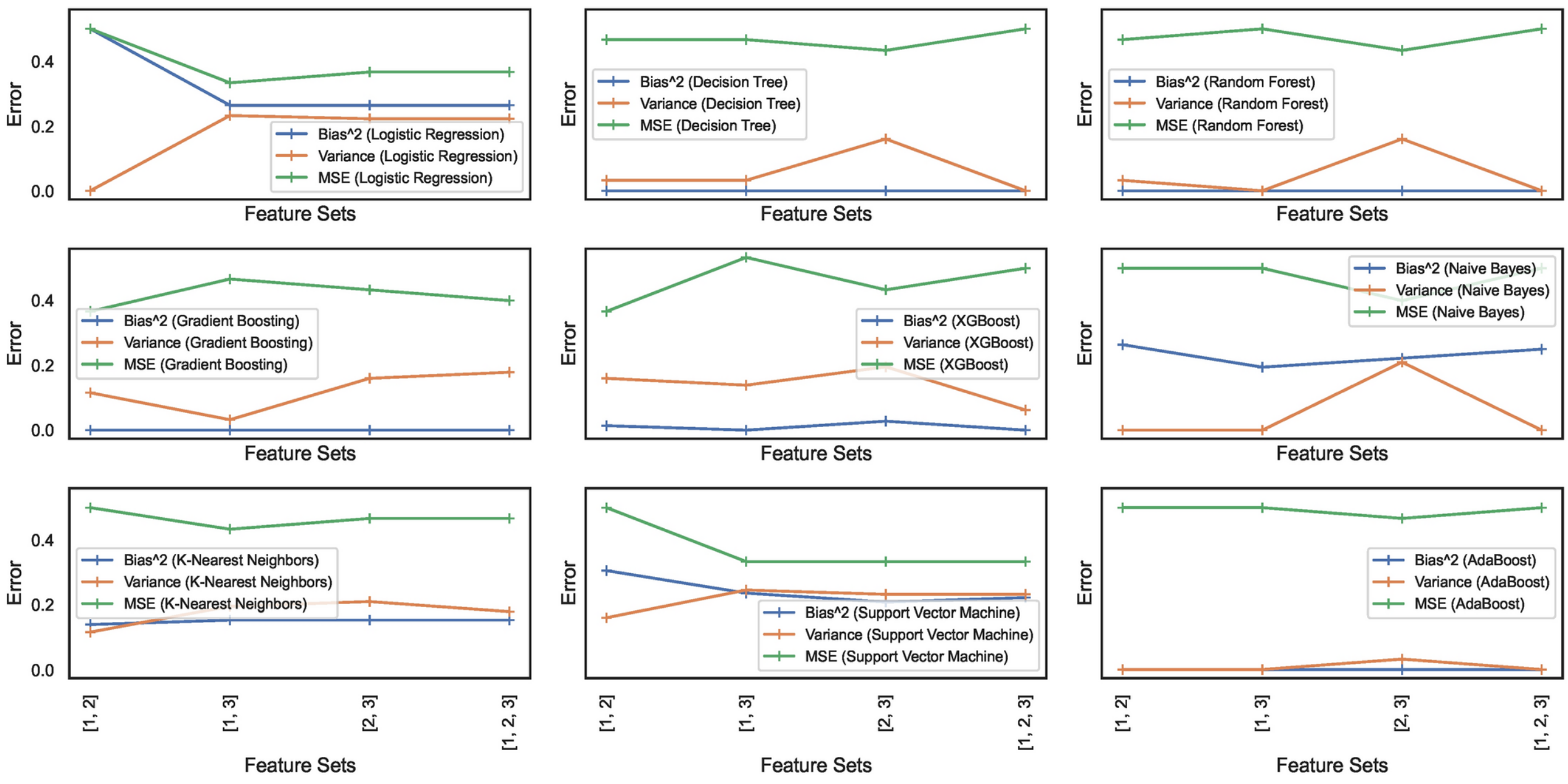

**Fig. 17**. Bias-variance curves for models trained on dataset $DS_0$.

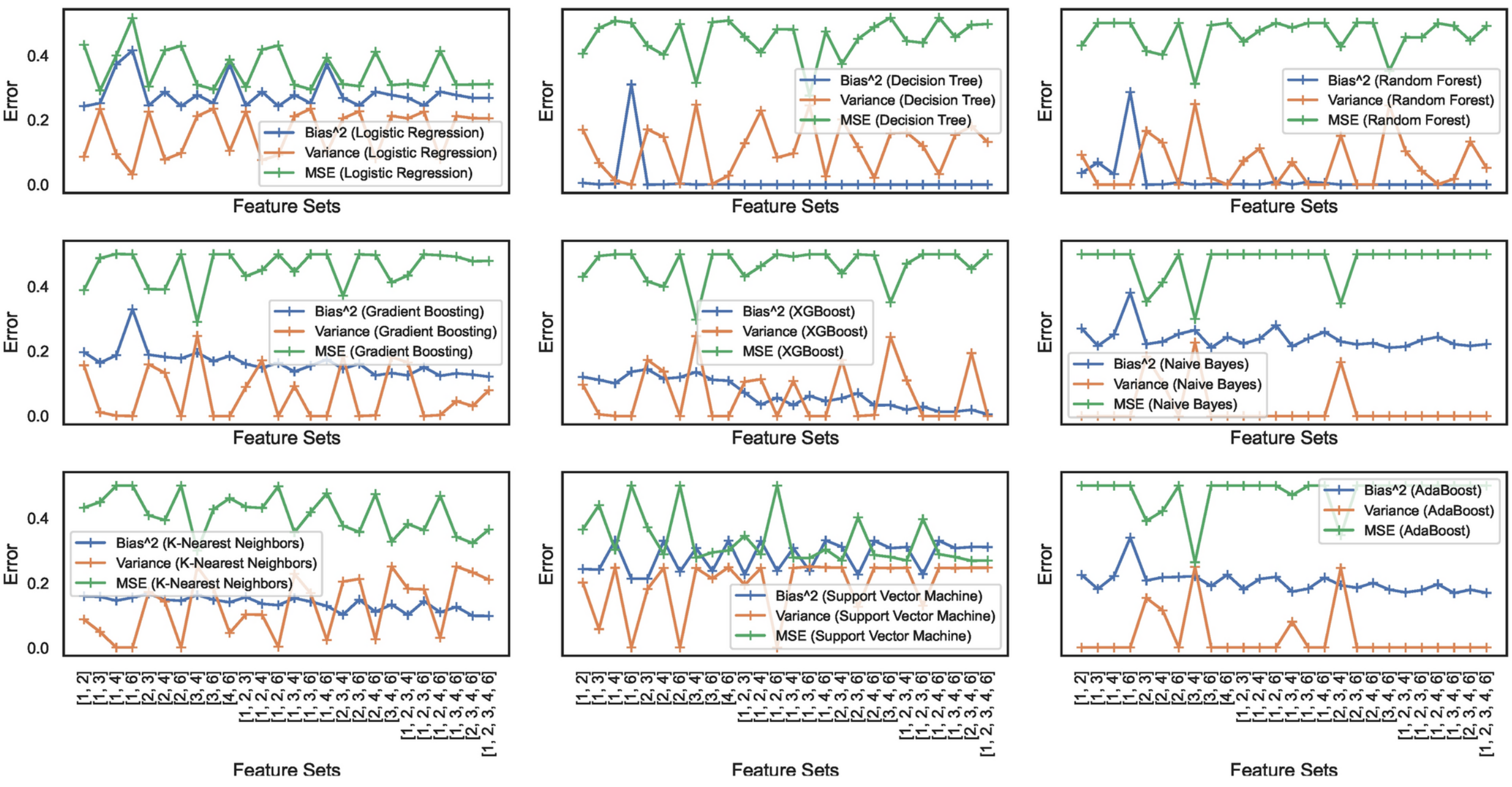

**Fig. 18**. Bias-variance curves for models trained on dataset $DS_2$.

On the other hand, feature sets handcrafted out of augmented dataset $DS_2$ showcase trade-offs at specific combinations of features only. As depicted in Fig. 18, different models exhibit trade-offs at certain specific feature sets. This implies that finding a feature set enabling a perfect trade-off between bias and variance exhibited by the models still requires further investigation. Moreover, the observation reveals that means squared errors for individual models show elevation compared to models trained to $DS_0$. The RF, GB, and XGB showcase declining bias and variance trends; however, with increased MSE. The other models also showcase similar behaviour, except the SVM showcasing a stabilizing trends for the bias, variance and MSE. Analytical works explaining and unveiling such relations are possible candidates for continued investigation in future studies. Therefore, the nine models classifying the handwritten, drawn spiral images by healthy and PD patients need further investigation to ensure bias-variance trade-off could be analyzed at individual levels and then generalized.

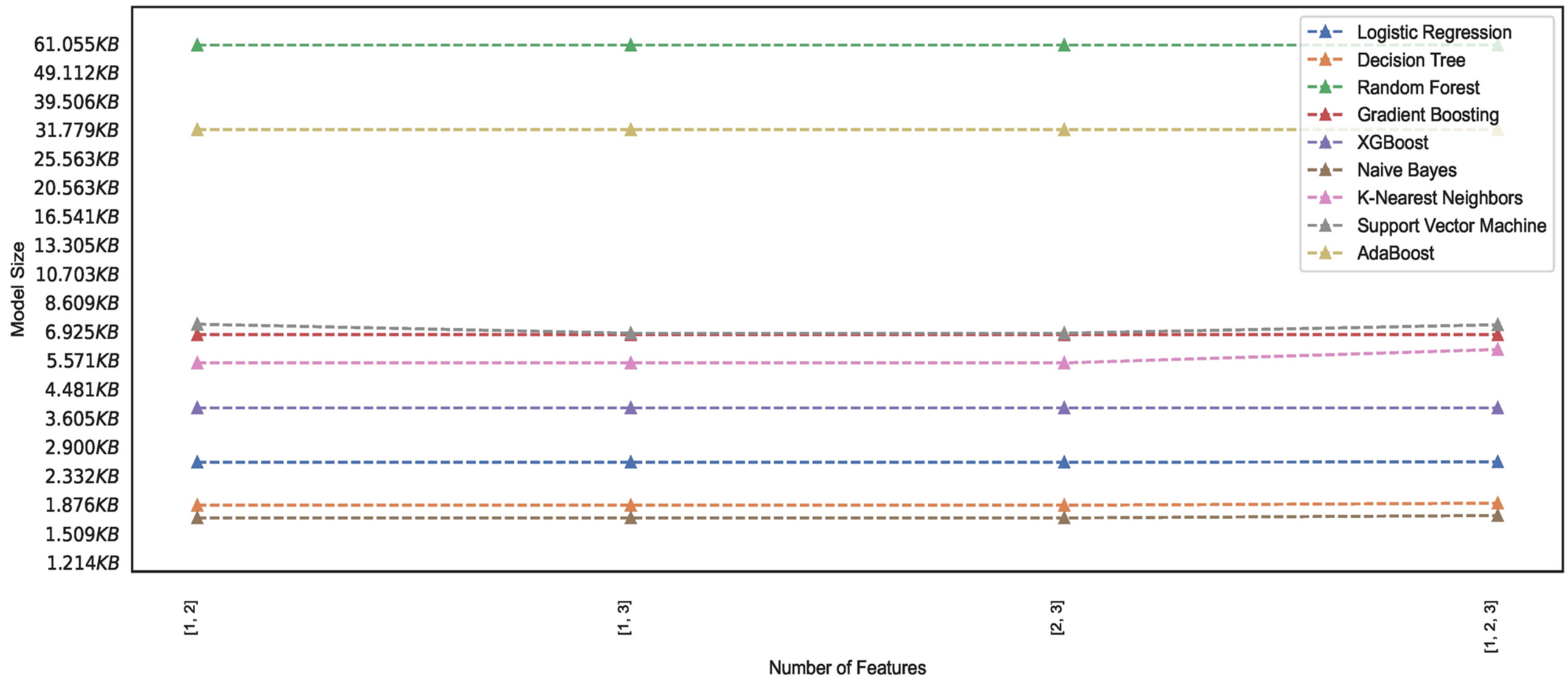

**Fig. 19**. Memory footprint of models trained on dataset $DS_0$.

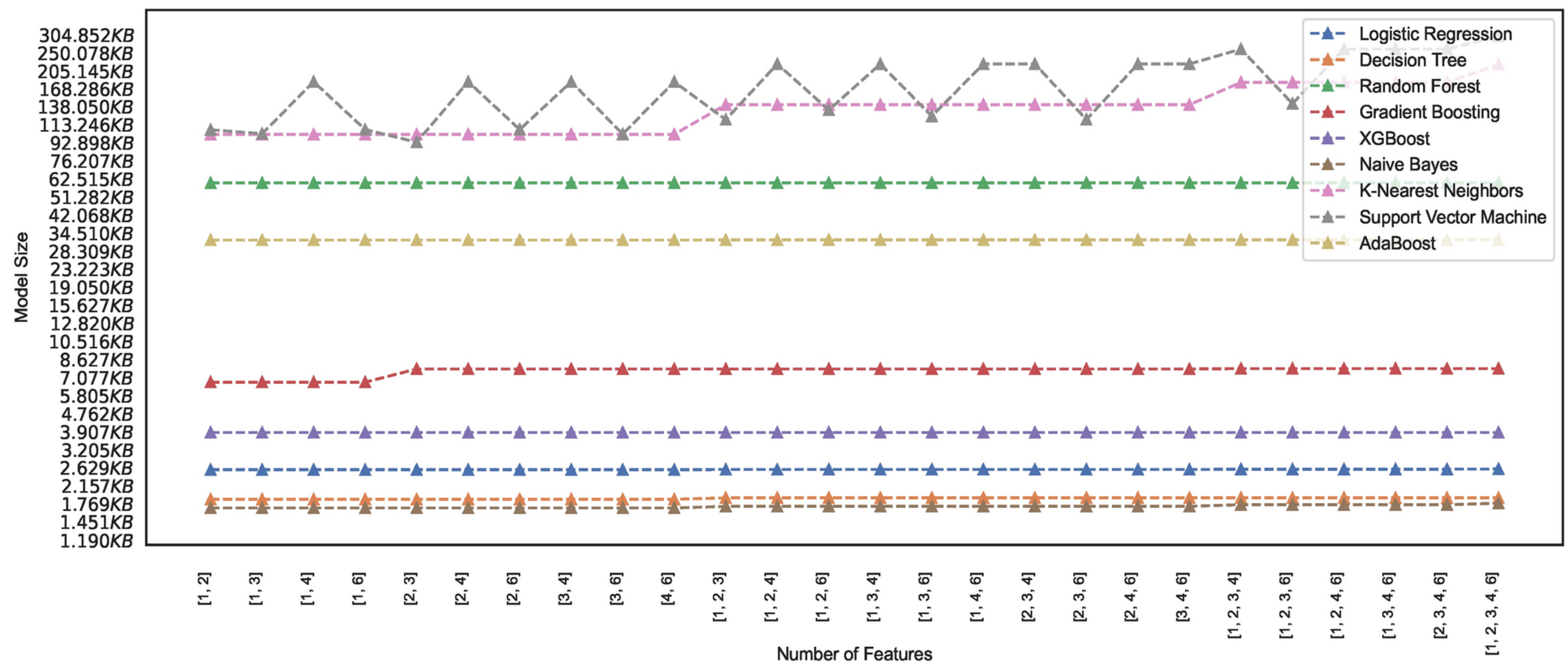

**Fig. 20**. Memory footprint of models trained on dataset $DS_2$.

*Model memory footprint*

Efficient memory management in ML models is critical, particularly when deploying models in resource-constrained environments, mobile devices, or cloud-based systems with limited memory capacity. Researchers and practitioners, therefore, must seek insight into the memory requirements of the models and explore techniques to reduce the memory footprint while maintaining performance. The study presents ML models' memory footprint for classifying healthy and PD patients. Figures 19 and 20 represent the memory footprints of models trained on features combinations derived from $features_set_DS_0 = [1, 2, 3]$ and $features_set_DS_0 = [1, 2, 3, 4, 6]$ , respectively. The RF and ADB models depict higher memory requirements than those in Fig. 19. Expect the RF and ADB models, a maximum of three feature vectors employed for training all the models on spiral image dataset depict lower memory requirements ranging between 1KB to 7 KB. While the second-highest model, ADB, showcases a memory requirement of approximately 33 KB, the RF model showcases the highest memory requirement of approximately 61 KB. Constant memory requirements could be observed for other models; however, KNN and SVM show elevated requirements with an increase in the number of feature counts of two ([2, 3]) to three ([1, 2, 3]) features. Observation reveals that some models showcase constant memory requirements, whereas others exhibit increasing memory requirements and increased feature space.

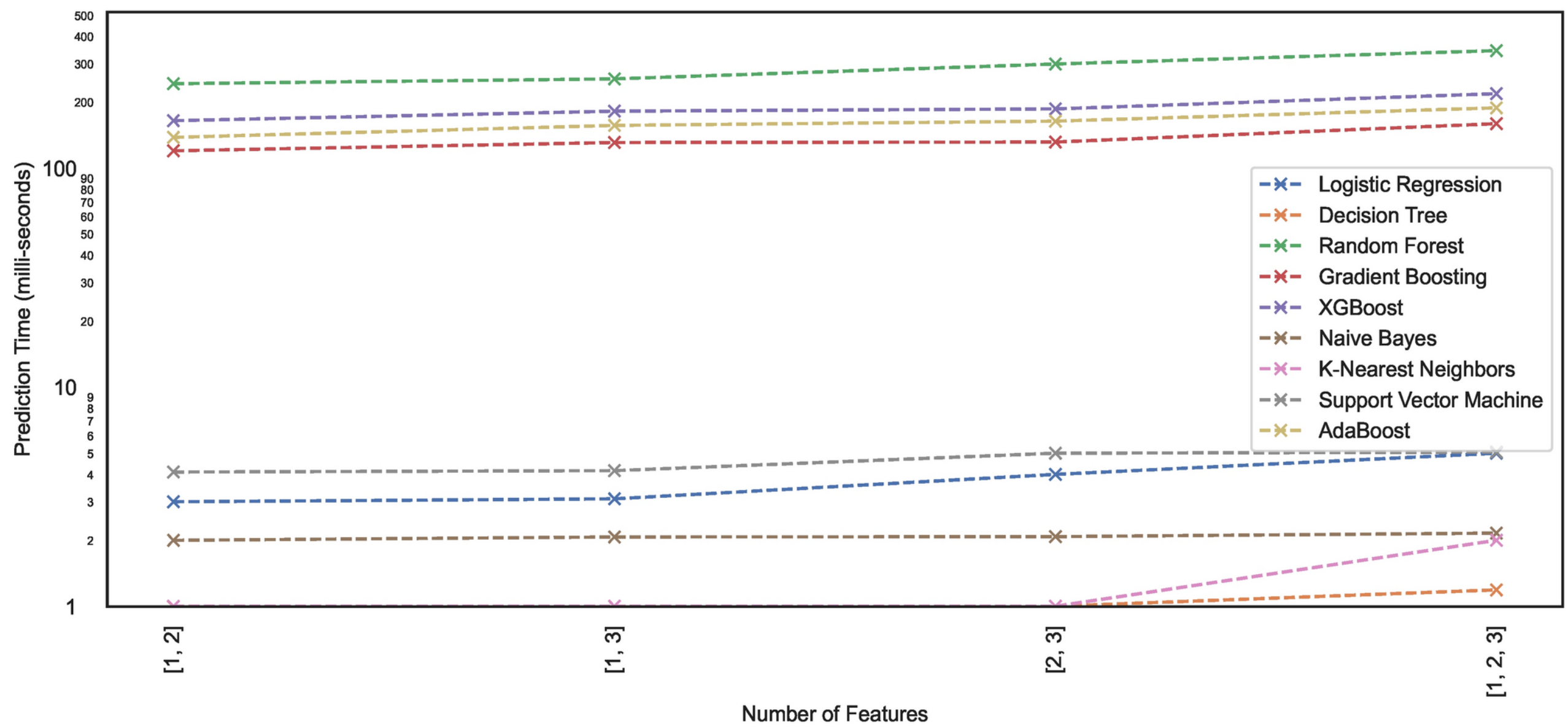

**Fig. 21**. Prediction time of models trained on dataset $DS_0$.

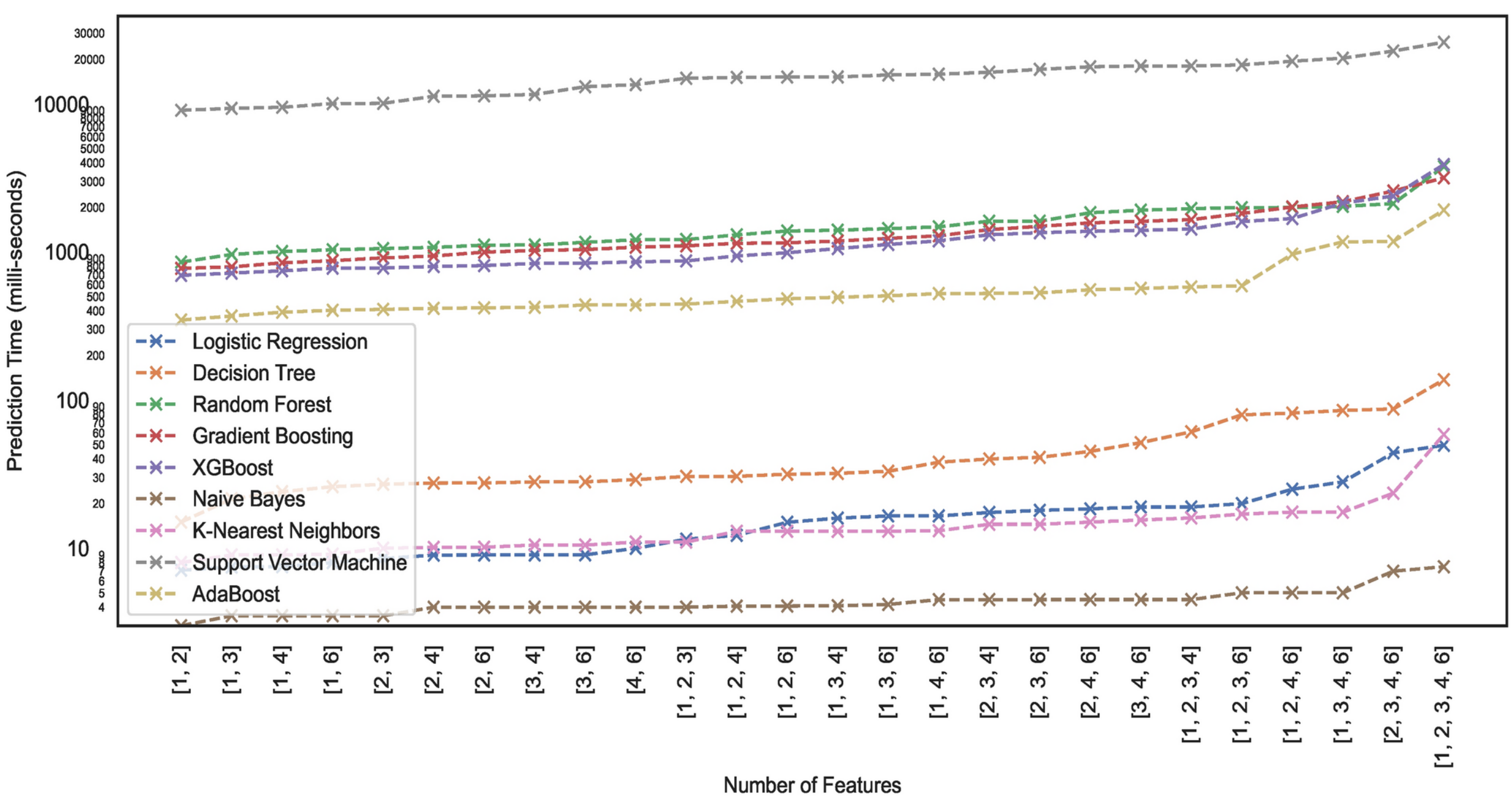

**Fig. 22**. Prediction time of models trained on dataset $DS_2$.

Comparing the memory requirements of models trained on $DS_0$ and $DS_2$ showcases the various models' varied differences in memory size requirements. Figure 20 depicts the memory usage by various models. Observing the memory footprint reveals that some of the models trained on augmented dataset $DS_2$ showcase elevation in memory requirements while the other models exhibit the same memory requirement as observed in models trained on small size dataset $DS_0$. The models GB, XGB, LR, DT, and NB trained on $DS_0$ exhibit memory space requirements ranging between 1KB to 7.5KB approximately, while models ADB, RF, KNN, and SVM showcase increasing memory space requirement when trained on $DS_2$. As depicted in Fig. 20, the SVM memory requirements are dynamic and rely mainly on different combinations of features. However, all other models, except the kNN and GB, showcase a consistent memory requirement across the entire feature space. Observing the memory footprint, in general, reveals that for some models, the memory requirements are independent of the feature space; however, some models showcase an increase in the memory requirement with the increase in the number of feature vectors employed for training the models. This study recommends selecting model features that ensure consistency in model size with low memory requirements to meet the demand for resource-constrained devices. Edge-computing devices with memory constraints could benefit from this explainable feature of training the models since model sizes suitable for memory-constrained devices can be easily figured out. This study focuses on classifying images of hand-drawn spirals by healthy and PD patients based on ML models with low memory requirements only.

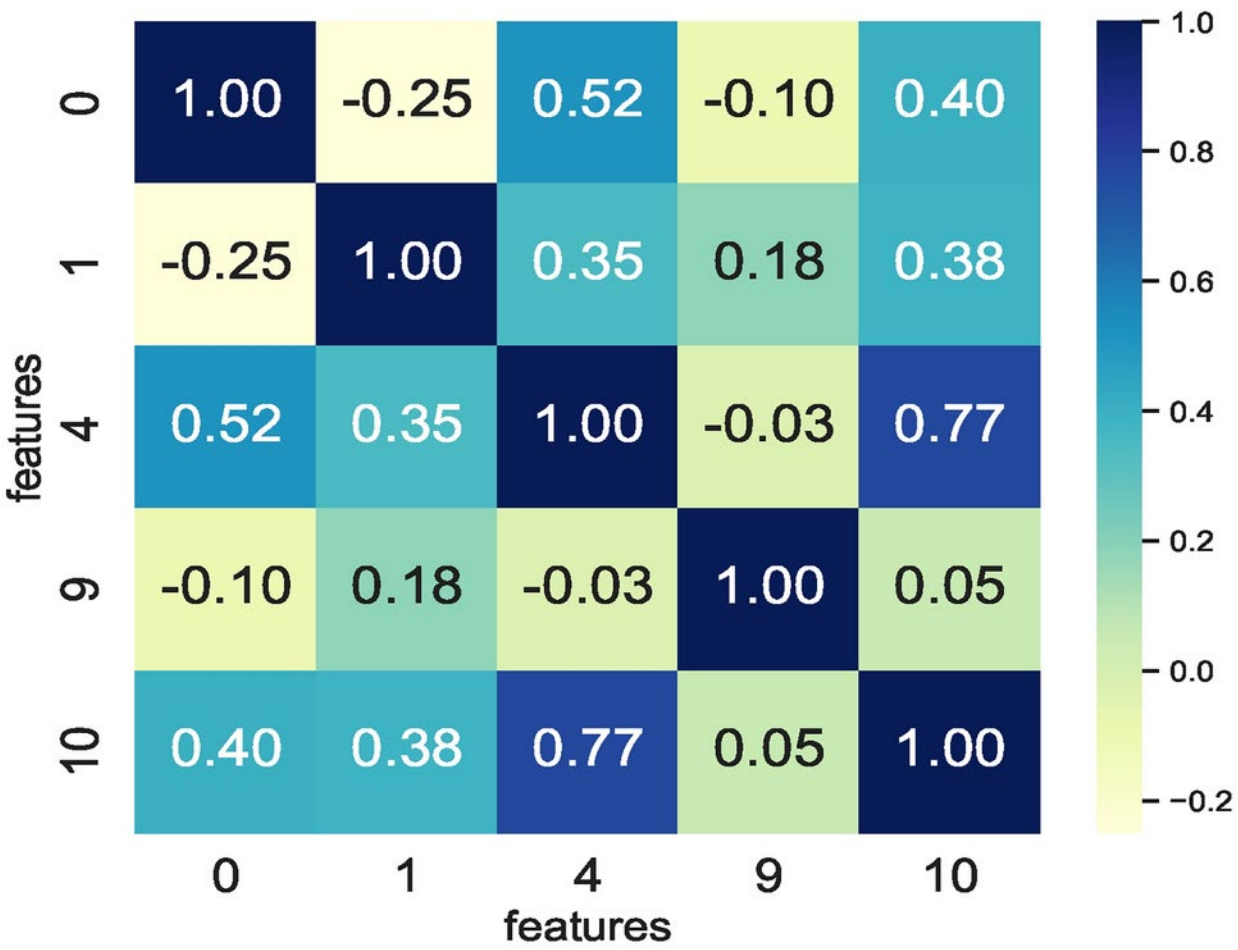

**Fig. 23**. Correlation matrix defining correlation of feature combinations handcrafted from $DS_1$.

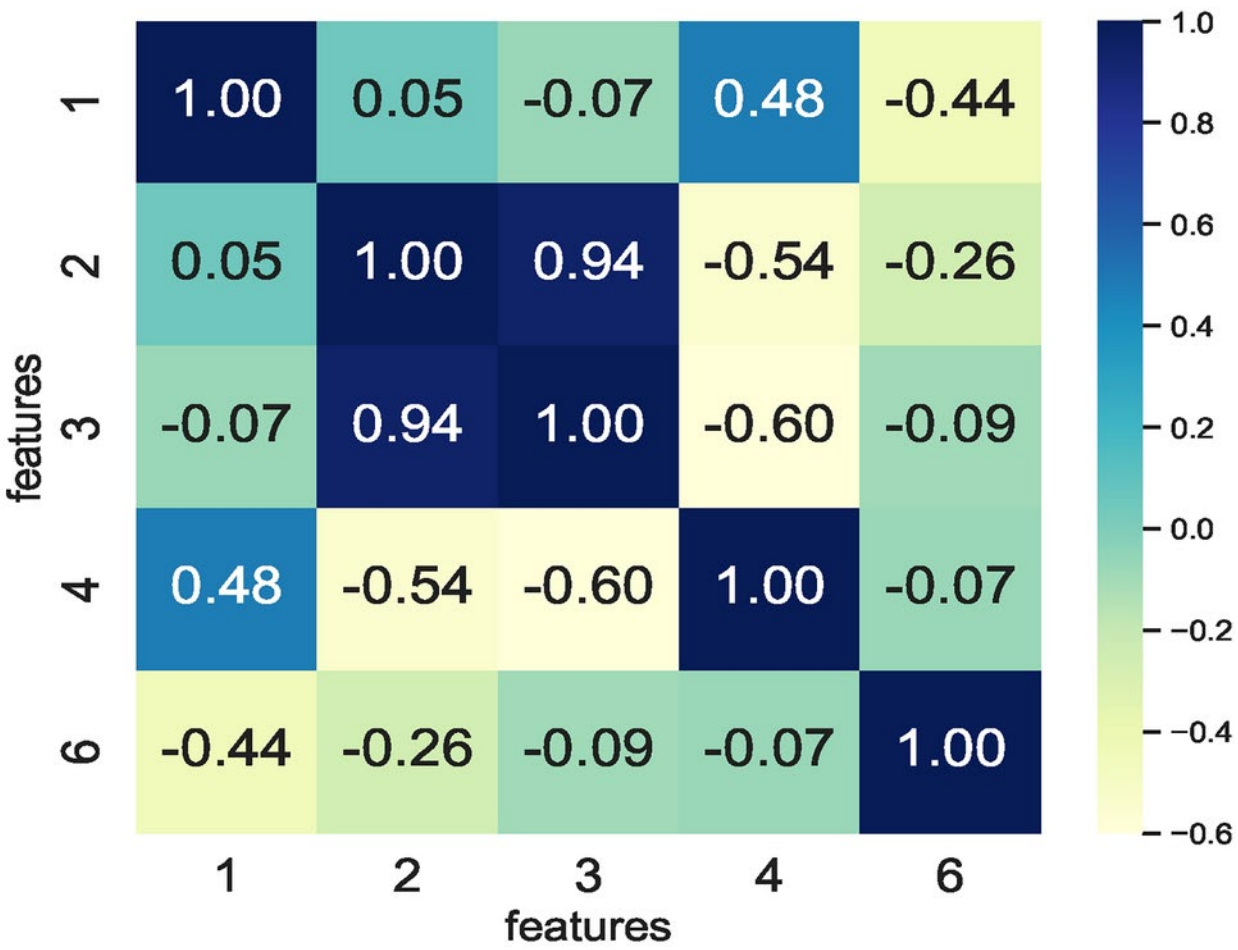

**Fig. 24**. Correlation matrix defining correlation of feature combinations handcrafted from $DS_3$.

*Prediction time*

Prediction time typically refers to the time a trained model takes to produce a prediction or output for a given input. Model prediction time is particularly crucial in edge computing environments. Since edge devices often have limited computational resources compared to powerful cloud servers, model prediction time becomes critical. In edge computing, devices and sensors at the network edge collect and process data, and ML models may be deployed directly on these edge devices. This study lists the prediction times exhibited by nine models employed for classifying the hand-drawn spirals of the PID. Figures 21 and 22 illustrate the prediction time of these models employed in this study. The results show that a few of the models showcase higher prediction times compared to the others. For instance, the ML models – RF, XGB, ADB, and GB take $> 100ms$ to generate predictions on the $DS_0$ while trained on the same dataset, the ML models – SVM, LR, NB, KNN, and DT can generate fast predictions $< 10ms$. Moreover, the ML models trained on the augmented dataset $DS_2$, being the large sized dataset $DS_0$, showcase different prediction behaviours. Some models retain the prediction rates while others showcase variations, either increasing with the size of the dataset $DS_0$ or even the opposite. However, all the ML models showcase an incline in their prediction times with the increase in the size of the dataset.

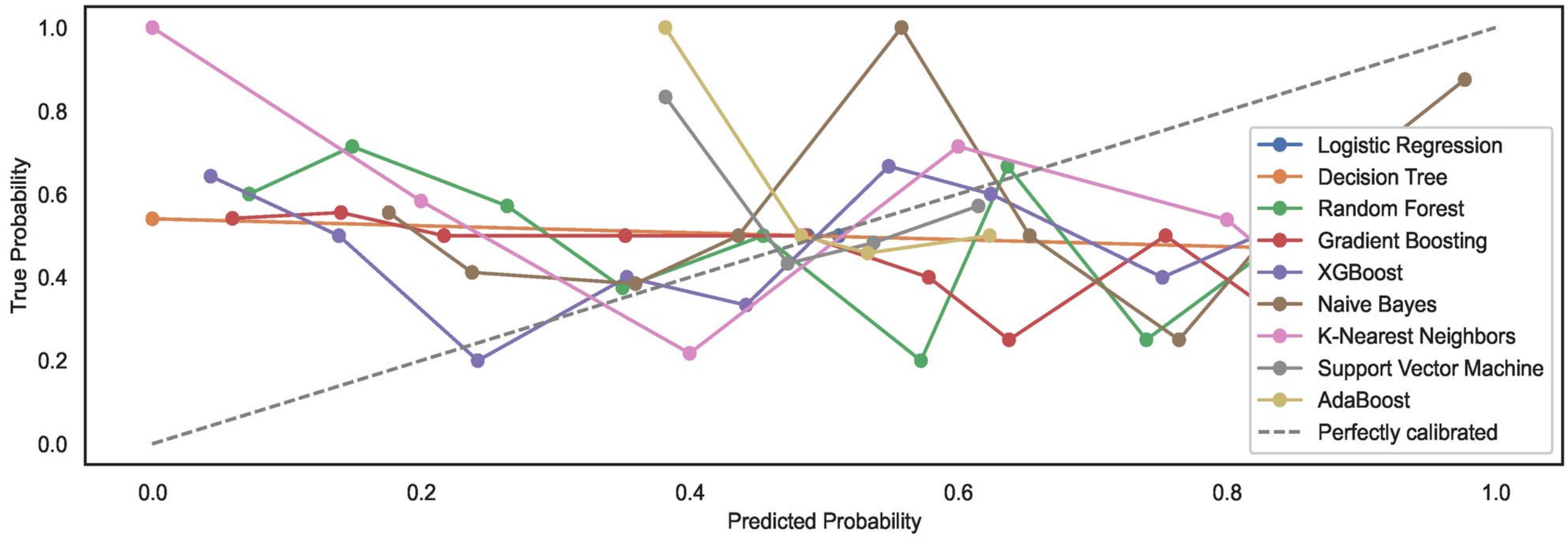

**Fig. 25**. Calibration curves for models trained on $DS_1$.

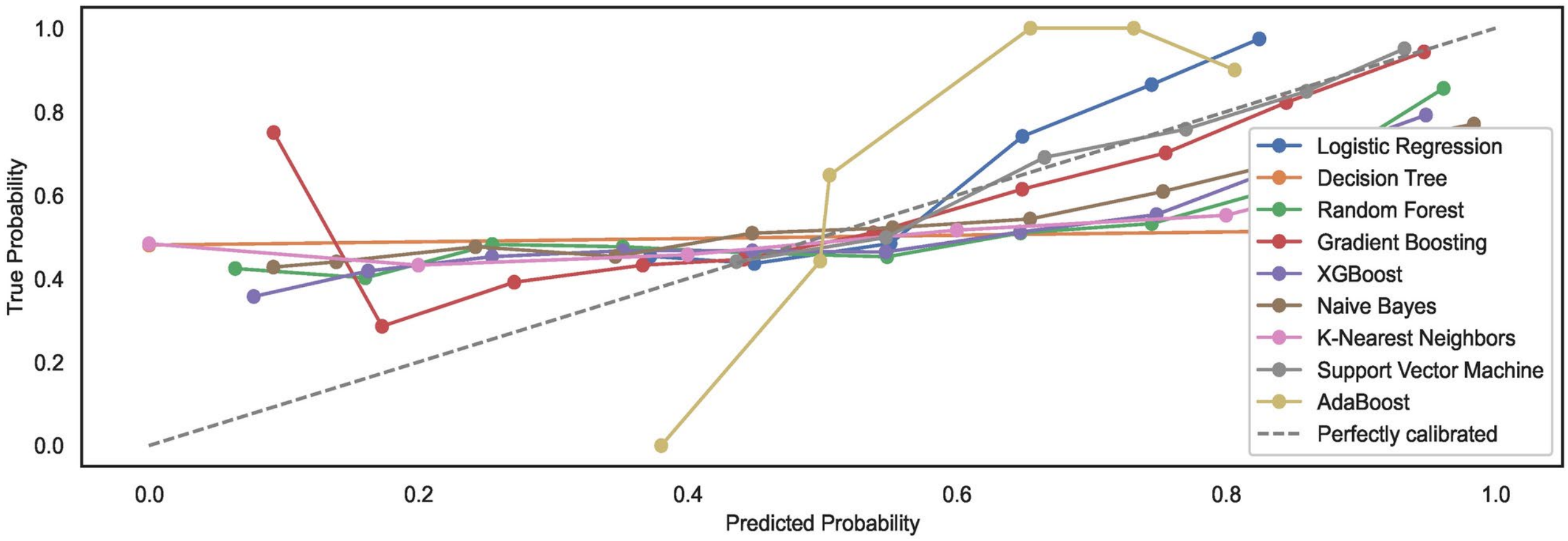

**Fig. 26**. Calibration curves for models trained on $DS_3$.

## ML model training on $DS_1$ and $DS_3$

The correlation matrix shown in Fig. 23 represents five feature vectors, and the combination of these five feature vectors represents the feature space seen as input by the nine ML models. Again, the RFS removes the highly correlating features here while identifying feature sets specific to each model. Following the same approach employed for dataset $DS_0$, the feature set enabling similar prediction accuracies by all nine models is selected. Therefore, feature set $feature_set_DS_0 = [0, 1, 4, 9, 10]$ represents the best possible feature set among all the RFS-generated features to train the models. In contrast to the prior dataset $DS_0$, reducing the feature space by eliminating features results in low model prediction accuracy. As such, the feature space generated by the RFS trains the models and reduces the number of features impacting model accuracies.

Training models to achieve an optimum accuracy level required deriving a new feature space from the augmented dataset $DS_3$ instead of using the feature space derived from $DS_1$ since model prediction scores remained low in the models trained on feature space derived from the dataset $DS_1$. The correlation matrix in Fig. 24 shows the correlation within the feature space derived from the augmented dataset $DS_3$. Observing the RFS-generated feature space consisting of $feature_set_DS_3 = [1, 2, 3, 4, 6]$ for $DS_3$ reveals that the models' prediction accuracies remain the same. The $feature_set_DS_3 = [1, 2, 3, 4, 6]$ is a feature set handcrafted from the augmented dataset $DS_3$. The feature space derived from $DS_3$ for training the models ensured that models achieved nearly the same prediction accuracy. This implies that applying canny edge detection to the images of hand-drawn spirals by healthy and PD patients lowers the performance of models. Moreover, the feature space derived from augmented datasets remains the same, implying that augmenting the dataset $DS_1$ ensures consistency in the feature space. This in turn, enables models to source predictions at the same level whilst performing classification on healthy and PD patients.

*Model calibration*

The study employs un-calibrated models with default configuration settings, as mentioned in the prior sections. Expected model behaviours showcase a non-alignment with the perfect calibration line when the nine models are trained on dataset feature space derived from $DS_1$. Compared to $DS_1$, the results showcase alignments of models' calibration curves to some extent with the perfect calibration line while the models are trained with the augmented dataset $DS_3$. Again, an augmented dataset does not lead to ideal $S - shaped$ curves that result from calibrated models.

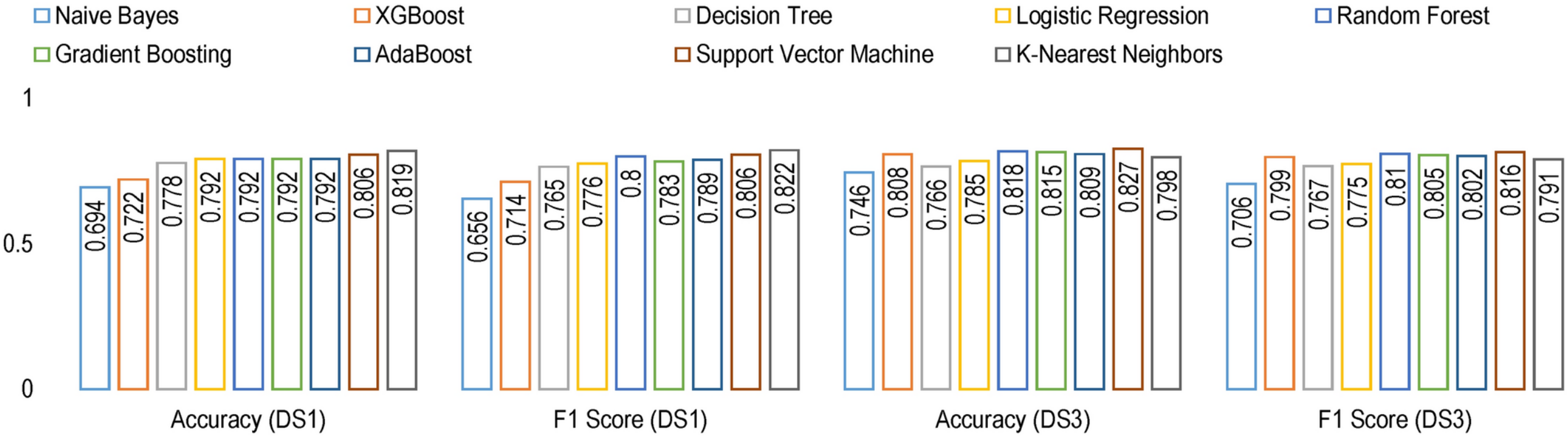

**Fig. 27**. Results of model evaluation results on $10 - fold$ cross validation.

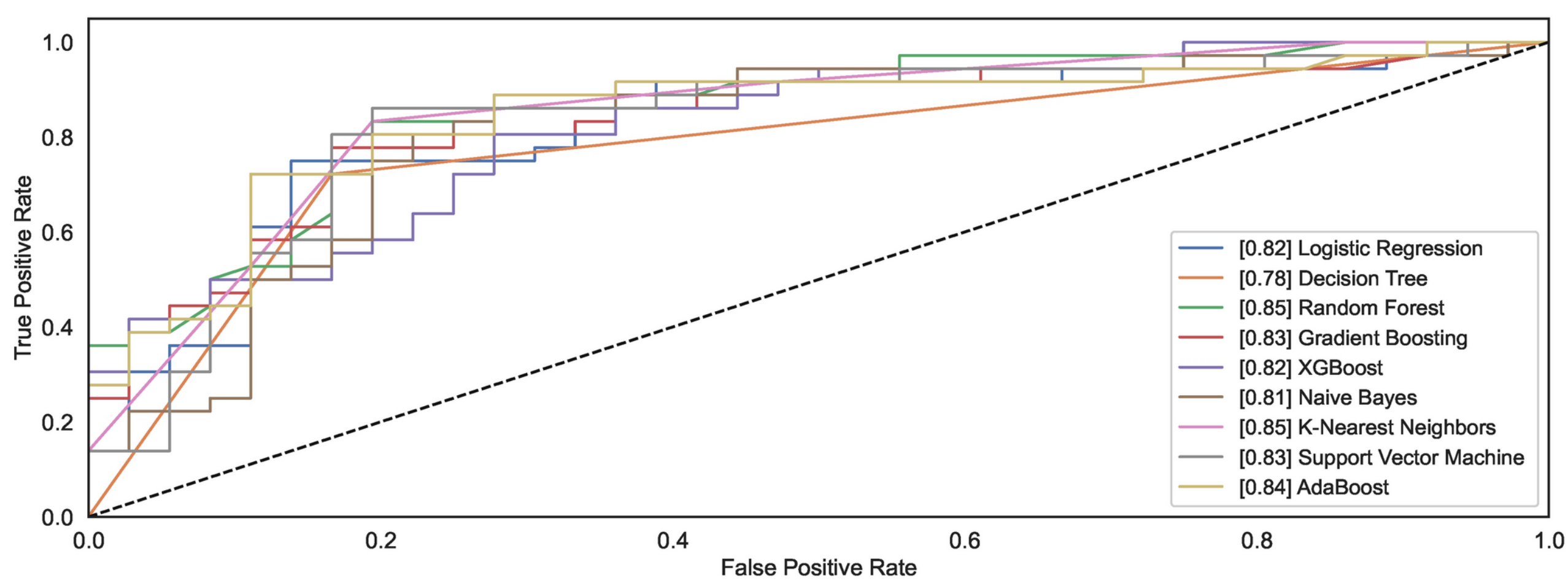

**Fig. 28**. ROC curves for models trained on $DS_1$.

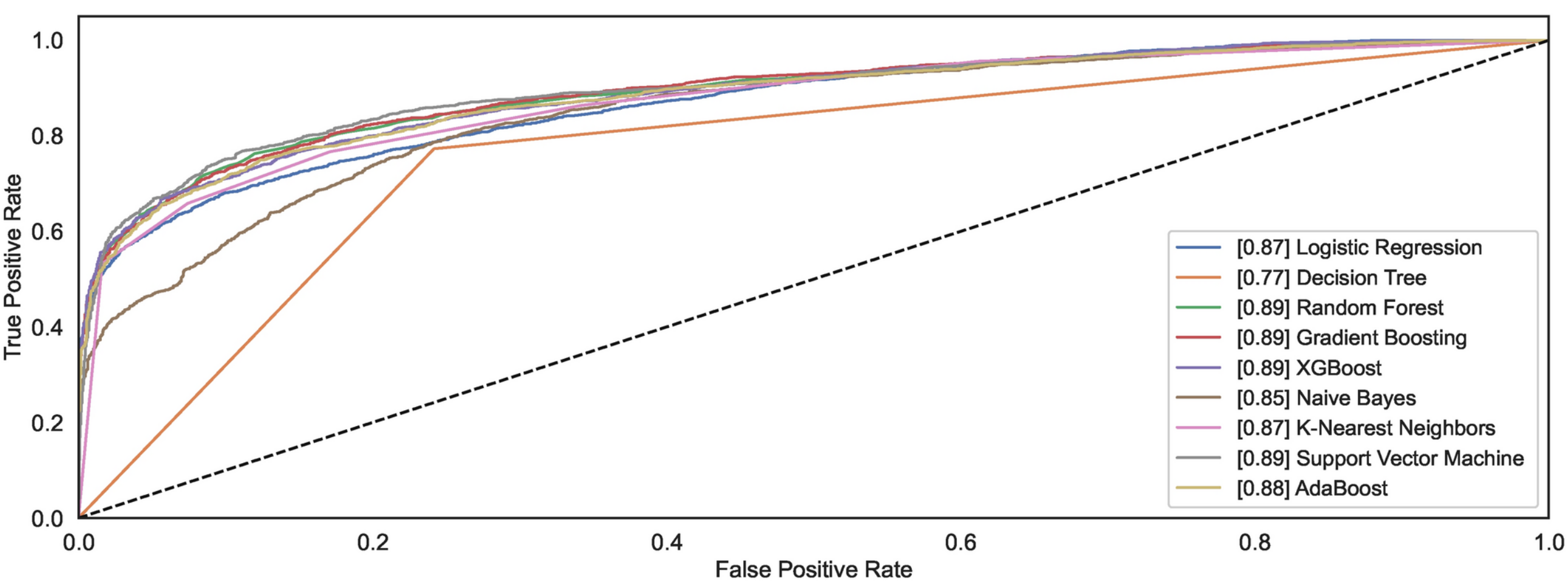

**Fig. 29**. ROC curves for models trained on $DS_3$.

For observing the difference, the study shows that large-size datasets automatically calibrate the models to some extent. However, images of hand-drawn spirals by healthy and PD patients, when subjected to pre-processing with a canny edge detector, lower the prediction accuracies of ML models. In addition, the models trained on small-size datasets exhibit model calibration curves that remain non-aligned to the ideal perfect calibration model state curve, implying the model prediction accuracies eventually decrease with un-calibrated models. The study limits the scope to un-calibrated models only; however, the task of realizing the classification of calibrated models is left over for future work. Figures 25 and 26 show the calibration curves for the models trained on small and large datasets used to derive the feature space size, respectively. The models trained on large dataset $DS_3$ showcase curves nearly aligning with the perfect model calibration line, nonetheless not exhibiting the perfect $S-shape$ curves resulting in calibrated models. The study highlights that calibrating models before training may improve the prediction accuracies of the models further. Estimating the calibrated performance has been left over for future works.

*Model training*

Figure 27 shows the cross validation accuracy and F1-score for the nine models evaluated using stratified $10-fold$ cross validation.

## Performance evaluation on $DS_1$ and $DS_3$

*Model ROC*

Figures 28 and 29 show the ROC plots for the models trained on $DS_1$ and $DS_3$. The $DS_1$ results show that the models classify spiral drawings of healthy and PD patients with an AUC value ranging between 0.81 to 0.85. All classifiers show nearly the same AUC values; however, only the DT decreases AUC levelling at 0.78. RF and KNN show the same value of AUC 0.85 while the AUC values for the models NB, LR, XGB, GB, SVM, and ADB remain lower than RF and KNN. RF and KNN show the highest AUC of 0.85. The models trained on augmented dataset $DS_3$ depict increased AUC values. Classifiers RF, GB, XGB, and SVM show an increase in the AUC values, reaching up to the AUC of 0.89. However, the AUC for DT remains low even if the size of the dataset increases. NB, LR and KNN almost share the same AUC of 0.85, 0.87 and 0.87, respectively. The ADB just follows the models with the highest AUC values. The observation reveals that datasets $DS_1$ and $DS_3$ show consistent AUC for all models except LR. The AUC for both the datasets almost follow each other despite different. This implies that the canny dataset showcases consistent performances across the range of models.

*Model learning*

Figure 31 shows the learning curves for models trained on $DS_1$ range and the level of cross-validation accuracies of various models that range between 71% to 82% approximately. The KNN model shows the highest accuracy of 82%, while XGB shows a lower accuracy level than the other models. The SVM and ADB almost show the same cross-validation accuracy. Training accuracies almost remain above 80% for all the models. However, for NB and SVM, it lowers gradually with increasing count of samples. Figure 32 shows the training and cross-validation accuracies for models trained on $DS_2$. The training accuracy for all ML models is high and ranges between 78% to 100%, approximately. However, the cross-validation accuracy range compared to training accuracy is low and ranges between 73% to 83%. While the highest cross-validation accuracies are observed in RF, GB, XGB, and SVM, the LR, DT, KNN, and ADB models showcase lower cross-validation performance. The NB shows the lowest cross-validation accuracy among all the nine models. Observing both the learning curves for models trained on $DS_1$ and $DS_3$, a performance improvement in cross-validation accuracy can be observed. However, the performance stays behind the cross-validation performance exhibited by $DS_0$ and $DS_2$. This implies canny edge detection; although it ensures consistency in the models' performance, however levels down the cross-validation accuracy of the models at the same time.

*Model confusion matrix*

The confusion matrices in Fig. 30 show the estimated TP, TN, FP, and FN values for the nine models. The TP for all the models is nearly identical except for RF and XGB. Out of 36 image samples of hand-drawn spiral images, 29 to 31 are correctly identified by the classifiers. However, the models show differences in identifying the TNs. LR, DT, GB, and XGB showcase accounts of TNs lower in count than TNs identified by RF, KNN, SVM, and ADB. The model exhibiting the lowest TNs is NB, with just 21 samples. The confusion matrix shown in Fig. 30 for the models trained on dataset $DS_3$ shows that nearly all the models showcase consistent behaviour in identifying the TPs and TNs, except with some minor differences. However, the NB model shows the lowest TNs among all the nine models. All nine models showcase the TP and TN rates. This implies the augmented spiral dataset based on canny edge detection enables all models to behave consistently, providing the same level of accuracy across the entire range of employed models in the study. While this might be a trade-off representing a suitable dataset for training the models for equal performance, the accuracy levels of models are lower than model cross-validation accuracies enabled by the datasets $DS_0$ and $DS_2$. This shows that a canny edge detection-based dataset can generalize model behaviours; however, it may simultaneously reduce the performance of the models.

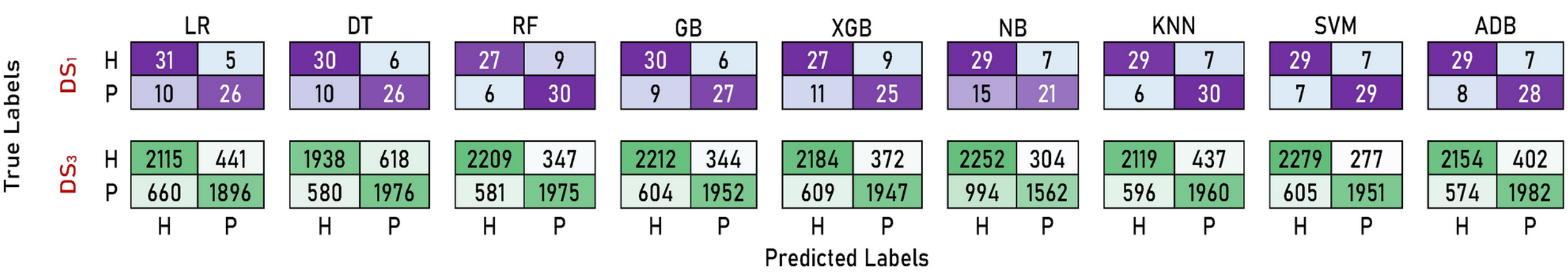

**Fig. 30**. Confusion matrices–results of 10-fold cross validation on $DS_1$ and $DS_3$. Labels *H* and *P* denote the Healthy and Parkinson's class of patients, respectively.

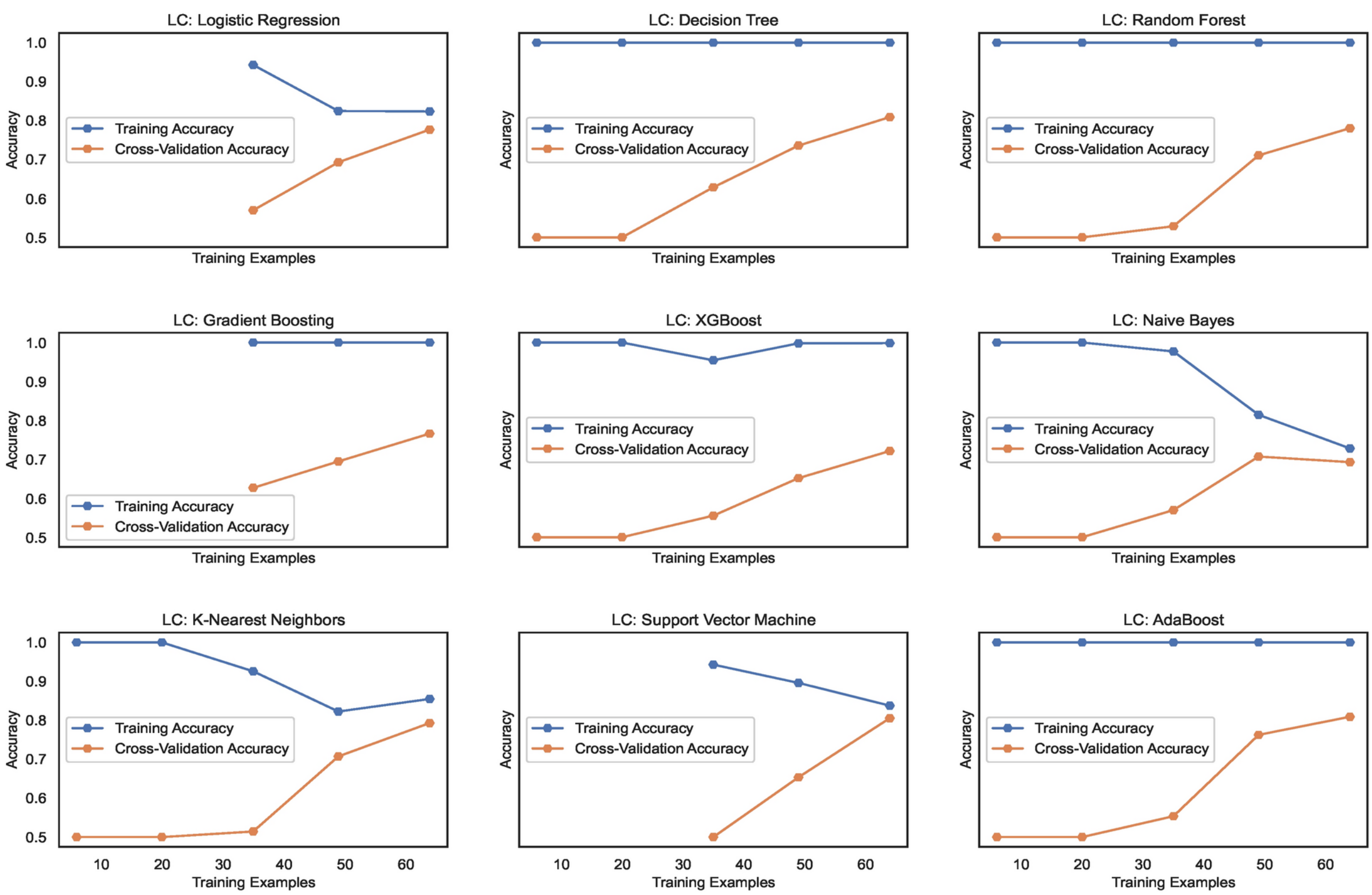

**Fig. 31**. Learning curves for models trained on $DS_1$.

*Bias-variance trade-off*
Figures 33 and 34 show the bias-variance curves for models trained on datasets $DS_1$ and $DS_3$. Feature combinations for tracing the bias-variance curves represent different models' behaviour on different feature sets. The curves in Fig. 33 depict the bias-variance for models trained on the combinations of $feature_set_DS1 = [0, 1, 4, 9, 10]$. Results reveal that the means squared error remains high for all models trained on canny image dataset $DS_1$. The plots showcase bias-variance trade-offs for different models whilst showcasing a constant level of bias-variance for the model KNN and NB. The KNN bias-variance curves almost lie parallel to each other, implying certain combinations of features result in an optimum model. However, for models other than KNN, there appears to be a considerable insignificant bias-variance trade-off and the curves for the models XGB, NB, and SVM show minor deviations. The results show that different models exhibit different bias-variance trade-offs at specific feature combinations in the small dataset $DS_1$. Moreover, Fig. 34 shows bias-variance curves traced for models trained on dataset $DS_3$. It should be noted that nearly all the models showcase stabilized variations in the bias, variance and MSE. The variance is observed to be constant in the all the models when trained on $DS_3$ while the MSE increases with the increased size of the dataset. The DT model only showcases a declining trend in the bias compared to the other ML models. Again, different models exhibit trade-offs at certain specific feature sets. However, only the KNN shows a good trade-off between the bias and variance as both appear to be in the close error range, although the MSE being high. XGB is the next possible candidate, showing trade-offs between the bias and the variance. Models other than KNN and XGB show considerable error differences between the bias and variance. Apparently, the only model showing optimal fit is the KNN. Furthermore, increasing the size of the feature space does not affect the bias and variance statistics here.

*Model memory footprint*
Figures 35 and 36 represent the memory footprints of models trained on features combinations derived from $features_set_DS_1 = [0, 1, 4, 9, 10]$ and $features_set_DS_3 = [1, 2, 3, 4, 6]$, respectively. Compared to the dataset $DS_0$, the canny image dataset $DS_1$ leads to an increase in the model sizes ranging from 1.5KB to 61KB approximately. While most model sizes range below 10KB, RF and ADB models showcase high memory requirements. The RF shows the highest memory requirement among all of the models employed. The NB and DT models show almost similar model sizes whilst showing a slight increase in the memory footprint with the increasing size of the feature space. The LR and XGB almost show constant model size over the entire range of feature space. In contrast, the KNN, GB, and SVM depict an increase of 3KB in the size of models size for the increasing feature space implementing a minimum of two features [0, 1] and max of five features [0, 1, 4, 9, 10].

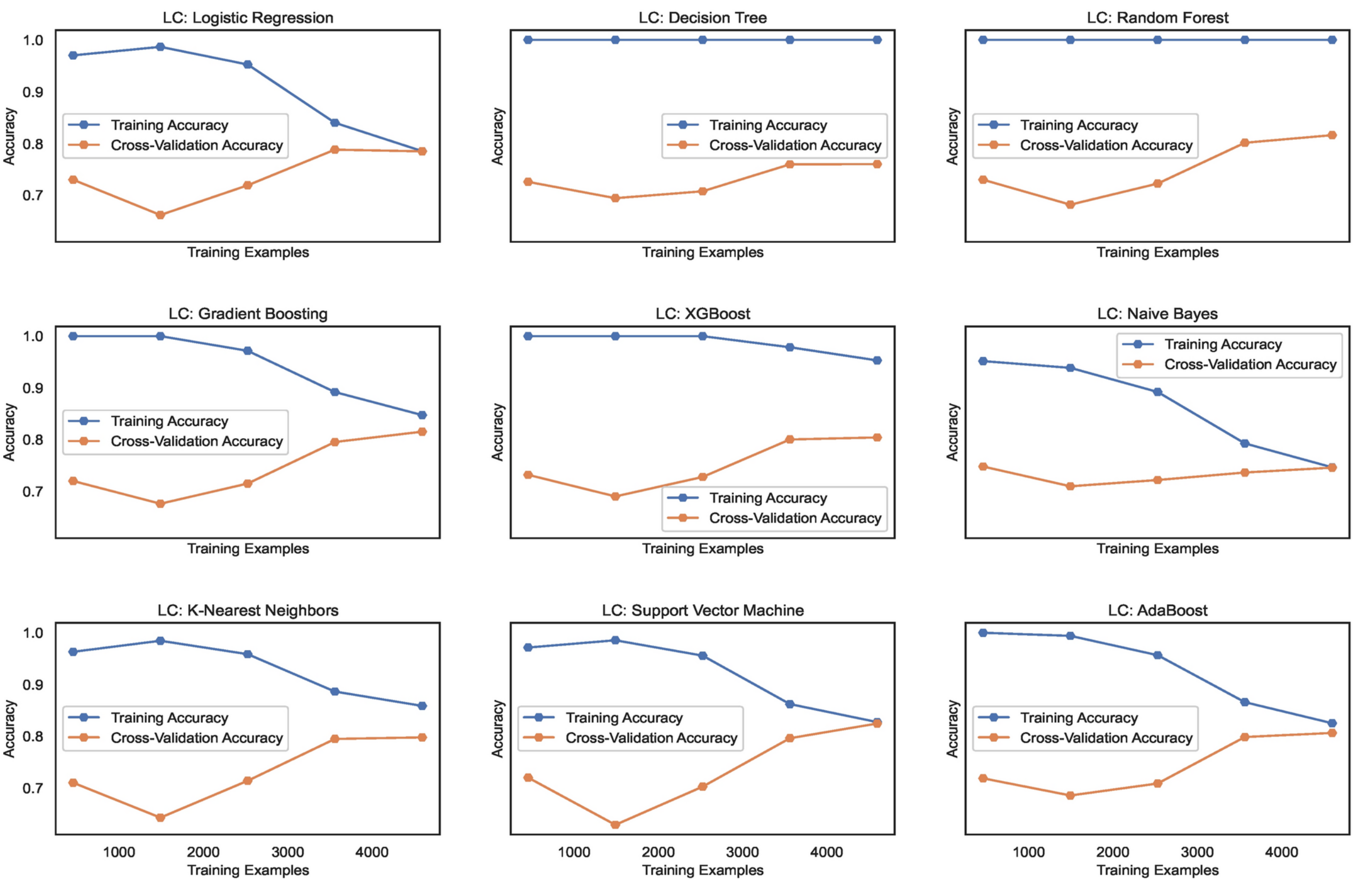

**Fig. 32**. Learning curves for models trained on $DS_3$.

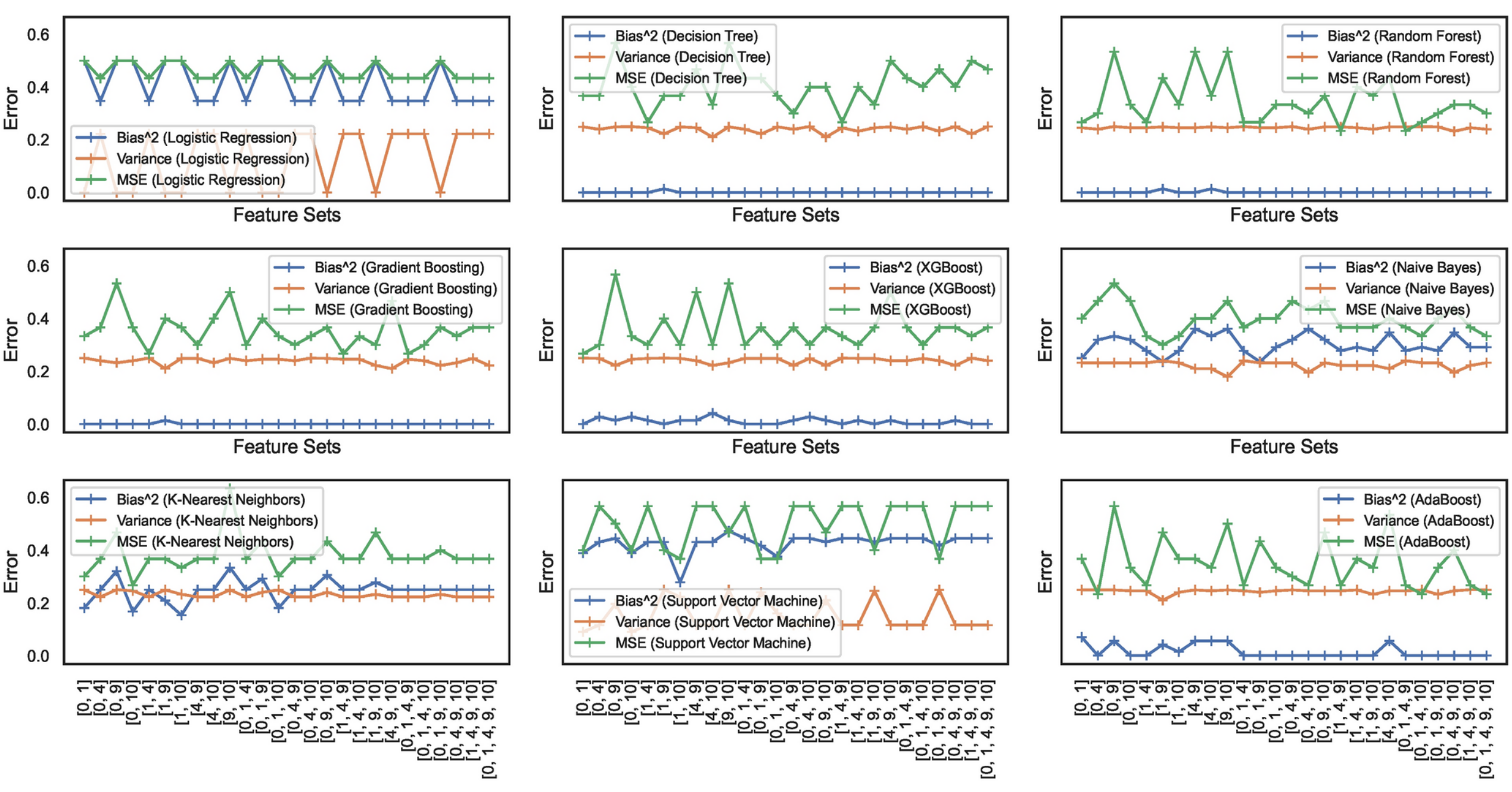

**Fig. 33**. Bias-variance curves for models trained on dataset $DS_1$.

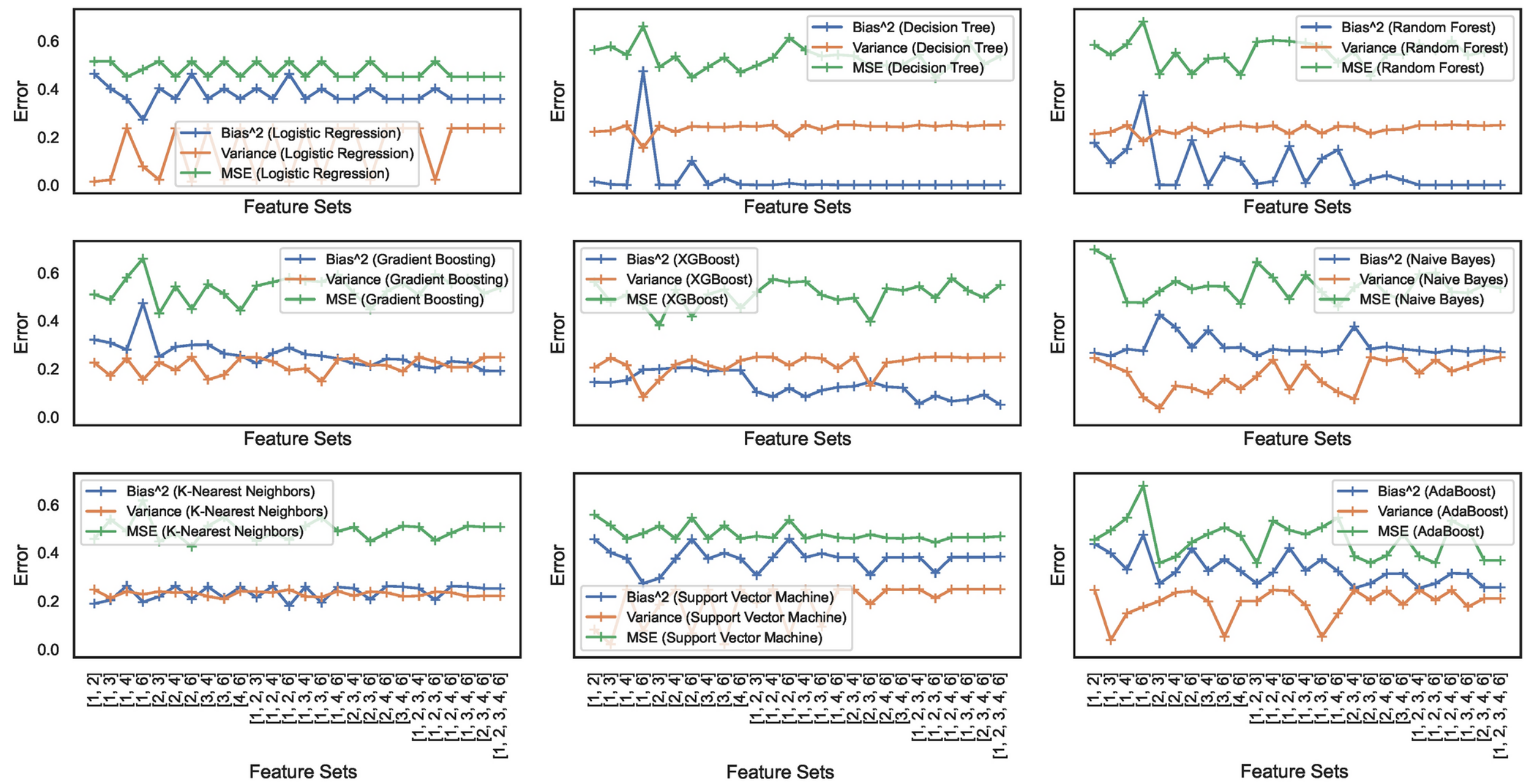

**Fig. 34**. Bias-variance curves for models trained on dataset $DS_3$.

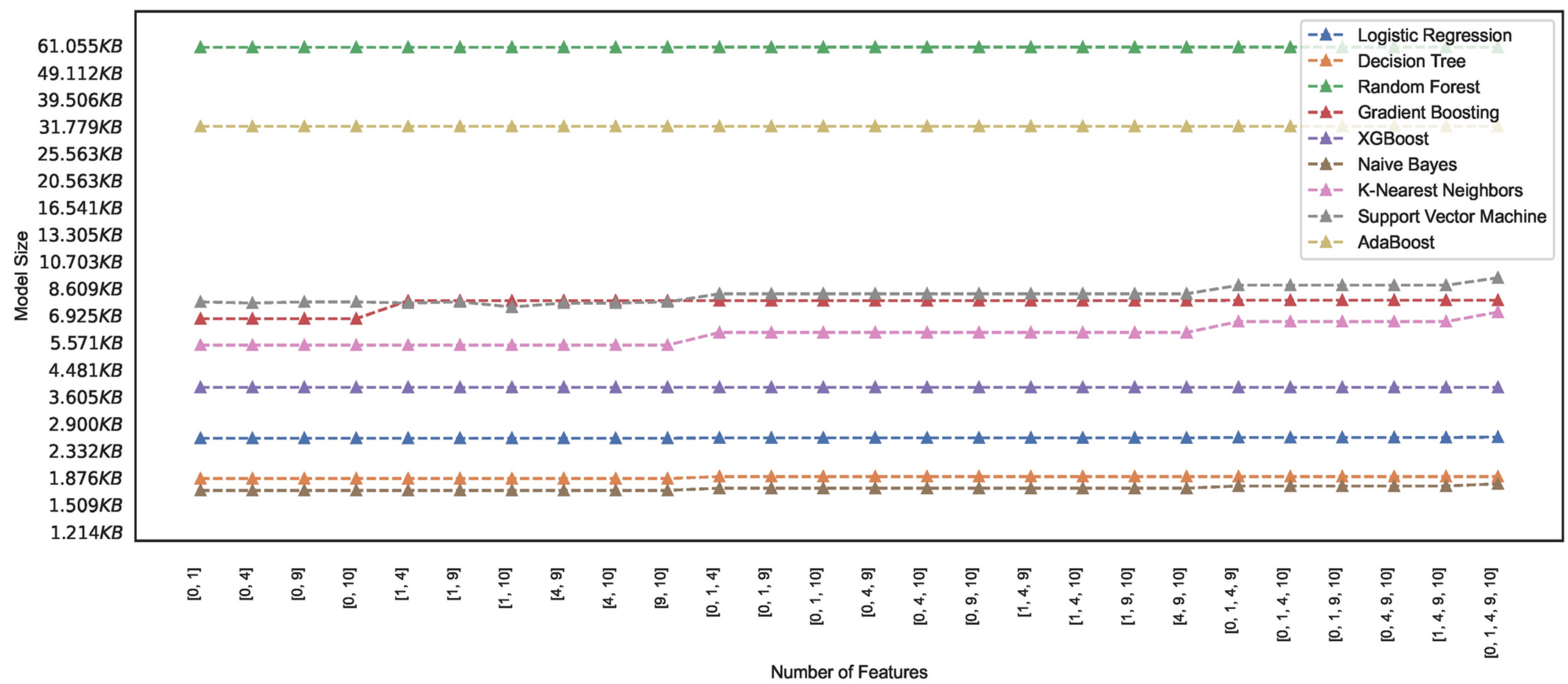

**Fig. 35**. Memory footprint of models trained on dataset $DS_1$.

Nonetheless, the models RF and ADB showcase no change in model sizes over the entire feature; they represent the models with high memory requirements for operating on canny image dataset $DS_1$.

In par with $DS_1$, the models trained on augmented canny image dataset $DS_3$ to reveal a different memory footprint as shown in Fig. 35. The study results reveal that models ADB, RF, KNN, and SVM showcase increasing trends in their memory footprint. The models' size increases with an increasing number of features and shows steep inclination at certain specific feature combinations. This implies that model size relies on feature combinations, and some models exhibit different memory footprints on different sizes of datasets employed for training the models. Increasing two to three features increases the 11KB of memory for KNN, while the same step in feature space increases the 32KB memory requirement for SVM. However, the model size for the KNN and ADB is nearly constant over the entire range of feature space, and the two models show the same memory footprint of 32KB and 61KB as it appears when the models are trained on small size canny dataset $DS_1$. On comparing the memory footprint of two models trained on two datasets $DS_1$ and $DS_3$, observation reveals that some models show consistent memory footprint while trained on both datasets. However, other models show an increasing memory footprint with increasing feature space. This concludes that some models show consistent behaviour in terms of model sizes when trained on a canny image dataset of hand-drawn spiral images of healthy and PD patients. The identified models that show constant memory space requirements can be the best match to constraint memory edge computing devices for the online classification of patients.

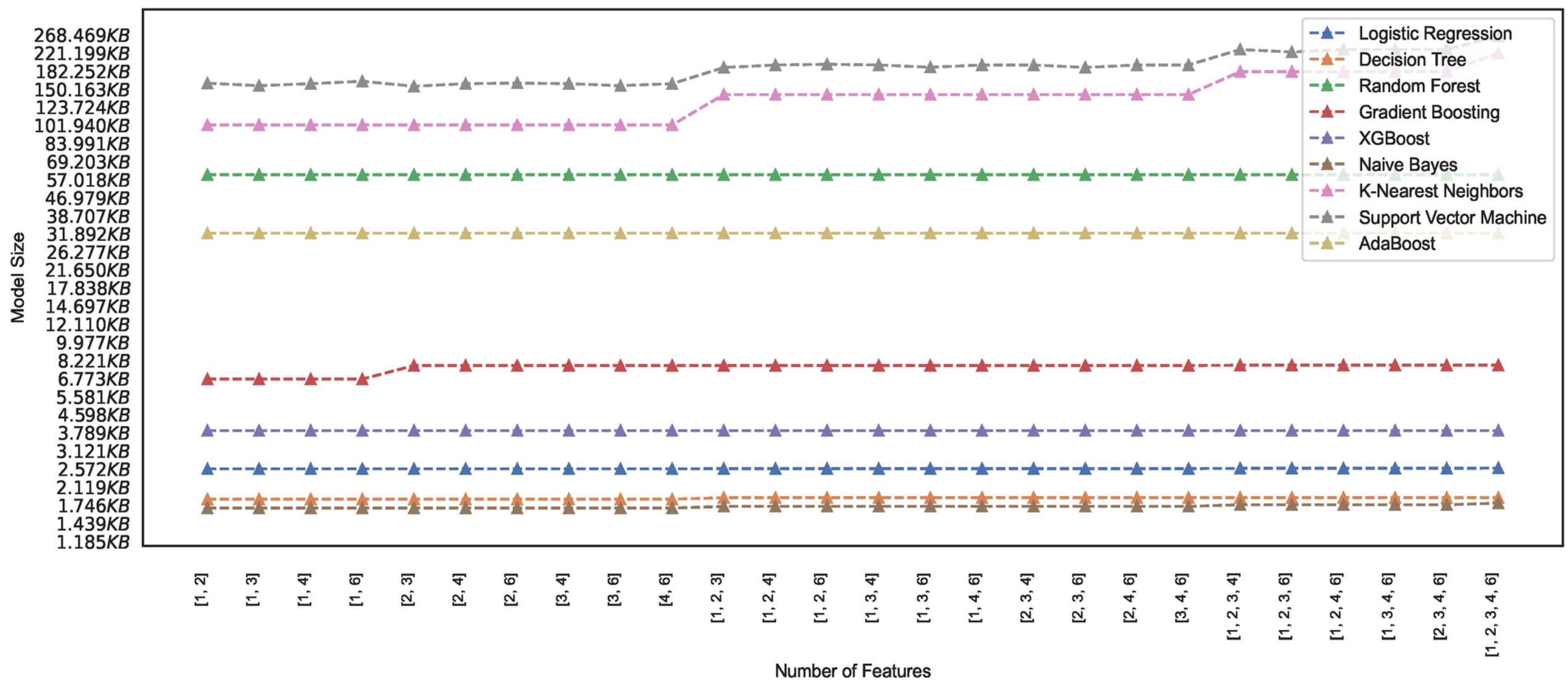

**Fig. 36**. Memory footprint of models trained on dataset $DS_3$.

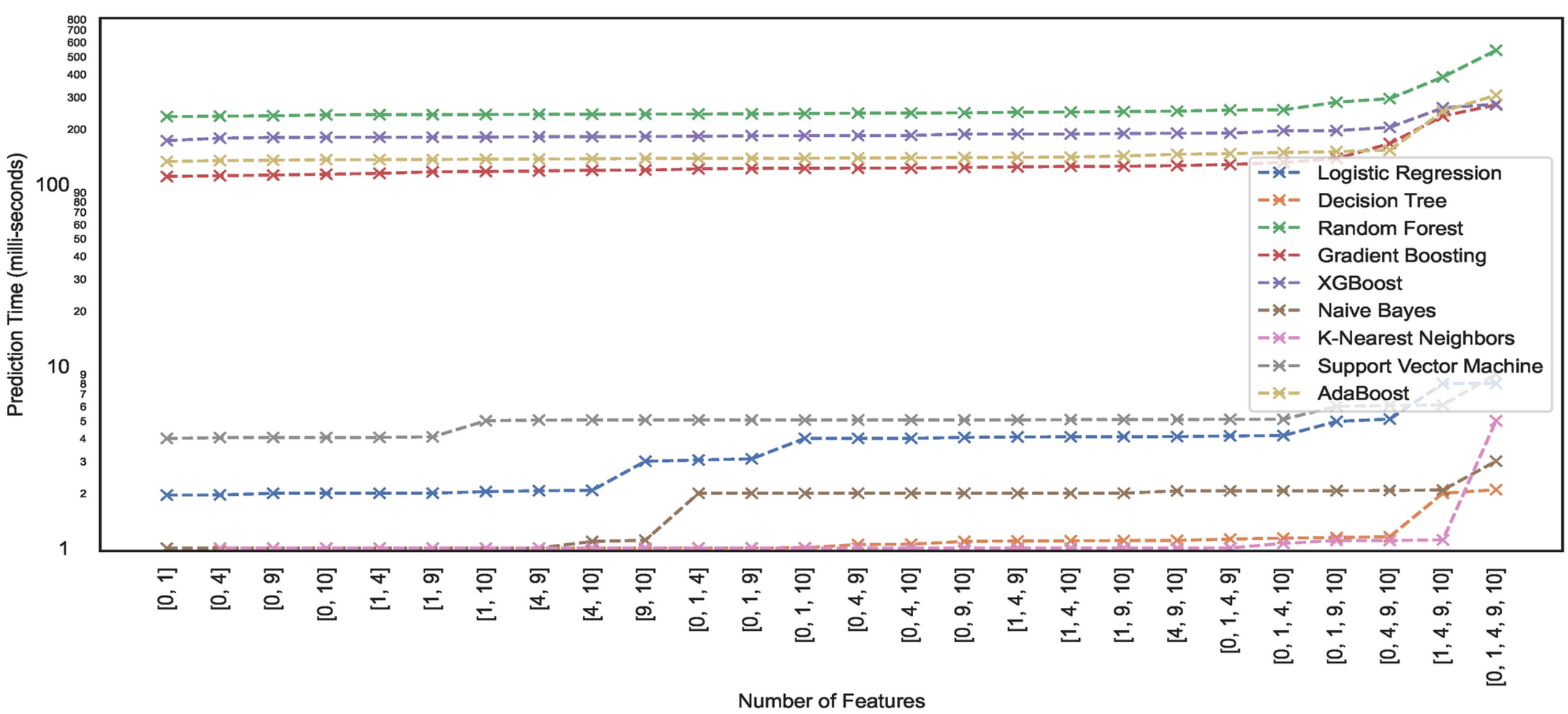

**Fig. 37**. Prediction time of models trained on dataset $DS_1$.

*Prediction time*

Figures 37 and 38 show the prediction time traced for models trained on datasets $DS_1$ and $DS_3$, respectively. The prediction time footprint in Fig. 37 shows that the nine models employed show different prediction times on the input dataset $DS_1$. While the models KNN, NB, LR, and SVM show prediction times of one order of magnitude, the other models - GB, ADB, XGB, and RF show prediction times higher than two orders of magnitude. This difference in prediction time represents a significant variation in the model prediction time. The models with low prediction times would be desirable compared to models with high prediction times. It should be noted that the models with prediction times with high orders of magnitude show almost the same prediction time over the entire range of feature space, while the models with low prediction times depict an increase in the prediction times over increasing feature space.

Observing Fig. 37 reveals a steep increase in the prediction times occurs on changing feature space from four to five features, and almost all the models exhibit this behaviour. This implies that model prediction times elevate drastically upon increasing the feature space. Since the study evaluated models on the data $DS_1$ and $DS_3$ on the features space consisting of five features only, future studies may investigate the impact of even higher numbers of features. Comparing the results of prediction times of models trained on dataset $DS_0$ to models trained on $DS_3$ shows a large variation in the model behaviours. The SVM with a prediction time of one order in magnitude shows an elevation of four orders in magnitude, A significant variation in the prediction time of the SVM model. Likewise, the models ADB, XGB, GB, and RF show an elevation of two orders in magnitude. However, the SVM model shows a minor elevation in prediction time for the increasing size of feature space. On the other hand, the models NB, KNN, and LR depict a gradual increase in the prediction times, however, just below two orders in magnitude. The DT models approach the above two orders in magnitude prediction time over the entire feature space. Overall, the models show an increasing trend in the number of features, increasing from four to five. Again, to investigate how the higher number of features impact the prediction time, a new study may be desired since the scope of this study was limited to RFS that generated five feature vectors at max.

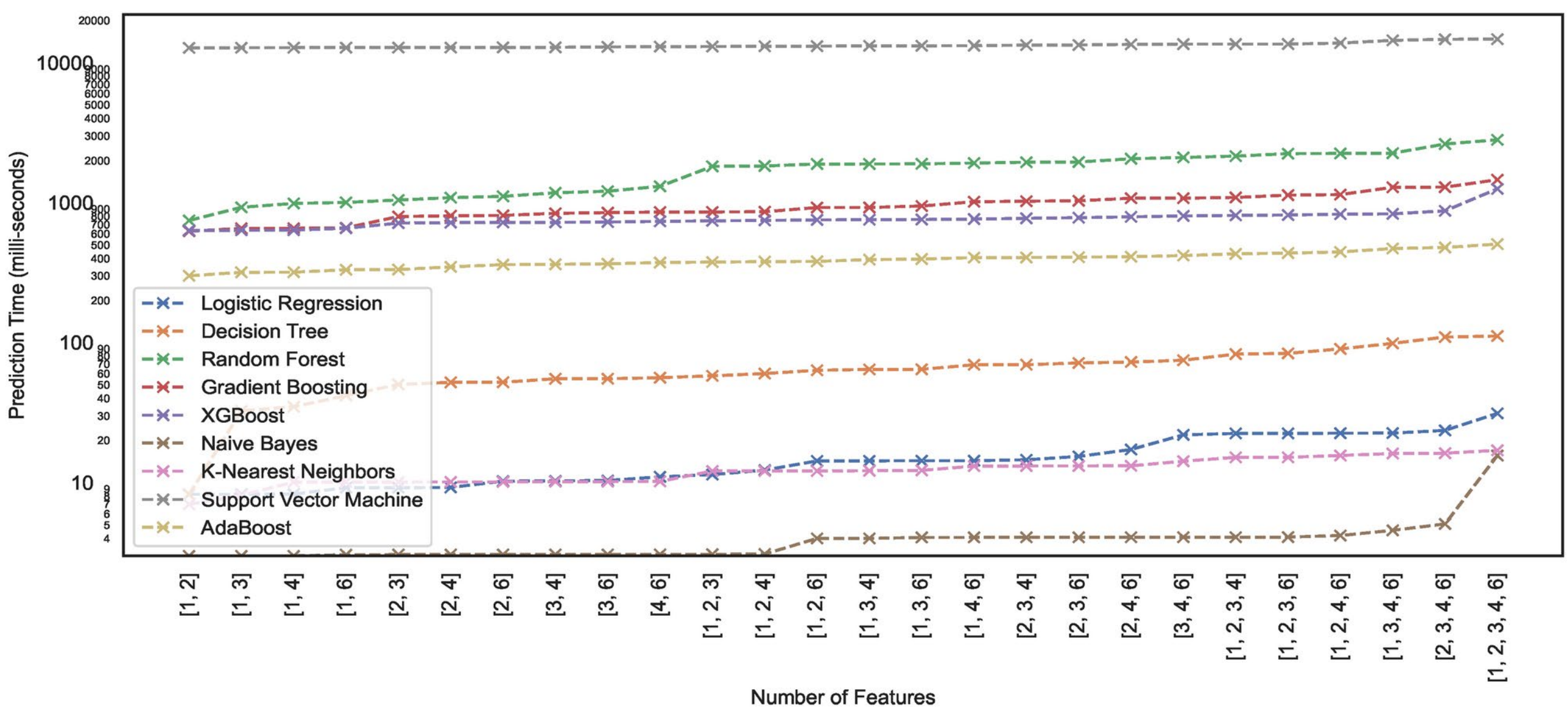

**Fig. 38**. Prediction time of models trained on dataset $DS_3$.

## Comparative analysis

*Model calibration*

Using un-calibrated models with default settings and alignment with the perfect calibration line on large datasets ($DS_2$) demonstrates promising results for PD classification. Large datasets appear to partially calibrate the models automatically, albeit not achieving the ideal S-shaped curves characteristic of fully calibrated models. The study highlights the potential benefits of calibrating models before training to enhance prediction accuracies further. Nonetheless, the impact of preprocessing with canny edge detector results in lower prediction accuracies, hence requires further investigation. Clearer interpretations of calibration curve figures (Figs. 9 and 10) would enhance the understanding of the results.

*ROC*

The ROC curves analysis demonstrates that DT and ADB models achieve the highest AUC of 0.92 on dataset $DS_0$, closely followed by RF and GB with AUC values of 0.90 and 0.91, respectively. Augmented dataset $DS_2$ significantly improves classification results, with six models achieving AUC values above 0.9 and RF, XGB, and SVM reaching the highest AUC of 0.97. On dataset $DS_1$, all classifiers show consistent AUC values ranging from 0.81 to 0.85, except for DT, which decreases to 0.78. Augmented dataset $DS_3$ also exhibits improved AUC values, with RF, GB, XGB, and SVM reaching up to 0.89. However, LR consistently performs poorly across all datasets. Overall, the study indicates that large datasets and augmentation positively impact model performance, while the canny dataset maintains consistent performances across the range of models.

*Learning curves*

The learning curve analysis provides valuable insights into the training accuracies and performance of various ML models on different datasets. On dataset $DS_0$, models exhibit a range of training accuracies, with Gradient Boosting and Decision Tree models achieving the highest accuracy of approximately 91.7%, while Logistic Regression and Naive Bayes models show lower accuracies below 75%. On dataset $DS_2$, the average accuracy of all models increases to approximately 86%, with the K-Nearest Neighbors model achieving the highest accuracy of 92%. However, the performance on datasets $DS_1$ and $DS_3$ lags behind $DS_0$ and $DS_2$, with cross-validation accuracies ranging between 71% to 83%. The impact of canny edge detection on ensuring consistent model performance is observed, but it also leads to a decrease in prediction accuracy. Overall, the study highlights the strengths and weaknesses of different ML models on different datasets, providing valuable guidance for future model selection and development in PD classification.

| | DS0 Bias | DS0 Variance | DS0 MSE | DS2 Bias | DS2 Variance | DS2 MSE |
|---|---|---|---|---|---|---|
| LR | ≈ 25 | ≈ 20 | > 25 | ≈ 25 | < 20 | > 25 |
| DT | ≈ 0 | < 20 | > 40 | ≈ 0 | < 20 | > 40 |
| RF | ≈ 0 | < 20 | > 40 | ≈ 0 | < 15 | > 40 |
| GB | ≈ 0 | ≈ 20 | ≈ 40 | < 20 | ≈ 10 | > 40 |
| XBG | ≈ 0 | ≈ 20 | > 40 | < 20 | ≈ 5 | ≈ 50 |
| NB | ≈ 25 | ≈ 20 | > 40 | > 25 | < 5 | ≈ 50 |
| KNN | < 20 | ≈ 20 | > 40 | < 20 | ≈ 15 | < 40 |
| SVM | ≈ 20 | ≈ 20 | < 40 | ≈ 25 | ≈ 10 | < 40 |
| ADB | ≈ 0 | ≈ 0 | > 40 | ≈ 25 | < 5 | ≈ 50 |

| | DS1 Bias | DS1 Variance | DS1 MSE | DS3 Bias | DS3 Variance | DS3 MSE |
|---|---|---|---|---|---|---|
| LR | ≈ 40 | ≈ 20 | ≈ 50 | ≈ 40 | > 40 | ≈ 10 |
| DT | ≈ 0 | ≈ 20 | ≈ 40 | ≈ 0 | ≈ 20 | > 50 |
| RF | ≈ 0 | ≈ 20 | > 25 | < 5 | ≈ 25 | > 50 |
| GB | ≈ 0 | ≈ 25 | ≈ 40 | ≈ 20 | ≈ 20 | > 50 |
| XBG | ≈ 0 | ≈ 25 | ≈ 40 | ≈ 10 | ≈ 20 | > 50 |
| NB | > 25 | > 25 | ≈ 40 | ≈ 25 | ≈ 15 | > 50 |
| KNN | ≈ 25 | ≈ 25 | ≈ 40 | ≈ 20 | ≈ 20 | > 50 |
| SVM | > 40 | < 20 | > 40 | ≈ 40 | < 20 | ≈ 40 |
| ADB | ≈ 0 | ≈ 25 | ≈ 40 | ≈ 25 | ≈ 20 | > 40 |

**Fig. 39**. Summary of results approximating bias and variance for four trained models. The values shown are expressed as percentages.

| | DS0 | DS2 | Size Increase |
|---|---|---|---|
| LR | 2.6 | 2.6 | × |
| DT | 1.9 | 1.9 | × |
| RF | 61 | 61 | × |
| GB | 7 | 7-8.6 | ✓ |
| XBG | 3.8 | 3.8 | × |
| NB | 1.7 | 1.7 | × |
| KNN | 5.7-7 | 102-220 | ✓ |
| SVM | 7 | 113-304 | ✓ |
| ADB | 32 | 32 | × |

| | DS1 | DS3 | Size Increase |
|---|---|---|---|
| LR | 2.6 | 2.6 | × |
| DT | 1.9 | 1.9 | × |
| RF | 61 | 61 | × |
| GB | 7 | 7-8.6 | ✓ |
| XBG | 3.8 | 3.8 | × |
| NB | 1.7 | 1.7 | × |
| KNN | 5.7-7 | 102-220 | ✓ |
| SVM | 7-7.5 | 150-268 | ✓ |
| ADB | 32 | 32 | × |

**Fig. 40**. Summary of results approximating memory footprint for four trained models. The values shown represent the model sizes expressed in kilo bytes.

*Confusion matrix*

The analysis of confusion matrices provides valuable insights into the performance of the nine models on different datasets. Nearly all classifiers perform well on dataset $DS_0$ except for the LR model, which shows higher counts of FPs and FNs. The DT model stands out with the lowest FPs and FNs and the highest TPs and TNs, indicating its superior classification ability on a small dataset. Conversely, on the large dataset $DS_2$ , significant improvements in TP, TN, FP, and FN values are observed for all models. The KNN model achieves the lowest number of FPs and FNs and the highest TPs and TNs among all models, indicating its increased classification accuracy with the larger dataset. However, on the canny edge, detection-based dataset $DS_3$, all models exhibit consistent behaviour in identifying TPs and TNs, with lower accuracy levels than datasets $DS_0$ and $DS_2$. This suggests that canny edge detection enables consistent model performance; nonetheless, it may compromise overall model accuracy.

*Bias-variance*

Bias and variance are critical factors that impact the performance of ML models. Different models on datasets $DS_0$ and $DS_2$ showcase varying bias-variance trade-offs at specific feature combinations. While some models exhibit near-perfect trade-offs with low feature counts, others show dynamic behaviours depending on the model type and feature combinations. However, for feature sets handcrafted from augmented dataset $DS_2$, finding a feature set enabling a perfect trade-off remains an open question that requires further investigation. The models trained on canny image dataset $DS_1$ and dataset $DS_3$ show varying bias-variance trade-offs, with KNN demonstrating better performance in achieving optimal fit. Further individual-level analysis is needed to effectively generalize the bias-variance trade-off for the nine models. Continued investigation in this area will contribute to a deeper understanding of model behaviour and guide the selection of appropriate features and model types for improved performance in classifying spiral curves of healthy and PD patients. Figure 39 shows the summarized results of the bias variance statistics for the four different datasets.

*Model complexity*

Memory management is crucial for efficiently deploying ML models, especially in resource-constrained environments. The study analyzes ML models' memory footprints for classifying healthy and PD patients. The results, depicted in Figs. 19, 20, 35, and 36, show variations in memory requirements among the models. Models like RF and ADB exhibit higher memory requirements compared to others. However, most models trained on small datasets show consistent memory requirements, while models trained on augmented datasets showcase varying memory footprints. Comparing the memory footprints of models trained on different datasets, it is evident that some models display consistent memory requirements across both datasets, while others exhibit an increase in memory footprint with an increase in feature space. The study suggests selecting models with consistent memory requirements for edge computing devices with memory constraints. Models like KNN and GB exhibit consistent memory usage and are suitable for resource-constrained devices. Prediction time is crucial in ML models, particularly in edge computing environments with limited computational resources. The study evaluates the prediction times of the nine models used for classifying hand-drawn spirals of healthy and PD patients. The results, as shown in Figs. 21 and 22, demonstrate significant variations in prediction times among the models. Models like KNN, NB, LR, and SVM exhibit prediction times of one order of magnitude, while GB, ADB, XGB, and RF have prediction times higher than two orders of magnitude. Models with lower prediction times are preferred for edge computing environments. Additionally, the study observes a steep increase in prediction times when increasing the feature space from four to five features, indicating the impact of feature space size on prediction times.

| | DS0 | DS2 | Prediction |
|---|---|---|---|
| LR | 3-5 | 1-7 | fast |
| DT | 1-1.2 | 15-100 | fast |
| RF | 250-380 | 900-3000 | slow |
| GB | 120-150 | 800-2800 | slow |
| XBG | 180-200 | 700-3000 | slow |
| NB | 2 | 2-8 | fast |
| KNN | 1-2 | 8-40 | fast |
| SVM | 4-5 | 9000-30000 | slow |
| ADB | 150-180 | 350-1500 | slow |

| | DS1 | DS3 | Prediction |
|---|---|---|---|
| LR | 2-10 | 7-30 | fast |
| DT | 1-2 | 7-90 | fast |
| RF | 250-600 | 700-2500 | slow |
| GB | 120-250 | 600-1000 | slow |
| XBG | 180-250 | 600-1000 | slow |
| NB | 1-3 | 1-15 | fast |
| KNN | 1-4 | 7-15 | fast |
| SVM | 4-8 | 15000-18000 | slow |
| ADB | 150-180 | 300-600 | slow |

**Fig. 41**. Summary of results approximating prediction time analysis for four trained models. The values shown represent the model prediction times expressed in milli-seconds.

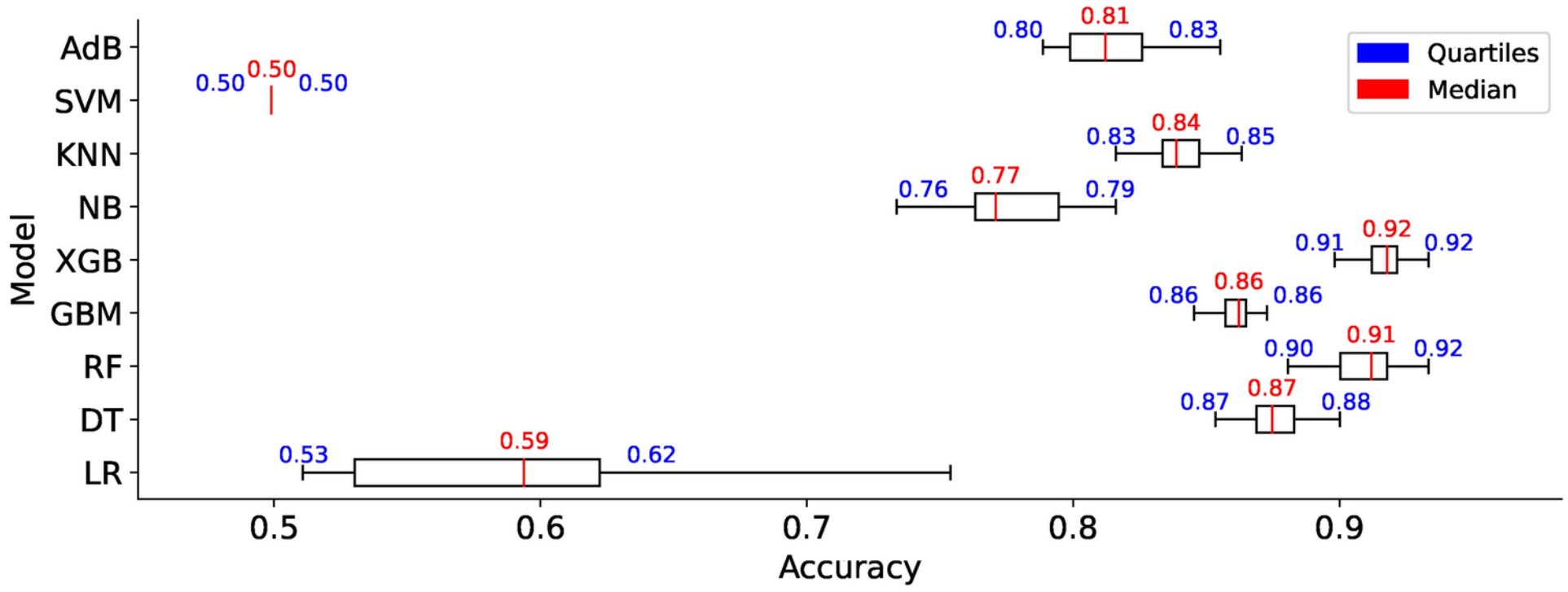

**Fig. 42**. Comparison of accuracies of models trained on $DS_2$.

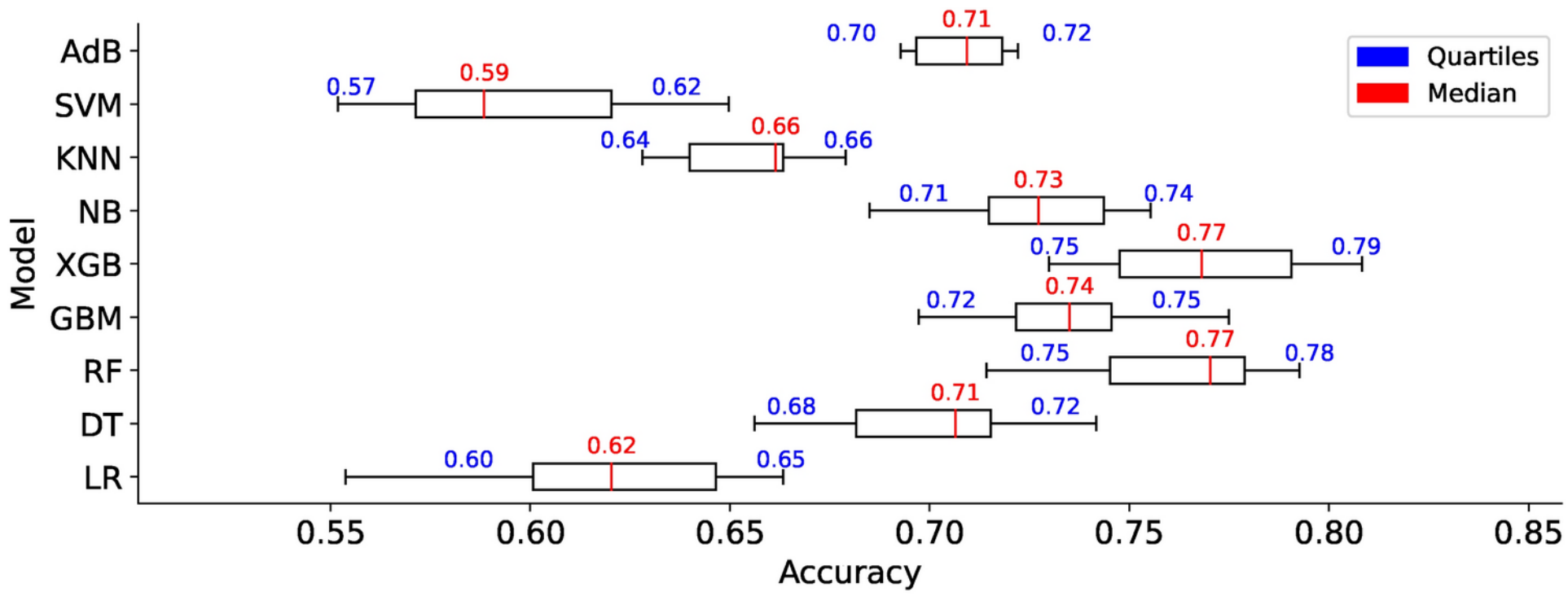

**Fig. 43**. Comparison of accuracies of models trained on $DS_3$.

*Implications of memory footprint and prediction time*

In this study, we estimated the models' memory footprint and prediction time to assess their suitability for deployment on mobile and resource-constrained devices. Figures 40 and 41 provides the summary of model sizes and prediction time behaviour towards the four datasets. Fast prediction times are essential in real-time applications, such as healthcare, where timely and accurate results are crucial. For instance, in diagnosing

Parkinson's disease from medical images, quick model responses ensure that healthcare professionals can make rapid decisions, leading to more effective and timely treatments. On the other hand, memory footprint determines how efficiently a model can run on devices with limited storage and processing power. Considering the memory footprint, we ensured the models could be deployed on devices like smartphones or embedded systems without compromising performance. Smaller memory footprints allow models to run more efficiently on such devices, enhancing scalability and making them practical for real-world applications. Incorporating low memory usage and fast prediction times is critical to creating models that are accurate and feasible for deployment in environments with resource limitations. This ensures that machine learning models can deliver quick, reliable results while remaining computationally efficient, especially for critical applications in healthcare and other mobile-driven fields.

## Statistical validation - model comparison

*Normality test*

The study employs pairwise comparisons of the ML models to determine the statistical significance among the nine machine-learning models based on their computed accuracies. We choose a sample size of $N = 100$ observations (accuracies based on $10 - fold$ cross-validation) per model, with $(df = 98)$ degrees of freedom for each two compared models. The study employs the Shapiro-Wilk normality test[72] to assess whether the collected data of model accuracies for each model follows a normal distribution. The results of the Shapiro-Wilk test showcase a non-Gaussian or non-normal distribution of the collected data. Therefore, the study applies a non-parametric test to statistically compare and find differences between the models trained on the four different datasets. **Note:** The study specifically evaluates differences for the augmented datasets $DS_2$ and $DS_3$ only since the $DS_0$ and $DS_1$ are small-sized datasets. Figures 42 and 43 showcase a comparison of the models trained on the two augmented datasets. Most models in Fig. 42 exhibit accuracies above 75%, except for the LR and SVM models, which remain below a 70% accuracy level. As observed from the figure, the median skews in most models, indicating a non-Gaussian distribution. In contrast to the models in Fig. 42, the larger boxes in Fig. 43 indicates greater variability within the observed accuracies. In other words, the wider interquartile range (IQR) suggests more dispersed data points, reflecting varying prediction accuracies in models as a consequence of training on $DS_3$.

*Non-parametric test*

The Mann-Whitney U test[73] is a non-parametric statistical test used to assess whether two independent groups exhibit significant differences in their continuous (ordinal or interval level) dependent variable distributions. It is an alternative to the independent samples t-test when normality and variance homogeneity assumptions are unmet. A significant two-tailed p-value greater than 0.05 indicates that the observed data lacks statistical significance. Notably, the Mann-Whitney test does not assume the data's normality, making it robust against violations of normal distribution assumptions. We apply the Mann-Whitney test to compare the nine models trained on the two augmented datasets. These models are trained with optimal feature vector sets that were determined to yield higher accuracies in previous analyses. For the Mann-Whitney test, we define the **Null Hypothesis:** There is no difference between the two datasets $DS_2$ and $DS_3$, and The **alternative hypothesis:** states that a significant difference exists between these datasets, $DS_2$ and $DS_3$. The Mann-Whitney U p-value indicates the probability of observing data as extreme as the observed results under the assumption of no effect (null hypothesis). However, it considers only the direction specified by the alternative hypothesis. A large two-tailed p-value $> 0.05$, suggests that the observed data lacks statistical significance. This indicates insufficient evidence to reject the null hypothesis in favor of the alternative hypothesis. Figure 44 illustrates the results of the Mann-Whitney U test comparing models trained on the two datasets. The figure clearly indicates that most models exhibit significant differences in their prediction performances, except for the comparisons between SVM and LR, and XGB and NR, which show no statistically significant differences. However, the results for the other models demonstrate significant differences, leading to the acceptance of the alternative hypothesis.

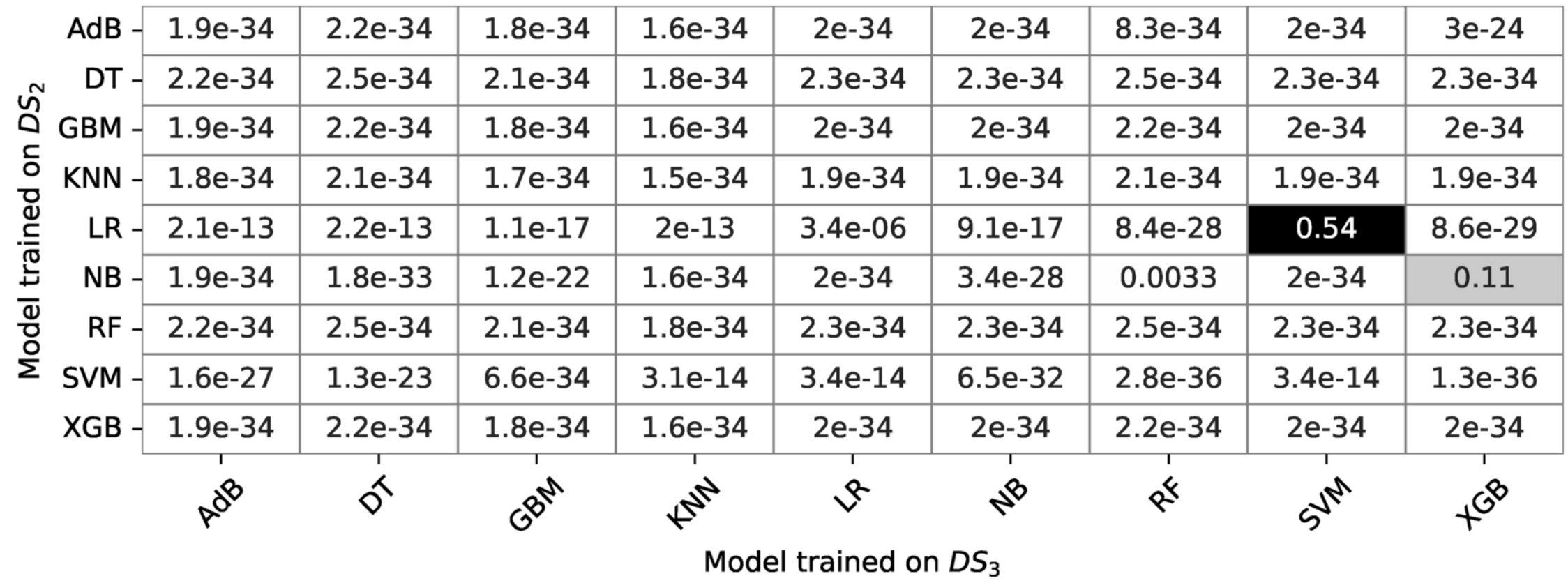

**Fig. 44**. Model comparison: Mann-Whitney U test results for models trained on $DS_2$ and $DS_3$.

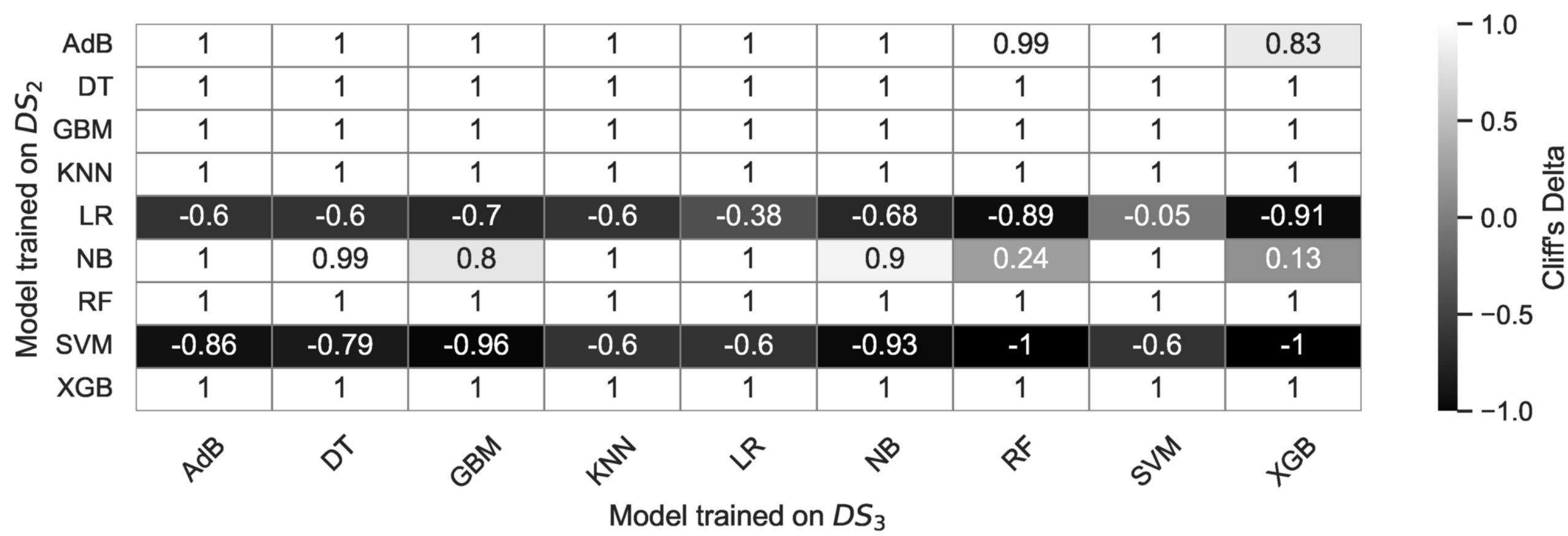

**Fig. 45**. Model comparison: Estimated Cliff's Delta $\delta_C$ values for the compared models trained on $DS_2$ and $DS_3$.

*Interpretation of Mann-Whitney U test results*
The Mann-Whitney U test results shown in the Fig. 44 reveal the statistical comparisons of nine models trained on two different datasets. Each value represents the p-value obtained by applying the Mann-Whitney U test to the accuracy scores of two models, assessing whether there is a significant difference in their performance.

*Key observations*
Very small p-values ($< 0.05$)—For several pairs of models, the p-values are extremely small (e.g., 1.9e-34, 2.2e-34), suggesting that these models differ significantly in their performance. For example, models like AdB vs. DT, or XGB vs. LR, show significant differences in their accuracies. These results indicate that one model outperforms the other consistently across both datasets.

Larger p-values ($\geq 0.05$)—A few p-values, such as 0.54 (between models like RF and SVM), suggest no significant difference in model performance. These results indicate that the models perform similarly across both datasets. This implies that, in certain conditions, the performance differences between these models may not be substantial, and the choice between them may depend on other factors, such as computational complexity or interpretability.

Model Performance Patterns—Models such as AdB, DT, GBM, and SVM appear to consistently show small p-values, indicating that they are significantly different from other models like LR and NB. This pattern suggests that ensemble methods like AdB and GBM could potentially be more robust across datasets compared to simpler models like LR or NB, which show less variability in their performance. However, it is important to note that ensemble methods are not considered in the current study, but they could be explored in future work. Conversely, models like XGB exhibit a larger range of p-values, indicating that its performance can vary significantly depending on the dataset. This could reflect its sensitivity to hyperparameters or dataset characteristics, which may require fine-tuning to achieve optimal performance.

*Effect size measures*
Cliff's Delta[74] is a non-parametric effect size measure used to assess the magnitude of differences between two distributions, particularly when comparing the performance of two models or groups. Unlike traditional measures such as the t-test, which assumes the normality of the data, Cliff's Delta is robust to non-normal distributions. It can be used with ordinal or continuous data. It quantifies the probability that a randomly selected value from one distribution will be larger or smaller than a randomly selected value from another distribution.

A Cliff's Delta ($\delta_C$) value of 1 indicates a strong performance difference where Model 1 trained on $DS_2$ significantly outperforms Model 2 trained on $DS_3$, while $-1$ suggests the opposite, with Model 2 being the superior model. Values close to 0 imply little to no difference in performance. The value of $\delta_C$ ranges from $-1$ to $+1$, with $-1$ indicating no overlap, $+1$ indicating complete separation, and 0 indicating full overlap. Unlike Standardized Mean Difference (SMD), Cliff's $\delta_C$ measures the effect across the entire distribution, not just the center, making it more comprehensive. This metric is especially useful in machine learning model comparisons as it provides a clear, robust way to evaluate models without relying on assumptions about the underlying data distribution. For observed model accuracies, let $n_X$ represent the number of observations for Model 1 and $n_Y$ represent the number of observations for Model 2. Cliff's delta $\delta_C$ is expressed by the following Eq. (21).

$$\delta_C = \frac{1}{n_X \cdot n_Y} \sum_{i=1}^{n_X} \sum_{j=1}^{n_Y} \left[ \Phi(x_i > y_j) - \Phi(x_i < y_j) \right], \quad \delta_C \in [-1, 1] \tag{21}$$

The $\Phi(\cdot)$ is an indicator function, returning 1 if the condition inside is true and 0 otherwise. The subscripts *i* and *j* have the following meanings: *i* represents the index of an individual observation from the first model (Model 1), where $x_i$ is the *i*-th observed accuracy in Model 1. Similarly, *j* represents the index of an individual observation from the second model (Model 2), where $y_j$ is the *j*-th observed accuracy in Model 2.

We estimate Cliff's Delta to compare the performance of different machine learning models against each other. By calculating Cliff's Delta for each pairwise comparison between models, we were able to assess the magnitude and direction of performance differences (see Fig. 45). A positive Cliff's Delta indicated that Model 1 outperformed Model 2, while a negative value suggested the opposite. For example, when comparing Logistic Regression (LR) with Decision Trees (DT), a negative value $-0.38$ indicated that the Decision Tree model performed better. On the other hand, when comparing models like Decision Trees (DT) with themselves, the value of 1 indicated identical performance. This method allowed us to quantitatively measure and interpret performance differences between pairs of models, providing insights into which models consistently performed better or worse across different comparisons. More specifically, the positive values, such as 1 and values close to 1 (e.g., 0.98, 0.9), indicate that Model 1 outperforms Model 2, with values closer to 1 signifying a stronger performance of Model 1. The Negative values, such as $-0.38$, $-0.6$, and $-0.89$, indicate that Model 1 (e.g., LR, DT, RF) performs worse than Model 2 (e.g., LR, DT, RF), with larger negative values suggesting a more significant difference in performance. A value of 0 represents cases where the two models have similar performance, such as the "SVM vs. LR" comparison, where both models show equal accuracy.

## Overall findings and recommendations

Unlike prior studies that often focus on a single preprocessing method or dataset, this work systematically evaluates the combined effects of dataset augmentation and preprocessing (e.g., Canny edge detection and Hessian filtering) across multiple datasets. While earlier research primarily reported performance improvements with preprocessing, our results reveal that Canny preprocessing degrades model accuracy across various ML models, challenging established assumptions. Additionally, previous studies rarely investigated the interplay between dataset size, model memory footprint, and prediction latency. In contrast, our detailed analysis demonstrates significant increases in memory usage and latency for models like *SVM* and *XGB* with larger datasets, providing a more nuanced understanding of scalability. The use of robust statistical measures, such as Mann-Whitney U test and Cliff's delta, showcases that our comparisons are statistically grounded, offering a rigorous framework absent in much of the prior work. These advancements establish a more comprehensive perspective on ML model evaluation, particularly in resource-constrained healthcare applications.

Our analysis provides key insights into the trade-offs between memory footprint, prediction time, and model performance, which are crucial for selecting machine learning models in resource-constrained environments. We observed significant variations in memory usage among different models, with some showing consistent memory requirements regardless of dataset size, while others, especially those trained on augmented datasets, exhibited more variability. This highlights the importance of selecting models that offer consistent memory efficiency, particularly for edge computing applications where memory resources are limited. Prediction time also emerged as a critical factor, with simpler models like KNN and NB exhibiting faster prediction times, whereas more complex models such as XGB and RF showed significantly higher times. This underscores the need to balance model accuracy with prediction time, especially in real-time or computationally constrained environments.

The use of statistical techniques, including the Mann-Whitney U test and Cliff's Delta, further enhanced our understanding of the performance differences between model pairs. These analyses not only quantified the statistical significance of performance differences but also provided insights into the magnitude of the effect, helping identify models that consistently outperform others. Our findings suggest that models like DT, KNN and ADB, which show stable performance across datasets and exhibit favorable memory and time efficiencies, are more suitable for resource-constrained environments. In contrast, more complex models may require further optimization to achieve a balance between performance and resource usage, making simpler models preferable for deployment in environments with limited computational resources.

## Limitations and future research

The use of spiral drawing data for Parkinson's disease (PD) classification, while valuable, has certain limitations that should be addressed in future research. Spiral drawing tasks, though useful for assessing motor control and tremor, may not capture the full spectrum of PD-related symptoms, such as cognitive decline or non-motor symptoms. Additionally, this approach is limited by inter-individual variability, as factors like motor skill, drawing ability, or even other neurological conditions can affect the accuracy of classification models. Combining spiral drawing data with other diagnostic methods, such as gait analysis, speech analysis, or neuroimaging, could provide a more comprehensive and robust understanding of PD. This multi-modal approach would enable the integration of both motor and non-motor features, potentially improving classification accuracy and broadening the applicability of machine learning models in clinical settings.

In the future, exploring calibration classification tasks and its implications could address limitations and offer valuable insights for advancing PD diagnosis using ML models. Future research in this area may pave the way for more reliable and interpretable clinical decision support systems by focusing on improved calibration and explainability. The study highlights the potential benefits of calibrating models before training to enhance prediction accuracies further. Nonetheless, the impact of preprocessing with canny edge detector on lower prediction accuracies requires further research. More detailed individual-level analysis is needed to effectively generalize the bias-variance trade-off for the nine models. Continued investigation in this area will contribute to a deeper understanding of model behaviour and guide the selection of appropriate features and model types for improved performance in classifying spiral curves of healthy and PD patients. The study provides valuable insights into the memory management of ML models, aiding in selecting appropriate models for edge computing and other memory-constrained applications. However, further research is needed to explore the memory behaviours of models on larger feature spaces and different datasets for a comprehensive understanding of model memory requirements. Comparing models trained on datasets $DS_0$ and $DS_3$, significant variations in prediction time behaviours are noted, with SVM showing a notable elevation of four orders of magnitude. The study suggests that future research should explore the impact of higher numbers of features on prediction times for a more comprehensive understanding of model behaviours.

To build on these findings, future research should focus on combining spiral drawing data with other biomarkers or clinical tests to improve the generalizability of the models. Further exploration of data calibration methods, particularly in relation to feature selection and the impact of different preprocessing techniques, is needed to enhance prediction accuracy. Additionally, investigating the potential of hybrid models that incorporate both traditional clinical diagnostic methods and machine learning predictions could be an exciting next step. This approach may not only provide more reliable results but also increase the interpretability and clinical utility of the models, paving the way for more effective decision support systems in PD diagnosis and treatment planning.

*Implications for long-term monitoring and disease progression*
To address the implications of our findings for long-term monitoring, we emphasize the importance of considering changes in handwriting due to the progression of Parkinson's Disease. Handwriting patterns are known to deteriorate over time, which could influence the performance of machine learning models trained on static datasets. Incorporating longitudinal data, capturing handwriting samples at different stages of the disease, can provide valuable insights into these progressive changes. This approach would enhance the robustness of the models by enabling them to account for temporal variations, making them more applicable for long-term disease monitoring. Future research can focus on developing models that adapt to these progressive shifts, ensuring reliable predictions over extended periods and facilitating personalized tracking of disease progression.

## Conclusion
In conclusion, this study thoroughly investigated machine learning (ML) models for hand-drawn spiral classification in both healthy individuals and Parkinson's disease (PD) patients. We scrutinized model performance, bias-variance trade-offs, prediction times, and memory footprints, providing nuanced insights into their practical suitability. Notably, Decision Tree and Adaptive Boosting models exhibited strong performance on the original dataset, while Random Forest, Extreme Gradient Boosting, and Support Vector Machines excelled on augmented datasets. The analysis of bias-variance trade-offs and prediction times holds particular significance, especially in resource-constrained edge computing environments. Efficient memory management emerged as critical, with certain models displaying consistent memory requirements across diverse datasets. These findings offer valuable guidance for ML model selection and deployment in PD diagnosis, emphasizing the ongoing need for research into larger feature spaces and diverse datasets. This trajectory aims to refine the real-world applicability of these models, aligning them more effectively with the dynamic landscape of PD diagnosis and beyond.

## Data availability
No datasets were generated or analysed during the current study. The study employed open dataset available for download at https://www.kaggle.com/datasets/kmader/parkinsons-drawings.

## Acknowledgements

This work was supported by the Faculty of Electronics, Telecommunications, and Informatics, Gdansk University of Technology, Poland.

## Author contributions

S.B. conceived the experiment, S.B. and P.S. conducted the experiment, S.B. and P.S. analysed the results. All authors reviewed the manuscript.

## Declarations

### Competing interests
The authors declare no competing interests.

### Additional information
**Correspondence** and requests for materials should be addressed to S.B.

**Reprints and permissions information** is available at www.nature.com/reprints.

**Publisher's note** Springer Nature remains neutral with regard to jurisdictional claims in published maps and institutional affiliations.